\documentclass{article}

\PassOptionsToPackage{numbers,sort&compress}{natbib}

\usepackage[preprint]{neurips_2026}

\usepackage[utf8]{inputenc} 
\usepackage[T1]{fontenc}    
\usepackage{hyperref}       
\usepackage{url}            
\usepackage{graphicx}       
\usepackage{capt-of}        
\usepackage{wrapfig}        
\usepackage{subcaption}     
\usepackage{booktabs}       
\usepackage{amsfonts}       
\usepackage{amsmath}        
\usepackage{enumitem}       
\usepackage{nicefrac}       
\usepackage{microtype}      
\usepackage[dvipsnames]{xcolor}         
\usepackage{tikz}          
\usepackage{pifont}         

\usepackage{tabularx}
\usepackage{array}
\usepackage{longtable}
\usepackage{xltabular}      
\usepackage{seqsplit}       
\usepackage{colortbl}       
\newcolumntype{L}[1]{>{\raggedright\arraybackslash}p{#1}}
\newcolumntype{C}[1]{>{\centering\arraybackslash}p{#1}}
\newcolumntype{Y}{>{\raggedright\arraybackslash}X}
\newcommand{\cmark}{\textcolor{ForestGreen}{\ding{51}}}
\newcommand{\mmark}{\textcolor{BurntOrange}{\(\triangle\)}}
\newcommand{\xmark}{\textcolor{BrickRed}{\ding{55}}}

\hypersetup{
    colorlinks=true,
    linkcolor=RoyalBlue,
    citecolor=RoyalBlue,
    urlcolor=RoyalBlue
}

\usepackage{xspace}
\newcommand{\bench}{MLS-Bench\xspace}
\newcommand{\numtasks}{140\xspace}

\usepackage{listings}
\usepackage[framemethod=tikz]{mdframed}
\usepackage{xstring}
\IfFileExists{inconsolata.sty}{\usepackage{inconsolata}}{}  

\definecolor{ealEdit}{HTML}{1A7F37}      
\definecolor{ealEditable}{HTML}{0969DA}  
\definecolor{ealNum}{HTML}{8C8C8C}
\definecolor{openaiTint}{HTML}{E8F7F4}
\definecolor{anthropicTint}{HTML}{F8EDE5}
\definecolor{googleTint}{HTML}{E8F0FE}
\definecolor{deepseekTint}{HTML}{EAEDFF}
\definecolor{qwenTint}{HTML}{EEE9F8}
\definecolor{defaultTint}{HTML}{F0F0F0}
\definecolor{openaiRule}{HTML}{10A37F}
\definecolor{anthropicRule}{HTML}{D97757}
\definecolor{googleRule}{HTML}{4285F4}
\definecolor{deepseekRule}{HTML}{4D6BFE}
\definecolor{qwenRule}{HTML}{6B4FBB}
\definecolor{defaultRule}{HTML}{888888}

\lstdefinestyle{eal/python}{%
  language=Python, basicstyle=\ttfamily\scriptsize,
  keywordstyle=\color{black},
  commentstyle=\color{ealNum}\itshape,
  stringstyle=\color{black},
  numbers=none,
  frame=leftline, framerule=0.4pt, framesep=4pt, rulecolor=\color{black!25},
  breaklines=true, breakatwhitespace=true,
  columns=fullflexible, keepspaces=true, showstringspaces=false,
  tabsize=4, xleftmargin=0.4em, xrightmargin=0.3em,
  upquote=true, escapeinside={(*@}{@*)},
  aboveskip=0.3em, belowskip=0.3em,
}
\lstdefinestyle{eal/plain}{style=eal/python, language=}
\lstdefinestyle{prompt}{%
  basicstyle=\ttfamily\scriptsize,
  backgroundcolor=\color{white},
  frame=single,
  rulecolor=\color{black},
  framesep=4pt,
  breaklines=true,
  breakatwhitespace=false,
  columns=fullflexible,
  keepspaces=true,
  showstringspaces=false,
  upquote=true,
  xleftmargin=4pt,
  xrightmargin=4pt,
  literate={→}{{$\to$}}1 {—}{{---}}1 {–}{{--}}1,
}

\newcommand{\ealbar}[1]{\textcolor{#1}{\rule[-0.15ex]{1.4pt}{0.9em}}}
\newcommand{\ealmod}{\ealbar{ealEdit}}
\newcommand{\ealedi}{\ealbar{ealEditable}}
\newcommand{\ealun}{\hphantom{\ealbar{ealEdit}}}
\newcommand{\eallno}[1]{\textcolor{ealNum}{\tiny#1}}
\newcommand{\elnM}[1]{\ealmod\ \eallno{#1}}
\newcommand{\elnE}[1]{\ealedi\ \eallno{#1}}
\newcommand{\elnU}[1]{\ealun\ \eallno{#1}}
\newcommand{\ealelide}[1]{\textcolor{ealNum}{\tiny\itshape\ \ \ \ ... #1 lines elided ...}}
\newcommand{\ealregion}[1]{\textcolor{ealNum}{\tiny\itshape\ \ \ \ -- #1 --}}

\newcommand{\hcsetcommentcolors}[1]{%
  \def\hcbg{defaultTint}\def\hcrule{defaultRule}%
  \IfSubStr{#1}{openai}{\def\hcbg{openaiTint}\def\hcrule{openaiRule}}{}%
  \IfSubStr{#1}{gpt}{\def\hcbg{openaiTint}\def\hcrule{openaiRule}}{}%
  \IfSubStr{#1}{claude}{\def\hcbg{anthropicTint}\def\hcrule{anthropicRule}}{}%
  \IfSubStr{#1}{anthropic}{\def\hcbg{anthropicTint}\def\hcrule{anthropicRule}}{}%
  \IfSubStr{#1}{gemini}{\def\hcbg{googleTint}\def\hcrule{googleRule}}{}%
  \IfSubStr{#1}{google}{\def\hcbg{googleTint}\def\hcrule{googleRule}}{}%
  \IfSubStr{#1}{deepseek}{\def\hcbg{deepseekTint}\def\hcrule{deepseekRule}}{}%
  \IfSubStr{#1}{qwen}{\def\hcbg{qwenTint}\def\hcrule{qwenRule}}{}%
}
\newcommand{\humancommentbox}[3]{%
  \hcsetcommentcolors{#2}%
  \begin{mdframed}[
    leftline=true, topline=false, bottomline=false, rightline=false,
    linecolor=\hcrule, linewidth=0.5pt,
    backgroundcolor=\hcbg,
    innerleftmargin=6pt, innerrightmargin=4pt,
    innertopmargin=3pt, innerbottommargin=3pt,
    skipabove=4pt, skipbelow=4pt
  ]%
  \itshape\small\textit{Expert Assessment.}~#3%
  \end{mdframed}%
}
\newcommand{\humancomment}[3]{%
  \par\smallskip\humancommentbox{#1}{#2}{#3}\par\smallskip
}

\definecolor{takeawayBg}{HTML}{EEF6FF}
\definecolor{takeawayRule}{HTML}{2F6FD6}
\newenvironment{takeawaybox}{%
  \par\smallskip
  \begin{mdframed}[
    linecolor=takeawayBg, linewidth=0pt, roundcorner=5pt,
    backgroundcolor=takeawayBg,
    innerleftmargin=7pt, innerrightmargin=7pt,
    innertopmargin=5pt, innerbottommargin=5pt,
    skipabove=5pt, skipbelow=6pt
  ]%
  \small\itshape\raggedright\noindent\textcolor{takeawayRule}{\textbf{Takeaway.}}\quad
}{%
  \end{mdframed}%
  \par\smallskip
}
\newenvironment{takeawayitems}{%
  \par\vspace{-0.5\baselineskip}%
  \begin{enumerate}[label=(\alph*), leftmargin=1.6em, topsep=0pt, partopsep=0pt, itemsep=1pt, parsep=0pt]%
}{%
  \end{enumerate}%
}

\definecolor{designChoiceBg}{HTML}{FFF1F1}
\definecolor{designChoiceRule}{HTML}{C43B3B}
\newenvironment{designchoicebox}{%
  \par\smallskip
  \begin{mdframed}[
    linecolor=designChoiceBg, linewidth=0pt, roundcorner=5pt,
    backgroundcolor=designChoiceBg,
    innerleftmargin=7pt, innerrightmargin=7pt,
    innertopmargin=5pt, innerbottommargin=5pt,
    skipabove=5pt, skipbelow=6pt
  ]%
  \small\itshape\raggedright\noindent\textcolor{designChoiceRule}{\textbf{Design Choice.}}\quad
}{%
  \end{mdframed}%
  \par\smallskip
}

\title{\bench: A Holistic and Rigorous Assessment of AI Systems on Building Better AI}

\author{\textbf{Bohan Lyu}$^{\clubsuit}$\thanks{Core contributors}  \ \
        \textbf{Yucheng Yang}$^{\spadesuit *}$
        \textbf{Siqiao Huang}$^{\spadesuit *}$
        \textbf{Jiaru Zhang}$^{\heartsuit *}$
        \textbf{Qixin Xu}$^{\clubsuit *}$
        \textbf{Xinghan Li}$^{\mathsection *}$ \\
        \textbf{Xinyang Han}$^{\clubsuit *}$
        \textbf{Yicheng Zhang}$^{\spadesuit *}$
        \textbf{Huaqing Zhang}$^{\spadesuit *}$
        \textbf{Runhan Huang}$^\ddagger$
        \textbf{Kaicheng Yang}$^\mathparagraph$  \\
        \textbf{Zitao Chen}$^\spadesuit$
        \textbf{Wentao Guo}$^\diamondsuit$
        \textbf{Junlin Yang}$^\spadesuit$
        \textbf{Xinyue Ai}$^\dagger$
        \textbf{Wenhao Chai}$^\diamondsuit$
        \textbf{Yadi Cao}$^\parallel$ \\
        \textbf{Ziran Yang}$^\diamondsuit$
        \textbf{Kun Wang}$^\diamondsuit$
        \textbf{Dapeng Jiang}$^\spadesuit$
        \textbf{Huan-ang Gao}$^\spadesuit$
        \textbf{Shange Tang}$^\diamondsuit$ \\
        \textbf{Chengshuai Shi}$^\diamondsuit$
        \textbf{Simon S. Du}$^\mathsection$
        \textbf{Max Simchowitz}$^\circ$
        \textbf{Jiantao Jiao}$^\clubsuit$
        \textbf{Dawn Song}$^\clubsuit$
        \textbf{Chi Jin}$^\diamondsuit$
  \vspace{0.3em}     \\
  $^\clubsuit$UC Berkeley
  \;
  $^\diamondsuit$Princeton University
  \;
  $^\spadesuit$Tsinghua University
  \;
  $^\mathsection$University of Washington
  \\
  $^\heartsuit$Purdue University
  \quad
  $^\ddagger$Harvard University
  \quad
  $^\dagger$University of Pennsylvania
   \\
  $^\mathparagraph$Shanghai Jiao Tong University
  \quad
  $^\parallel$UC San Diego
  \quad
  $^\circ$Carnegie Mellon University
  \vspace{0.3em}     \\
  \texttt{bohan@berkeley.edu, yc-yang24@mails.tsinghua.edu.cn}\\
  \texttt{\{jiantao, dawnsong\}@cs.berkeley.edu, chij@princeton.edu}
  \vspace{-0.5cm}
}

\begin{document}

\maketitle

\begin{abstract}
  Modern AI progress has been driven by ML methods that are generalizable across settings and scalable to larger regimes. As large language models demonstrate advanced capabilities in reasoning, coding, and engineering tasks, it is increasingly important to understand whether they can discover such methods rather than only apply existing ones. We introduce \bench, a benchmark for evaluating whether AI systems can invent generalizable and scalable ML methods. \bench contains \numtasks tasks across 12 domains, each requiring an agent to improve one targeted component of an ML system or algorithm and demonstrate that the improvement generalizes across controlled settings and scales. We find that current agents remain far from reliably surpassing human-designed methods, and that engineering-style tuning is easier for them than genuine method invention. We further study the effects of test-time scaling, adaptive compute allocation, and context provision on agents' discovery performance, together with case studies of their behavior. Our analyses suggest that the bottleneck is not only in proposing new methods, but also in the scientific insight needed to plan, validate, and scale claims about them. More search, compute, or context alone does not remove this bottleneck. We build and maintain a community platform for cumulative and comparable iteration, and release the data and code at \url{https://mls-bench.com}.
\end{abstract}

\section{Introduction}

\begin{list}{}{\setlength{\leftmargin}{1em}\setlength{\rightmargin}{1em}}
\item {\itshape
``We want AI agents that can discover like we can, not which contain what we have discovered.''
\par}
\raggedleft
--- Richard S. Sutton, \emph{The Bitter Lesson}
\end{list}
\vspace{0.1em}

Large language models~(LLMs) have evolved from chatbots~\citep{brown2020language,openai2023gpt4,bai2023qwen,kimiteam2025kimi15,touvron2023llama2,team2023gemini,chiang2024chatbot} into agents that pursue long-horizon objectives~\citep{schick2023toolformer,yao2023react,zhou2023webarena,mialon2023gaia,openai2025o3o4systemcard,an2025qwen3,deepseekai2025r1,kimiteam2025kimik2,anthropic2025claude4systemcard, comanici2025gemini}, including software engineering~\citep{jimenez2023swebench,mundler2024swtbench, chen2026sweci, liang2026swe}, deep research~\citep{shao2025drtulu,schmidgall2025agentlab,tang2025airesearcher, wei2025browsecomp}, and mathematical theorem proving~\citep{hubert2025olympiad,lin2025goedelprover, chen2025seed, lin2025goedel, wang2025kimina}.
Attention has recently shifted to more frontier problems including open optimization tasks such as circle packing~\citep{novikov2025alphaevolve,yuksekgonul2026learning,wang2025thetaevolve,mang2025frontiercs, openevolve} and machine learning engineering, where agents compete on Kaggle-style tasks~\citep{huang2023mlagentbench,chan2024mlebench,qiang2025mledojo,chen2026autolab, zhang2024mleagent, yang2025reinforcement}.
%
However, even these more advanced settings still do not resemble how human researchers discover new general methods.

This mismatch comes from the target of evaluation. Most agent benchmarks reward \emph{engineering}: improving one fixed instance through data processing, tuning, debugging, and model selection. ML science asks for a method-level idea, such as a new architecture, objective, component, or optimizer, that can be validated beyond the setting that produced it~\citep{simon1996sciences,vincenti1990what,krenn2022scientific,he2016deepresidual,vaswani2017attention,ho2020denoisingdiffusion,mnih2015humanlevel,schulman2017ppo,radford2021learningtransferable,zhang2019rmsnorm,dao2022flashattention,kingma2014adam,jordan2024muon,liu2025muon}. The question is whether agents can create such methods, rather than merely improve performance on a fixed benchmark or dataset.

Existing benchmarks do not yet isolate this capability. ML-engineering benchmarks mix method choice with implementation and tuning~\citep{huang2023mlagentbench,chan2024mlebench,qiang2025mledojo,chen2026autolab,zhang2024mleagent,yang2025reinforcement}; end-to-end research benchmarks make attribution hard~\citep{chen2025mlrbench,starace2025paperbench}; and recent narrow discovery benchmarks remain tied to single components or subfields~\citep{vitvitskyi2026mining,rank2026posttrainbench,chen2026agent2rlbench,ouyang2025kernelbench}.

\begin{figure}[t]
\centering
\includegraphics[width=\textwidth]{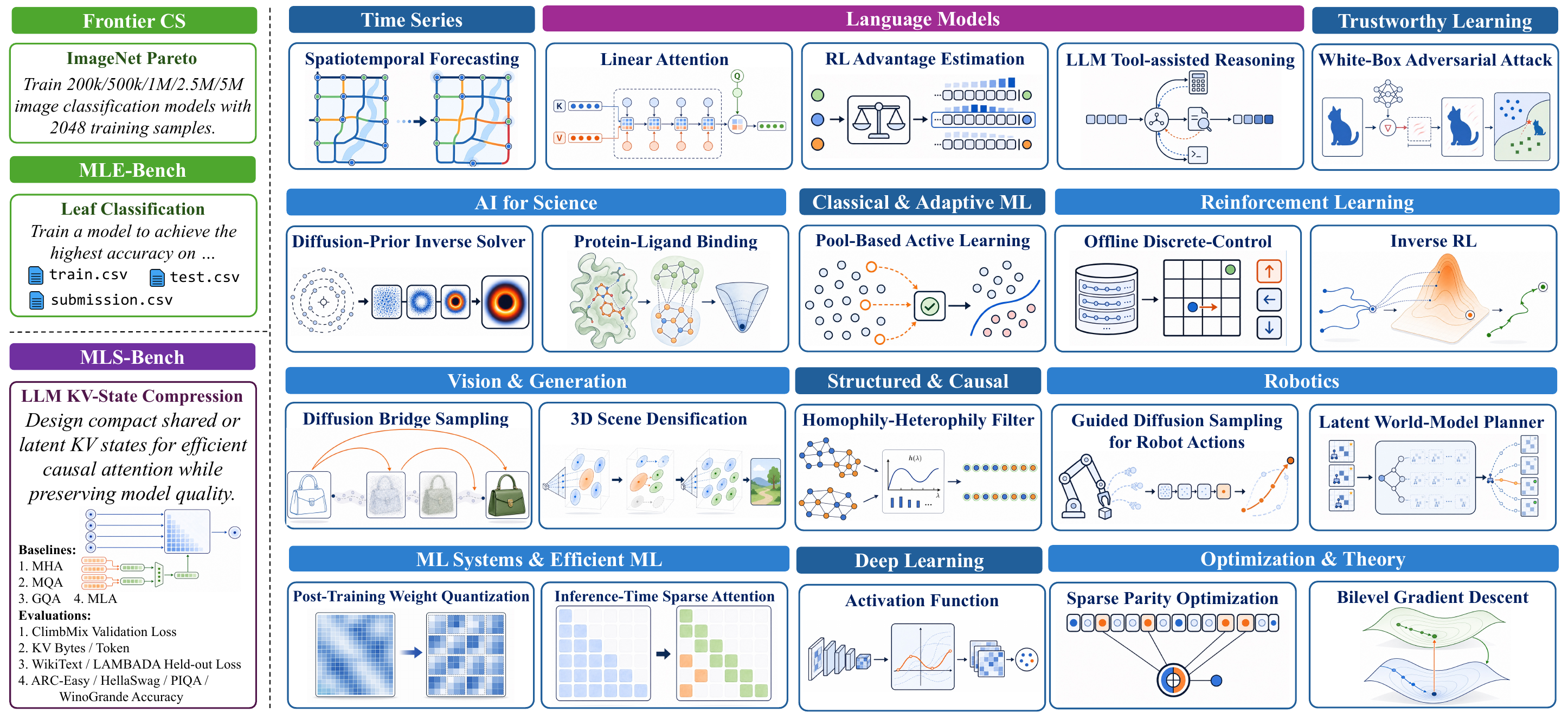}
\caption{\bench overview. Left: comparison of Frontier-CS, MLE-Bench, and \bench tasks. Right: 20 representative \bench tasks from \numtasks tasks across 12 domains.}
\label{fig:mls-main}
\vspace{-10pt}
\end{figure}

We introduce \textbf{\bench}~(\underline{ML} \underline{S}cience), a benchmark containing \numtasks tasks across 12 ML domains for evaluating whether AI systems can produce genuine, transferable ML method improvements. As shown in Figure~\ref{fig:mls-main}, each task asks an agent to improve a targeted component under controlled edit scopes, reproduced strong human baselines, and multiple evaluation settings. This design makes the submitted artifact attributable to the intended method rather than to evaluator changes, training-protocol hacks, or scale increases. We also curate \textbf{\bench-Lite}, a $30$-task challenging subset covering all 12 areas for rapid iteration and broader model tracking.

Our evaluation exposes a large method-discovery gap. Even with strong baselines in context and multiple opportunities to iterate, current frontier agents remain far from reliably matching human-designed methods inside the same scaffold. They are noticeably better at engineering-style tuning than at producing a new method that survives controlled validation, which makes \bench a demanding target for future foundation models and self-evolving frameworks.

We further study the influence of stronger inference-time support, more context, or greater freedom to experiment. These analyses show that current agents' limitation goes beyond proposing methods: they can search, tune, and recombine familiar ingredients, but struggle with the scientific judgment needed to form hypotheses, design experiments, allocate limited trials, and turn feedback into evidence for scalable claims. Human expert assessment also finds that genuinely new mechanisms are rare and often weakly justified.

We maintain \bench as a community benchmark with a growing leaderboard to guide the development of future foundation models and agent harnesses toward bootstrapping AI development.

\section{Related Work}

\paragraph{Automated scientific discovery.}
Computational methods have contributed to scientific discovery across diverse domains~\citep{udrescu2020ai,lam2023graphcast,appel1977solution,hubert2025olympiad,romera2024funsearch,jumper2021highly,watson2023rfdiffusion,swanson2025virtual,coley2019robotic,merchant2023gnome}, including computer science itself, spanning algorithms and systems~\citep{mirhoseini2021graph, chen2018learning,mankowitz2023alphadev}, and especially machine learning---including model architectures~\citep{zoph2017neural,liu2019darts}, training procedures~\citep{andrychowicz2016learning,finn2017model,hutter2019automl}, and data and loss design~\citep{cubuk2019autoaugment,gonzalez2020improved}.
More recently, LLMs have accelerated automated discovery across a broad spectrum: serving as collaborative scientific partners~\citep{gottweis2025coscientist,schmidgall2025agentlab}, optimizing specific algorithms and computational components~\citep{novikov2025alphaevolve,romera2024funsearch,ouyang2025kernelbench}, and driving fully autonomous research~\citep{lu2024aiscientist,yamada2025aiscientistv2,tang2025airesearcher}. A growing set of benchmarks evaluates these emerging capabilities~\citep{majumder2024discoverybench,chen2024scienceagentbench,liu2025researchbench,panigrahi2026heurekabench,bragg2025astabench}.

\paragraph{Self-evolving agents and evaluation.}
The paradigm of LLMs has evolved from single-turn question answering~\citep{brown2020language} toward agents that iterate over extended horizons~\citep{yao2023react,schick2023toolformer}.
Self-evolving systems iteratively refine solutions through evolutionary search~\citep{novikov2025alphaevolve,romera2024funsearch,chen2026avo,si2026executiongrounded}, open-ended self-improving loops~\citep{lange2025shinkaevolve,assumpcao2025codeevolve,panfilov2026claudini,karpathy2025autoresearch}, and test-time training~\citep{yuksekgonul2026learning,phan2025migrate,wang2025thetaevolve,surina2025algorithm,zuo2025ttrl}.
However, these systems have been demonstrated primarily on specific optimization problems, such as circle packing, contest-style algorithm search, kernel optimization, and activation-function search~\citep{novikov2025alphaevolve,yuksekgonul2026learning,wang2025thetaevolve,vitvitskyi2026mining}. Such settings are narrow in domain and do not capture whether a discovery is scalable and generalizable.

\paragraph{Benchmarking LLM agents for Coding and ML.}
Evaluation of LLMs on coding has progressed from code generation~\citep{chen2021evaluating,jain2024livecodebench,jimenez2023swebench,mundler2024swtbench,tang2023mlbench,chen2026sweci} toward code as a means to broader goals, including ML engineering~\citep{huang2023mlagentbench,chan2024mlebench,qiang2025mledojo,chen2026autolab} and open-ended scientific research~\citep{wijk2024rebench,nathani2025mlgym,garikaparthi2026researchgym,zhang2025mlrcbench,lupidi2026airsbench,wang2026firebench,mang2025frontiercs}.
While ML engineering evaluation is well-established, attempts to evaluate ML \emph{science}, i.e., whether AI can produce genuine method-level innovations, face limitations.
End-to-end research benchmarks~\citep{chen2025mlrbench} evaluate holistic workflows from ideation to manuscript, but their success criteria are broad, making it difficult to attribute individual method contribution.
Other benchmarks target specific ML components~\citep{rank2026posttrainbench,greenblatt2026aar, chen2026agent2rlbench,ouyang2025kernelbench,press2025algotune,imajuku2025ale, vitvitskyi2026mining}, leaving cross-domain generalization unmeasured.

\begin{figure}[!t]
\centering
\includegraphics[width=\textwidth]{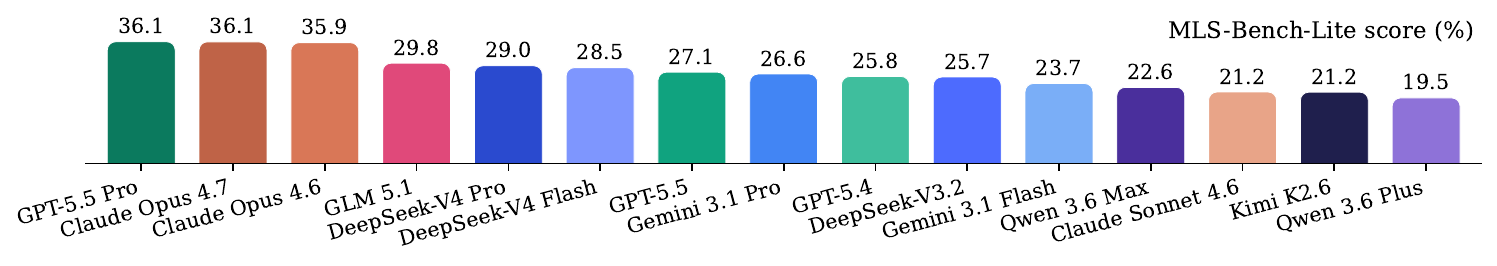}
\vspace{-15pt}
\caption{\bench-Lite Performance across 15 models.}
\label{fig:lite_intelligence}
\vspace{-10pt}
\end{figure}

\begin{table*}[!b]
\centering
\scriptsize
\setlength{\tabcolsep}{3pt}
\renewcommand{\arraystretch}{1.08}
\caption{Comparison with representative benchmarks. \cmark{} means explicit support, \mmark{} means partial or indirect support, and \xmark{} means absent. \textbf{Count}: source-reported primary evaluated units. \textbf{Scope \#}: number of source-reported coverage units. \textbf{New Method}: creation of a new scientific method rather than replication. \textbf{Generalize}: evaluation of whether the same method or artifact works across multiple settings. \textbf{Scalability}: scale-sensitive tasks or scalable evaluation design. \textbf{Control}: editable or submitted artifact restricted to the problem-relevant part under frozen evaluation.}
\label{tab:benchmark_comparison}
\resizebox{\textwidth}{!}{%
\begin{tabular}{llrrlccccl}
\toprule
\textbf{Benchmark} & \textbf{Setting} & \textbf{Count} & \textbf{\#} & \textbf{Scope} & \textbf{New Method} & \textbf{Generalize} & \textbf{Scalability} & \textbf{Control} & \textbf{Reference} \\
\midrule
ML-Bench~\citep{tang2023mlbench} & ML code & 169 & 18 & repos & \xmark & \xmark & \mmark & \mmark & Pass@K \\
MLAgentBench~\citep{huang2023mlagentbench} & ML experimentation & 13 & 4 & categories & \xmark & \xmark & \xmark & \xmark & baselines \\
MLE-bench~\citep{chan2024mlebench} & ML engineering & 75 & 15 & categories & \xmark & \xmark & \cmark & \cmark & medals \\
MLE-Dojo~\citep{qiang2025mledojo} & ML engineering & 200+ & 4 & domains & \xmark & \xmark & \cmark & \mmark & H-Rank \\
PostTrainBench~\citep{rank2026posttrainbench} & LLM post-training & 28 & 7 & evals & \xmark & \mmark & \cmark & \cmark & instruct \\
AutoResearch~\citep{karpathy2025autoresearch} & LLM training & 1 & 1 & setup & \mmark & \xmark & \mmark & \mmark & val\_bpb \\
SimpleTES~\citep{ye2026evaluationdriven} & Scientific discovery & 21 & 6 & domains & \cmark & \mmark & \cmark & \mmark & SOTA \\
MLGym~\citep{nathani2025mlgym} & ML experiments & 13 & 4 & domains & \mmark & \xmark & \mmark & \mmark & baselines \\
PaperBench~\citep{starace2025paperbench} & Paper replication & 20 & 1 & AI & \xmark & \xmark & \cmark & \mmark & rubric \\
MLR-Bench~\citep{chen2025mlrbench} & Research workflow & 201 & 9 & topics & \cmark & \xmark & \cmark & \xmark & judge \\
DiscoveryBench~\citep{majumder2024discoverybench} & Data discovery & 1167 & 6 & domains & \xmark & \xmark & \cmark & \cmark & facets \\
ScienceAgentBench~\citep{chen2024scienceagentbench} & Data workflow & 102 & 4 & fields & \xmark & \xmark & \mmark & \cmark & papers \\
AstaBench~\citep{bragg2025astabench} & Research assistance & 2400+ & 4 & areas & \mmark & \xmark & \mmark & \mmark & baselines \\
AutoLab~\citep{autolab2026benchmark} & Research loops & 23 & 3 & categories & \mmark & \xmark & \mmark & \cmark & medals \\
AIRS-Bench~\citep{lupidi2026airsbench} & ML SOTA tasks & 20 & 7 & categories & \xmark & \xmark & \cmark & \cmark & SOTA \\
FIRE-Bench~\citep{wang2026firebench} & Claim rediscovery & 30 & 1 & ML & \xmark & \mmark & \mmark & \mmark & claim-F1 \\
KernelBench~\citep{ouyang2025kernelbench} & Kernel optimization & 250 & 3 & levels & \xmark & \xmark & \mmark & \cmark & PyTorch \\
ALE-Bench~\citep{imajuku2025ale} & Algorithm optimization & 40 & 10 & genres & \mmark & \xmark & \cmark & \cmark & leaderboard \\
FrontierCS~\citep{mang2025frontiercs} & CS problems & 156 & 2 & tracks & \cmark & \mmark & \mmark & \cmark & expert \\
\midrule
\textbf{\bench} & \textbf{ML science} & \textbf{\numtasks} & \textbf{12} & \textbf{domains} & \cmark & \cmark & \cmark & \cmark & baselines \\
\bottomrule
\end{tabular}%
}
\end{table*}

\bench instead evaluates generalizable and scalable ML invention. Table~\ref{tab:benchmark_comparison} compares \bench with 19 representative benchmark datasets along these dimensions.

\section{\bench}
\label{sec:benchmark}

\bench evaluates whether AI systems can produce genuine, transferable algorithmic innovations. The benchmark is guided by the following principles:
(1)~\textbf{Holistic}: the benchmark covers the major areas the ML community actively pursues and their core research tasks.
(2)~\textbf{Atomic}: each task targets a single research question recognized by its research community as a coherent method-level contribution.
(3)~\textbf{Challenging}: every task includes strong human baselines recognized by the relevant community, including SOTA methods that we can reproduce.
(4)~\textbf{Generalizable}: solutions are evaluated across multiple settings.
(5)~\textbf{Reproducible}: all runs execute in controlled runtimes with pinned dependencies, fixed seeds, and locked package versions (Section~\ref{sec:scope}).
(6)~\textbf{Scientific innovation}: we enforce that performance gains come from the targeted method rather than from modifying the harness or shared training protocols, increasing model capacity, etc.
(7)~\textbf{Scalable}: evaluation scales are chosen to test whether methods remain effective when scaled up (Section~\ref{sec:rigor}).
(8)~\textbf{Unified scoring}: all metrics are normalized to a bounded scale based on baseline performance, enabling cross-task comparison (Section~\ref{sec:evaluation}).

\vspace{-5pt}
\begin{table}[!ht]
\centering
\small
\begin{minipage}[t]{0.70\textwidth}
\centering
\captionof{table}{Task distribution and representative topics across 12 domains.}
\label{tab:task_areas}
\vspace{0pt}
\resizebox{\textwidth}{!}{%
\begin{tabular}{lcl}
\toprule
\textbf{Area} & \textbf{Tasks} & \textbf{Topics} \\
\midrule
Language Models & 18 & Pretraining, Reasoning RL, Agents, Diffusion LMs \\
Classical \& Adaptive Learning & 14 & Few-Shot/Meta, Active/Continual/Federated, Calibration \\
Reinforcement Learning & 13 & Offline/Online RL, Meta/Multi-Agent RL, Inverse/Safe RL \\
Optimization \& Theory & 13 & Optimizers, Search/NAS/HPO, Bandits/Bounds \\
Robotics & 12 & World Models, Diffusion Policies, Imitation, Control \\
Vision \& Generation & 11 & 3D Vision, Diffusion, VAE/Flow, Image Generation \\
Deep Learning & 11 & Architectures, Losses, Augmentation, Normalization \\
ML Systems \& Efficient ML & 10 & KV Cache, Quantization, Kernels, Sparse Attention \\
AI for Science & 10 & Protein, Molecules, Climate/Weather, Inverse Problems \\
Time Series \& Forecasting & 10 & Forecasting, Imputation/Anomaly, Traffic, Quant Finance \\
Structured \& Causal Reasoning & 10 & Causal Discovery, Treatment Effects, Graph Learning \\
Trustworthy Learning & 8 & Attacks, Robust Training, Privacy, Unlearning \\
\bottomrule
\end{tabular}%
}
\end{minipage}\hfill
\begin{minipage}[t]{0.28\textwidth}
\centering
\vspace{0pt}
\includegraphics[width=\textwidth]{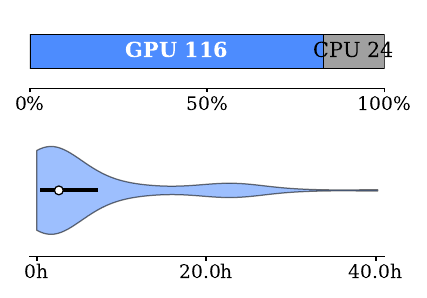}
\captionof{figure}{Compute profile: GPU vs.\ CPU task ratio and the distribution of GPU-hours per experiment.}
\label{fig:compute}
\end{minipage}
\vspace{-10pt}
\end{table}

\subsection{Overview}
\label{sec:scope}

\paragraph{Task Scope.}
\bench covers \numtasks tasks across 12 research areas; Table~\ref{tab:task_areas} lists the number of tasks and representative topics in each area. The tasks are built around community-recognized ML-science questions and turn them into executable, controlled, and comparable evaluations. Figure~\ref{fig:compute} shows the GPU/CPU task split and the distribution of H100 GPU-hours per experiment.

\paragraph{\bench-Lite.}
For convenient iteration and broader model tracking, we also curate \textbf{\bench-Lite}, a $30$-task subset covering all 12 domains. It keeps the central community-recognized questions in each area while remaining challenging. \bench-Lite's full list is given in Appendix~\ref{app:lite_tasks}. Running all \bench tasks requires $704.7$ H100-hours, while \bench-Lite requires only $99.2$ H100-hours, roughly one day on four H100 GPUs.

\paragraph{Task structure.}
A task specifies a research problem in executable form. It is defined by (i) a \textit{research question} that describes the research problem, its background and target; (ii) underlying \textit{codebase} with designated \textit{editable scopes} that constrain the regions the agents are able to edit; (iii) at least 3 strong human \textit{baselines} including the SOTA ones that we can reproduce; (iv) at least 3 \textit{evaluation settings} that probe generalization across benchmarks, environments, or base-model scales; (v) a \textit{seeds policy} that requires multi-seed evaluation for tasks whose scores carry non-negligible variance; (vi) a \textit{score normalization} that aggregates all metrics across all settings into a single comparable task-level score; and (vii) a \textit{capacity budget} that caps the agent's model size relative to the baseline when the task includes the modification of model components. The detailed contents of each task are listed in Appendix~\ref{app:task_catalog}.


\paragraph{Infrastructure.}
To ensure stable reproduction across diverse environments, the evaluation framework is built on a unified backend that supports multiple runtimes, including Apptainer, Docker, Conda, and Harbor~\citep{Harbor_Framework}. At the start of each run, the agent receives the task description, action and test budgets, task-relevant codebase files, and complete baseline implementations. Agents interact through four tools: \texttt{edit} modifies allowed code, \texttt{test} runs our harness and returns training and visible-test metrics, \texttt{submit} selects a previous test result as final, and \texttt{undo} reverts edits. See Appendix~\ref{app:prompts_and_tools} for the full system prompt, initial-prompt template, and tool schemas.




\subsection{Evaluation Rigor}
\label{sec:rigor}

We use three strategies to ensure that \bench reflects genuine method invention rather than confounders: (1) we constrain the agent's search to the algorithmic component under study while keeping the editable scope expressive enough to admit legitimate new methods (Section~\ref{sec:isolate}); (2) we select evaluation scales that preserve scalability evidence under a feasible compute budget (Section~\ref{sec:scale}); and (3) we guard against contamination and plagiarism (Section~\ref{sec:contam}).

\subsubsection{Isolating the algorithmic axis}
\label{sec:isolate}

An agent can raise its score by inventing a better method, but also by rewriting the evaluation harness, exploiting hyperparameters shared across methods, or inflating model capacity. \bench mechanically closes the latter routes so that only method invention is rewarded.

\paragraph{Scoping the method.} The editable scope of each task is restricted to the component under study, e.g., an architecture block or a training objective, while the evaluation harness remains frozen. Within this scope, we further differentiate between two kinds of hyperparameters: training-protocol knobs shared across methods (epochs, batch size) are locked into protected ranges so that the agent and every baseline run under the same setup, while method-defining hyperparameters (e.g., learning-rate schedule for an optimizer task) remain editable as part of the method itself. Based on these design choices, any score gain is therefore attributable to the component the task requires to study.

\paragraph{Baseline-calibrated scaffolding.}
While a loose scope may create exploitable loopholes, a tight one may prevent the agent from expressing legitimate new methods. We resolve this with a criterion, \emph{baseline-calibrated scaffolding}, that has two interacting parts: (1) a scope-design rule, where the editable scope of each task is set to be exactly wide enough to implement every established strong method for the problem as an edit sequence, and no wider; and (2) a validity check, where every baseline re-implemented inside this scope must reproduce its published reference performance, otherwise the task setup is rejected and revised. The two parts interact: the scope rule proposes a candidate setup, the reproduction check certifies or refutes that the scaffold and harness faithfully realize the original problem, and a task enters \bench only when both hold. This mechanism also removes potential bias caused by framework mismatch in \bench's evaluation.


\paragraph{Capping model capacity.} For tasks whose editable scope includes model components, we enforce a capacity budget on the submitted model. The evaluator counts the trainable parameters in the agent's submission and rejects it when it exceeds the task's capacity ceiling, preventing agents from obtaining higher scores by scaling the model rather than improving the method.

\begin{designchoicebox}
We pin evaluation to the algorithmic axis. Baseline-calibrated scaffolding ensures that the editable scope is tight enough to close shortcut routes, while reproducing every strong baseline in the same codebase certifies that this scope remains correct and expressive enough for legitimate methods.
\end{designchoicebox}

\begin{figure}[!t]
\centering
\includegraphics[width=\textwidth]{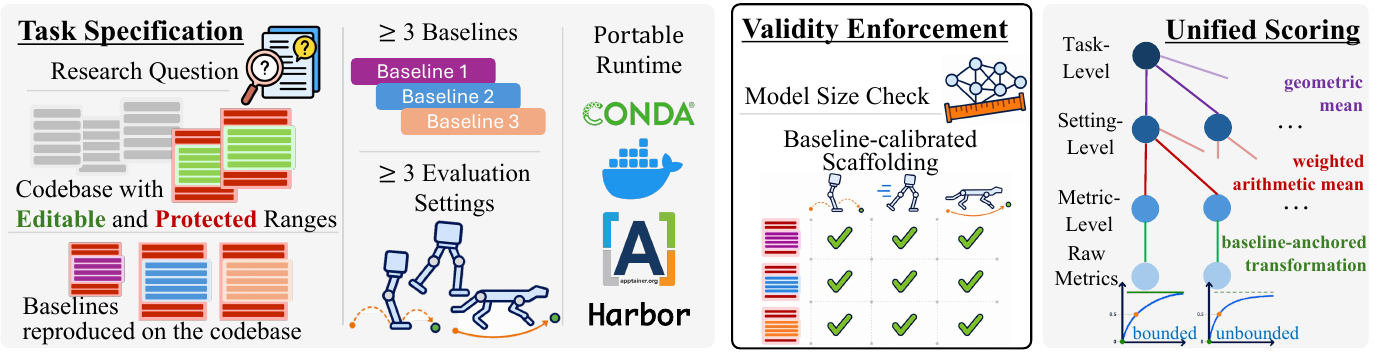}
\caption{\bench's design: task specification, validity enforcement, and unified scoring.}
\label{fig:mls-overview}
\vspace{-5pt}
\end{figure}

\subsubsection{Scalability and feasibility}
\label{sec:scale}

The scalability of a method is one of its most crucial features, and it is common that some methods help at small scale but fail to help at large scale~\citep{wen2025fantasticoptimizers, yang2022mup, everett2024scalingexponents, kaddour2023nogain, choshen2024misfitting}. However, computational feasibility is equally critical for benchmark design because excessive evaluation cost limits adoption and reduces the benchmark's utility as an iteration signal for method development. Below we outline how \bench reasons about and navigates this tension.

\paragraph{Relativity of scale.} Scale is inherently relative: for any evaluation scale, a larger one can lie beyond it, and the scaling behavior characterized at one regime can be revised when experiments push to larger ones. Qualitative phenomena such as emergence have been reported beyond previously studied scales~\citep{wei2022emergent}, though even their characterization is actively revised~\citep{schaeffer2023mirage}; and compute-optimal laws themselves have been re-derived~\citep{kaplan2020scalinglaws, hoffmann2022chinchilla} while single power-law fits break in new regimes~\citep{caballero2023broken}. The design problem is therefore not which exact scale to evaluate at, but how to preserve the strongest evidence for scalability compatible with a feasible compute budget.

\paragraph{Principled scale selection.} Our principle is that any setting must reproduce the published ranking of the existing baselines. This keeps the reduced task aligned with the original method-level comparison, so gains over baselines remain evidence of scalability rather than artifacts of an arbitrary small proxy. We keep native scales when feasible; otherwise, we reduce scale as little as possible to make evaluation tractable, while requiring the reduced setting to pass this ordering check.

\begin{designchoicebox}
By calibrating compute to preserve baseline ordering, \bench makes feasible evaluation settings informative about method-level scalability.
\end{designchoicebox}

\subsubsection{Contamination controls}
\label{sec:contam}

To prevent agents from succeeding by recalling public solutions rather than by inventing new ones, \bench adopts two complementary safeguards. (1) Each task contains the strongest established method that we could reproduce as a baseline, therefore a solution that merely retrieves a known method is unlikely to beat it. (2) Web search is disabled during our main experiments.

\subsection{Evaluation Metrics}
\label{sec:evaluation}

Every task in \bench is evaluated across multiple \emph{settings}, and each setting reports one or more raw \emph{metrics}. We aggregate metric scores within each setting and then aggregate across settings, producing a single bounded task score that is comparable across tasks.


\paragraph{Per-metric normalization.}
For each metric, we apply a \emph{baseline-anchored transformation}: the worst baseline anchors $0$ and the best baseline anchors $0.5$ on the internal $[0,1]$ scale. Because raw metrics differ in direction and units, we write the oriented metric score as $s(x)=\mathrm{sign}\cdot\operatorname{transform}(x)$, where $\operatorname{transform}$ uses one of two baseline calibrations:
{\scriptsize
\begin{equation}
  \setlength{\arraycolsep}{2pt}
  \operatorname{transform}(x)=
  \begin{cases}
    \bigl((x-x_\mathrm{floor})/(x_\mathrm{bound}-x_\mathrm{floor})\bigr)^\gamma,
    & \gamma=\frac{\log 0.5}{\log\bigl((x_\mathrm{ref}-x_\mathrm{floor})/(x_\mathrm{bound}-x_\mathrm{floor})\bigr)},\ \text{if } x_\mathrm{bound}\text{ exists},\\
    2\sigma((x-x_\mathrm{floor})/\lambda)-1,
    & \lambda=(x_\mathrm{ref}-x_\mathrm{floor})/\log 3,\ \text{else}.
  \end{cases}
  \label{eq:metric_normalization}
\end{equation}
}Here $x_\mathrm{floor}$ and $x_\mathrm{ref}$ are the worst and best baselines after applying any metric-specific preprocessing. The parameters $\gamma$ and $\lambda$ are chosen so that $s(x_\mathrm{ref})=0.5$.

\paragraph{Aggregation across levels.}
Within a setting, the score is the weighted arithmetic mean of its metric scores, $S_\mathrm{setting} = \sum_i w_i s_i / \sum_i w_i$, where $w_i$ is a human-labeled weight. Across settings, we take the geometric mean, $S_\mathrm{task} = \bigl(\prod_k S_{\mathrm{setting},k}\bigr)^{1/K}$, so a method cannot compensate for failure on one generalization setting by exploiting another.

\begin{table}[t]
\centering
\small
\caption{Results of frontier models compared with \textit{Human SOTA} across 12 areas. Area scores are \textbf{arithmetic means} over tasks within each area; Avg column is the \textbf{arithmetic mean} over 12 areas.}
\label{tab:main}
\resizebox{\textwidth}{!}{%
\begin{tabular}{l|cccccccccccc|c}
\toprule
 & LM & Rob & V\&G & RL & Sys & Sci & Opt & CAL & DL & TS & SCR & TL & Avg \\
\midrule
Human SOTA & 43.7 & 41.7 & 41.3 & 35.2 & 45.6 & 43.3 & 41.3 & 41.3 & 41.3 & 42.8 & 47.6 & 45.6 & 42.6 \\
\midrule
\multicolumn{14}{c}{\textit{Vanilla}} \\
\midrule
Claude Opus 4.6 & 26.9 & 38.4 & 12.7 & 21.8 & 33.9 & 26.2 & 31.3 & 30.1 & 36.6 & 26.5 & 26.6 & 26.7 & 28.1 \\
GPT-5.4 & 24.1 & 11.7 & 7.4 & 10.6 & 20.3 & 13.8 & 32.4 & 25.1 & 29.6 & 10.9 & 15.8 & 31.8 & 19.5 \\
Gemini 3.1 Pro & 39.0 & 23.2 & 21.6 & 10.9 & 29.7 & 33.4 & 42.7 & 18.6 & 40.4 & 16.2 & 12.4 & 30.1 & 26.5 \\
DeepSeek-V3.2 & 35.3 & 11.0 & 5.8 & 11.0 & 28.2 & 4.2 & 25.1 & 11.4 & 31.5 & 16.1 & 21.6 & 16.4 & 18.1 \\
Qwen 3.6 Plus & 22.0 & 15.3 & 14.3 & 14.2 & 13.6 & 12.0 & 23.9 & 20.6 & 29.8 & 5.8 & 1.9 & 24.4 & 16.5 \\
\midrule
\multicolumn{14}{c}{\textit{Agent}} \\
\midrule
Claude Opus 4.6 & 35.0 & 45.6 & 31.2 & 28.2 & 41.9 & 28.2 & 50.1 & 36.6 & 38.5 & 29.8 & 39.9 & 27.6 & 36.0 \\
GPT-5.4 & 34.7 & 21.1 & 21.4 & 12.4 & 22.5 & 23.1 & 45.4 & 34.6 & 31.6 & 19.5 & 28.4 & 39.4 & 27.8 \\
Gemini 3.1 Pro & 43.8 & 31.9 & 34.4 & 26.9 & 38.0 & 34.4 & 42.9 & 27.1 & 44.8 & 34.7 & 25.8 & 33.5 & 34.9 \\
DeepSeek-V3.2 & 41.1 & 30.5 & 18.2 & 12.4 & 28.3 & 20.7 & 27.7 & 18.9 & 33.7 & 25.7 & 32.5 & 26.2 & 26.3 \\
Qwen 3.6 Plus & 38.1 & 19.9 & 22.3 & 13.1 & 18.1 & 17.6 & 29.9 & 24.8 & 30.9 & 16.8 & 16.2 & 31.2 & 23.2 \\
\bottomrule
\end{tabular}
}
\vspace{-10pt}
\end{table}

\section{Experiments}
\label{sec:experiments}

\subsection{Setup}
\label{sec:setup}

\paragraph{Models.} We evaluate five frontier models on the full dataset: \texttt{Claude Opus 4.6}, \texttt{GPT-5.4}, \texttt{Gemini 3.1 Pro}, \texttt{DeepSeek-V3.2}, and \texttt{Qwen-3.6 Plus}. On \bench-Lite, we test ten more models: \texttt{Claude Opus 4.7}, \texttt{Claude Sonnet 4.6}, \texttt{GPT-5.5 Pro}, \texttt{GPT-5.5}, \texttt{Gemini 3.1 Flash Lite}, \texttt{DeepSeek-V4 Pro}, \texttt{DeepSeek-V4 Flash}, \texttt{Qwen-3.6 Max}, \texttt{Kimi K2.6}, and \texttt{GLM 5.1}. All models are run with high reasoning effort and a thinking-token budget of \(10{,}000\); we keep each provider's default sampling temperature. Web search is disabled, and seeds are fixed across all runs.

\paragraph{Settings.} In the main experiments, each agent is allowed at most \(20\) actions, including at most \(3\) \texttt{test} calls, before submitting a final proposal. We report two scores: \textit{Vanilla}, the first \texttt{test} result, and \textit{Agent}, the final submitted result. For the ten additional \bench-Lite models, each agent is allowed \(8\) actions and 1 \texttt{test} call, so we report only the \textit{Vanilla} result.
We report the best human baseline as \textit{Human SOTA}, scored with the same normalization as the agents. \textit{Human SOTA} is not fixed at 50 after aggregation: it can fall below 50 when different baselines win different metrics or settings, and can exceed 50 when a baseline reaches a metric's theoretical upper bound, which is mapped to 100. For high-variance tasks, all reported scores are multi-seed means, which give stable baseline orderings.
Experiments are executed and reproducible on H100 GPUs.
Some ablation and analysis experiments are evaluated on a subset where the property under study is well defined, and tasks within each subset are listed in Appendix~\ref{app:partial_experiment_tasks}.

\subsection{Main Results}
\label{sec:results}

Table~\ref{tab:main} reports per-area scores for the five models under \textit{Vanilla} and \textit{Agent} alongside \textit{Human SOTA}. Even with the full baseline implementations in context, frontier agents usually fail to match the strongest reproduced human methods when asked to implement a new algorithm. Iteration improves many submissions, but it mainly narrows the gap; it does not make current agents reliably competitive with methods already expressible inside the same scaffold. This unsaturated difficulty shows that \bench offers a durable target for the community to measure future progress.

\begin{takeawaybox}
Even when strong human baseline implementations are provided in context, current agents generally do not reach baseline-level performance. This shows both that \bench is a challenging target and that models are still weak at building on strong baselines to discover better methods.
\end{takeawaybox}




\subsection{Ablations}
\label{sec:ablations}

We study three questions behind the evaluation protocol: (1) whether agents are better at inventing new methods or tuning existing methods; (2) the effects of validity controls in our design; and (3) whether iterative refinement transfers across settings.

\begin{figure}[!t]
\centering
\begin{minipage}[b]{0.335\textwidth}
\centering
\includegraphics[width=\linewidth]{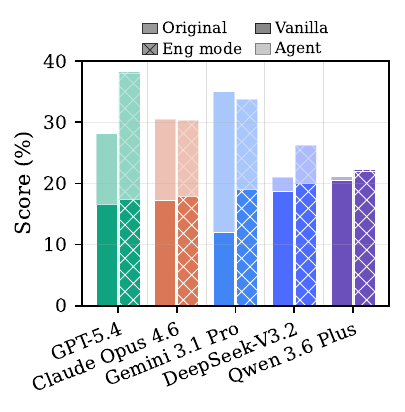}
\end{minipage}\hfill
\begin{minipage}[b]{0.325\textwidth}
\centering
\includegraphics[width=\linewidth]{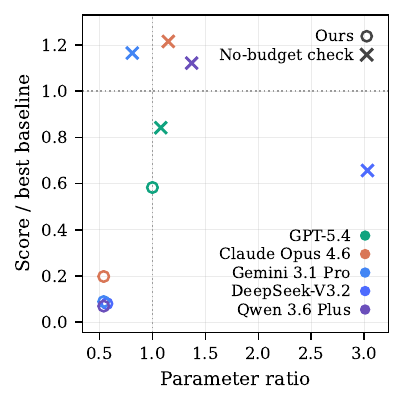}
\end{minipage}\hfill
\begin{minipage}[b]{0.33\textwidth}
\centering
\includegraphics[width=\linewidth]{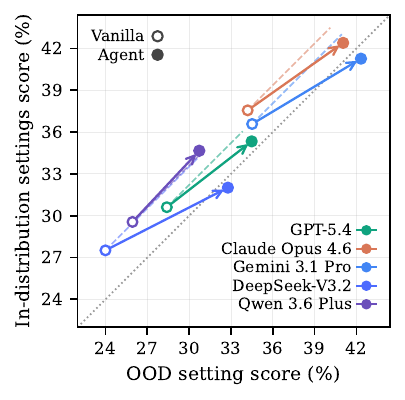}
\end{minipage}
\caption{Analysis on the evaluation protocol. \textbf{Left:} scientific-innovation prompt vs.\ engineering-optimization prompt. \textbf{Middle:} without the capacity-budget check, agents can obtain higher scores by increasing the submitted model's capacity. \textbf{Right:} in-distribution vs.\ OOD settings, from first proposal to final submission.}
\label{fig:exp_ablations_panel}
\end{figure}

\paragraph{Scientific innovation versus engineering optimization.}
Figure~\ref{fig:exp_ablations_panel}~(left) compares the scientific-innovation prompt with an engineering-optimization prompt. While \texttt{Claude Opus 4.6} and \texttt{Gemini 3.1 Pro} remain stable, other models gain especially after several iterations. The contrast shows that those agents, especially weaker models, are stronger at tuning parameters, applying known techniques, and polishing an existing implementation than at proposing and validating new methods.

\paragraph{Validity controls.}
We evaluate the capacity-budget control on computer vision and reinforcement learning tasks, where agents can adjust the model size. While removing the budget constraint does not consistently improve overall average performance, it opens a recurring shortcut: agents can increase the submitted model's capacity to obtain higher scores rather than improve the method itself. As illustrated by the specific cases in Figure~\ref{fig:exp_ablations_panel}~(middle), some over-capacity submissions surpass human SOTA. Our budget check rejects such submissions, forcing gains to come from method design rather than model-size increases. Additionally, we ablate the editable scope. Providing agents with a broader edit space does not enhance their effectiveness; rather, they frequently misuse this flexibility for off-target code modifications, which introduces implementation noise and degrades performance.

\paragraph{Domain generalization.}
For each task, we annotate one predefined out-of-distribution setting. We then track scores on the OOD setting versus the rest, from first proposal to final submission. Figure~\ref{fig:exp_ablations_panel}~(right) shows that, for most models, especially the strong ones, the initial in-distribution-vs-OOD gap shrinks by the final submission. This indicates that iterative refinement genuinely transfers across distributions, and \bench measures methods that travel rather than ones that merely fit the fixed settings.

\begin{takeawaybox}
\begin{takeawayitems}
\item Weaker models are better at engineering optimization than genuine method discovery.
\item Models can exploit increased capacity for higher scores, making the capacity-budget check necessary.
\item Strong models improve across settings while refining iteratively. Our multi-setting evaluation can probe whether a method truly generalizes.
\end{takeawayitems}
\end{takeawaybox}

\section{Analysis}
\label{sec:analysis}

Beyond the main evaluation, we study (1) test-time scaling, asking whether more tokens and compute budget can keep producing gains (Section~\ref{sec:test-time-scaling}); (2) adaptive compute allocation, placing agents in a realistic ML-science setting~(Section~\ref{sec:optimal-compute}); (3) context engineering, measuring how additional context changes model behavior (Section~\ref{sec:context-engineering}); and (4) human assessment, case studies, and error analysis, diagnosing where agents fail and what capabilities would be needed to improve (Section~\ref{sec:case-studies}).

\begin{figure}[!ht]
\centering
\includegraphics[width=\textwidth]{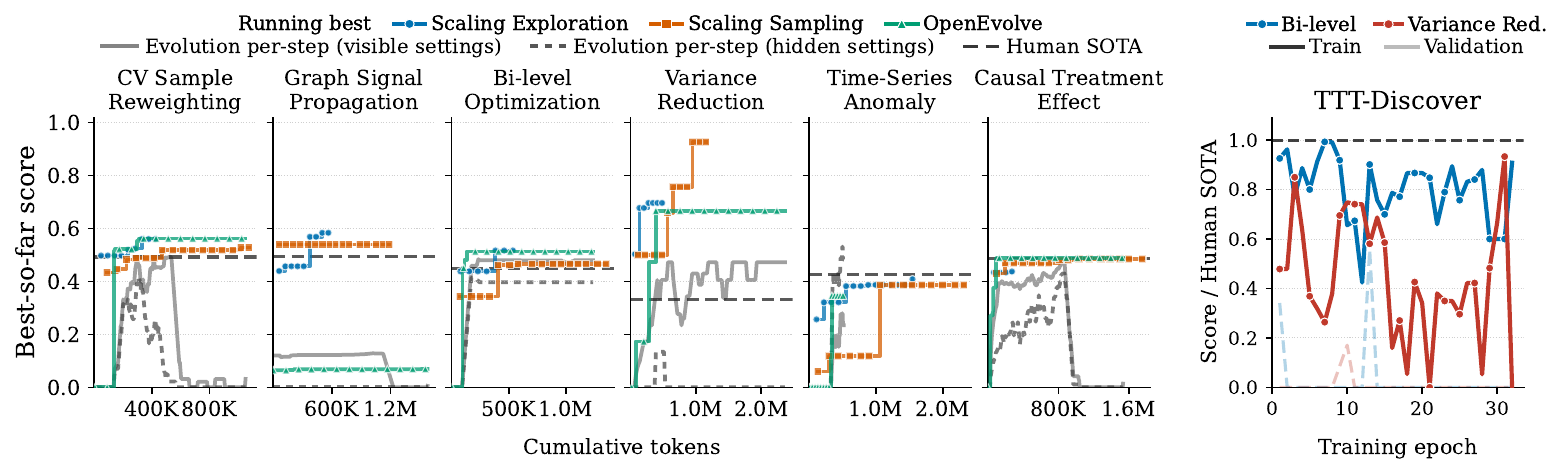}
\caption{Test-time scaling. \textbf{Left:} running-best score vs.\ cumulative tokens for the three inference-time setups. \textbf{Right:} TTT-Discover trained on two tasks, both overfitting visible settings.}
\label{fig:exp_test_time_scaling}
\vspace{-5pt}
\end{figure}

\subsection{Test-Time Scaling}
\label{sec:test-time-scaling}

\paragraph{Setup.} We evaluate four test-time scaling setups on low-latency \bench tasks: (1) \textbf{Scaling Sampling}, which samples many independent first proposals; (2) \textbf{Scaling Exploration}, which gives the agent more rounds of iterative experimentation; (3) \textbf{Test-time Evolution}, which runs an OpenEvolve population search with islands, mutation, selection, and execution feedback~\citep{openevolve}; and (4) \textbf{Test-time Training}, which updates the model from experimental feedback based on the TTT-Discover framework~\citep{yuksekgonul2026learning}. The first three inference-only setups use \texttt{Gemini 3.1 Pro}, while the test-time training setup uses \texttt{Qwen3.5-35B-A3B}. For the latter two setups, we make only two of the three settings visible to the model. Details for those setups are in Appendix~\ref{app:test_time_scaling_configs}.

\paragraph{Results.} Scaling helps on simpler tasks: \emph{scaling sampling} and \emph{scaling exploration} often raise the best score, but the gains quickly saturate. On the complex deep-learning task, more scale does not break the ceiling, let alone exceed Human SOTA. OpenEvolve and test-time training overfit the objective: their visible-setting scores are maintained or improved, while hidden-setting performance declines. These results suggest that test-time optimization needs sufficient setting diversity; otherwise agents improve observed cases rather than transfer.

\begin{takeawaybox}
\begin{takeawayitems}
\item Extra test-time compute can improve easier cases, but current scaling methods quickly hit a ceiling; on the harder tasks, that ceiling remains below the strongest provided human baseline.
\item Under partial feedback, scaling can optimize visible settings rather than the underlying method, raising observed scores while damaging hidden-setting performance.
\end{takeawayitems}
\end{takeawaybox}

\subsection{Verifier-Limited Compute Allocation}
\label{sec:optimal-compute}

The discovery systems above assume fast-verifier regimes. However, ML method discovery is often verifier-limited: proxy runs cannot by themselves establish scalable conclusions, and decisive evaluations are costly. ML scientists work under limited compute budgets. We therefore simulate the setting faced by ML scientists, especially LLM pretraining researchers, where an agent adaptively allocates limited compute across experiments and scales.

\begin{figure}[!h]
\centering
\includegraphics[width=\textwidth]{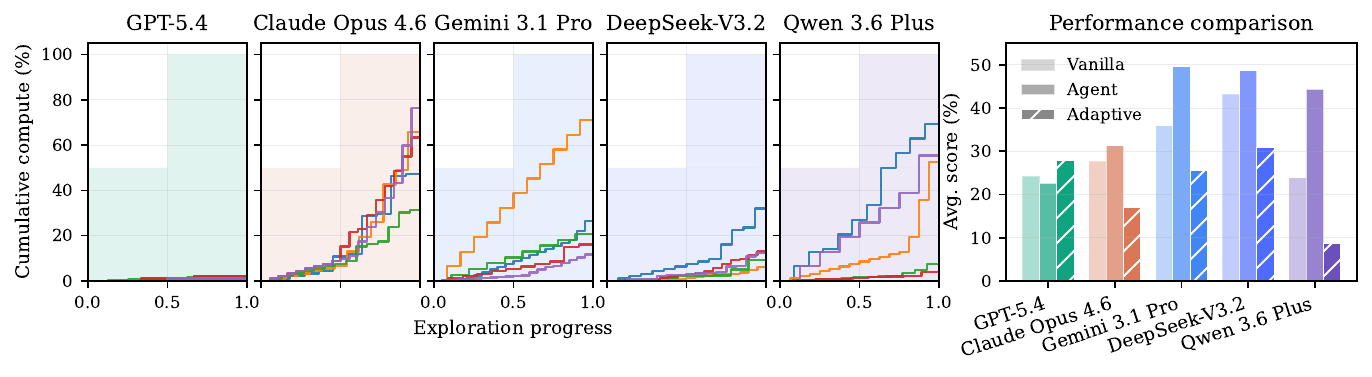}
\caption{Adaptive compute-allocation experiment. \textbf{Left:} cumulative compute budget consumed along the exploration trajectory for each of the five agents. \textbf{Right:} final-submission score.}
\vspace{-5pt}
\label{fig:exp2-allocation}
\end{figure}

\paragraph{Setup.} In the main experiment, for pretraining tasks, the standard \emph{Agent} protocol allows three full 345M-parameter runs. In this experiment, we convert the compute of the first two runs into an adaptive budget. Compute is measured as \(N{\cdot}D\), where \(N\) is model size and \(D\) is training tokens. During exploration, the agent chooses a proxy model size from \(\{51\mathrm{M},124\mathrm{M},199\mathrm{M},345\mathrm{M}\}\) and sets the token count for each \texttt{test} call. We allow at most 50 actions and 20 \texttt{test} calls. We run this experiment on five LLM pretraining tasks.

\paragraph{Results.} The adaptive protocol gives agents strictly more experimental choices than \textit{Vanilla} or the fixed \textit{Agent} setting, yet performance generally drops as shown in Figure~\ref{fig:exp2-allocation}. First, improvement is not monotonic in compute spent: \texttt{GPT-5.4} uses little budget yet is the only model that improves, while \texttt{Claude Opus 4.6} spends aggressively and still loses. \texttt{Gemini 3.1 Pro}, \texttt{DeepSeek-V3.2}, and \texttt{Qwen-3.6 Plus} follow roughly linear spending trajectories, whereas \texttt{Claude Opus 4.6} shows an accelerating, almost exponential pattern.

This result shows that the bottleneck extends beyond proposing a new method. When agents are given more autonomy to act as ML scientists, they often perform worse. The failure suggests that current models are not only weak at proposing new methods; they also lack the scientific judgment needed to build and validate evidence, a bottleneck that may be even more severe in realistic discovery workflows.

\begin{takeawaybox}
Scientific discovery in ML is a full workflow, not just a proposal step. In a realistic compute-limited setting, current agents struggle with the broader judgment needed to choose informative experiments, allocate scarce trials, and turn feedback into evidence for scalable claims.
\end{takeawaybox}

\subsection{Context Engineering and Reasoning Patterns}
\label{sec:context-engineering}

\begin{wrapfigure}{r}{0.40\textwidth}
\centering
\vspace{-3.5em}
\includegraphics[width=\linewidth]{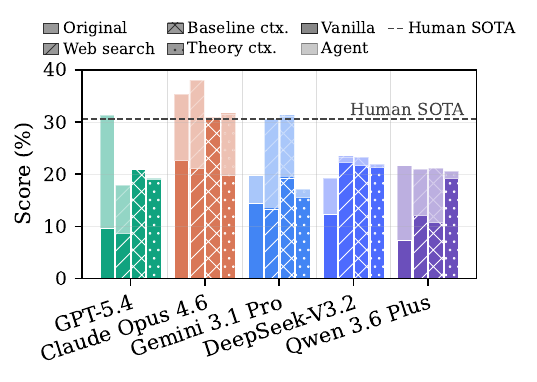}
\vspace{-2em}
\caption{Context engineering.}
\label{fig:context}
\vspace{-2em}
\end{wrapfigure}

We test whether additional context can help agents. We add three settings: (1) \textbf{Web search}, where we equip the agents with a strong search tool based on Tavily\footnote{\url{https://www.tavily.com/}}; (2) \textbf{Baseline ctx.}, which provides detailed derivations, key steps, and reasoning from the baseline papers; and (3) \textbf{Theory ctx.}, which provides background from relevant textbooks or theory-oriented literature.

Figure~\ref{fig:context} shows that these interventions provide generally modest gains. Whether context helps depends on the model's own capability: stronger models can extract some benefit. For all models, even when gains appear, they are easily matched by ordinary iterative refinement. This suggests that the bottleneck is not access to knowledge, but the ability to turn knowledge into testable hypotheses.

\begin{takeawaybox}
The bottleneck is not missing knowledge, but using it: turning context into testable hypotheses, relevant evidence, and implementations that survive evaluation.
\end{takeawaybox}

\subsection{Case Studies, Expert Assessment, and Error Analysis}
\label{sec:case-studies}

\begin{wrapfigure}{r}{0.40\textwidth}
\centering
\vspace{-2.5em}
\includegraphics[width=\linewidth]{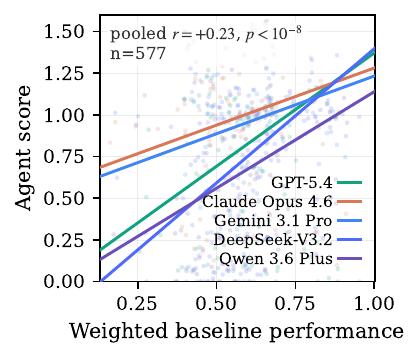}
\vspace{-1.5em}
\caption{Similarity-weighted baseline performance vs.\ agent performance.}
\label{fig:innovation}
\vspace{-2.0em}
\end{wrapfigure}

Expert assessment on agent submissions reveals dominant patterns: agents recombine ingredients drawn from the baselines they are shown and present the recombination as new, while truly novel components are rare and, when they appear, usually lack a stated reason they should help. Per-model style differs: \texttt{GPT-5.4} reaches for the most structurally different ideas but tends to overclaim novelty; \texttt{Claude Opus 4.6} is the most disciplined, favoring careful tuning over architectural rewrites and producing the cleanest implementations; \texttt{Gemini 3.1 Pro} attempts the boldest changes but rarely backs them with explicit hypotheses; \texttt{DeepSeek-V3.2} and \texttt{Qwen-3.6 Plus} default to hyperparameter search framed as method discovery.

We further probe this pattern with a statistic. For each run, we score code similarity to every baseline with 1/3-gram Jaccard and form $q{=}\sum_i s_i\beta_i/\!\sum_i s_i$, the code-similarity-weighted baseline score, where $\beta_i$ is the baseline score. To pool across tasks, we normalize both $q$ and the agent score by the task's best baseline. Figure~\ref{fig:innovation} plots these quantities. The pooled trend is significantly positive, and all five models follow the same direction. The per-model slope is significant for \texttt{DeepSeek-V3.2}, \texttt{Qwen-3.6 Plus}, and \texttt{GPT-5.4}, the three lowest-ranked models on the main leaderboard. This suggests a stratified failure mode: lower-performing models are more likely to mimic strong baselines than to explore new methods.

\begin{takeawaybox}
Current agents often produce combinations of the baselines they see. This pattern is especially clear for weaker models, whose performance tracks the baseline methods their code resembles.
\end{takeawaybox}

\section{Conclusion and Future Work}
\label{sec:conclusion}

We introduced \bench that evaluates whether AI systems can make reusable and scalable contributions to ML science. Across \numtasks tasks and 12 domains, current frontier agents remain far from reliably surpassing human-designed methods. \bench provides a common ground for measuring this gap as models and discovery methods advance.

Our analysis shows that the limitation of current agents lies not only in proposing better methods, but also in turning ideas into evidence. This evidence-building process requires a scientific discovery loop: deciding what to test, allocating limited trials, learning from feedback, and judging when a result supports a scalable claim. An agent can succeed on \bench only if it carries out this loop with sufficient rigor, not merely by proposing plausible methods.

Moreover, \bench captures a more realistic scientific discovery regime than prior self-evolving settings, which often rely on cheap verifiers for large-scale sampling. In ML science and many real scientific problems, verification is expensive, multi-stage, delayed, and only partially observable. \bench points to the next core challenge for foundation models and agent harnesses: \textbf{scalable discovery under unscalable verification}.

While \bench takes the first step and reveals limitations of current agents, ML science is too broad and fast-moving for one benchmark to exhaust. Future work should explore evaluation designs that remain rigorous while giving agents more freedom to pursue more open-ended questions.

\section*{Acknowledgment}

The authors thank Princeton AI Lab and Princeton Language and Intelligence (PLI) for their support of this work.
CJ acknowledges the support from NSF-OAC-2411299, NSF-IIS-2239297, Sloan Research Fellowship.

\bibliographystyle{plainnat}
\bibliography{verified}

@inproceedings{jimenez2023swebench,
  title     = {SWE-bench: Can Language Models Resolve Real-world Github Issues?},
  author    = {Carlos E. Jimenez and John Yang and Alexander Wettig and Shunyu Yao and Kexin Pei and Ofir Press and Karthik R. Narasimhan},
  year      = {2024},
  booktitle = {The Twelfth International Conference on Learning Representations}
}

@article{mundler2024swtbench,
  title={Swt-bench: Testing and validating real-world bug-fixes with code agents},
  author={M{\"u}ndler, Niels and M{\"u}ller, Mark N and He, Jingxuan and Vechev, Martin},
  journal={Advances in Neural Information Processing Systems},
  volume={37},
  pages={81857--81887},
  year={2024}
}

@article{tang2023mlbench,
  title     = {ML-Bench: Evaluating Large Language Models and Agents for Machine Learning Tasks on Repository-Level Code},
  author    = {Xiangru Tang and Yuliang Liu and Zefan Cai and Yanjun Shao and Junjie Lu and Yichi Zhang and Zexuan Deng and Helan Hu and Kaikai An and Ruijun Huang and Shuzheng Si and Sheng Chen and Haozhe Zhao and Liang Chen and Yan Wang and Tianyu Liu and Zhiwei Jiang and Baobao Chang and Fang, Yin and Yujia Qin and Wangchunshu Zhou and Yilun Zhao and Arman Cohan and Mark Gerstein},
  year      = {2023},
  journal   = {arXiv preprint arXiv:2311.09835}
}

@inproceedings{chan2024mlebench,
  title     = {MLE-bench: Evaluating Machine Learning Agents on Machine Learning Engineering},
  author    = {Jun Shern Chan and Neil Chowdhury and Oliver Jaffe and James Aung and Dane Sherburn and Evan Mays and Giulio Starace and Kevin Liu and Leon Maksin and Tejal Patwardhan and Aleksander Madry and Lilian Weng},
  year      = {2025},
  booktitle = {The Thirteenth International Conference on Learning Representations}
}

@article{qiang2025mledojo,
  title     = {MLE-Dojo: Interactive Environments for Empowering LLM Agents in Machine Learning Engineering},
  author    = {Rushi Qiang and Yuchen Zhuang and Yinghao Li and Dingu Sagar V. K and Rongzhi Zhang and Changhao Li and Ian Shu-Hei Wong and Sherry Yang and Percy Liang and Chao Zhang and Bo Dai},
  year      = {2025},
  journal   = {arXiv preprint arXiv:2505.07782},
}

@inproceedings{wijk2024rebench,
  title     = {RE-Bench: Evaluating Frontier AI R\&D Capabilities of Language Model Agents against Human Experts},
  author    = {Hjalmar Wijk and Tao Roa Lin and Joel Becker and Sami Jawhar and Neev Parikh and Thomas Broadley and Lawrence Chan and Michael Chen and Joshua Clymer and Jai Dhyani and Elena Ericheva and Katharyn Garcia and Brian Goodrich and Nikola Jurkovic and Megan Kinniment and Aron Lajko and Seraphina Nix and Lucas Jun Koba Sato and William Saunders and Maksym Taran and Ben West and Elizabeth Barnes},
  year      = {2025},
  booktitle = {Forty-second International Conference on Machine Learning}
}

@article{nathani2025mlgym,
  title     = {MLGym: A New Framework and Benchmark for Advancing AI Research Agents},
  author    = {Deepak Nathani and Lovish Madaan and Nicholas Roberts and Nikolay Bashlykov and Ajay Menon and Vincent Moens and Amar Budhiraja and Despoina Magka and Vladislav Vorotilov and Gaurav Chaurasia and Dieuwke Hupkes and Ricardo Silveira Cabral and Tatiana Shavrina and Jakob N. Foerster and Yoram Bachrach and William Yang Wang and Roberta Raileanu},
  year      = {2025},
  journal   = {arXiv preprint arXiv:2502.14499},
}

@inproceedings{zhang2025mlrcbench,
  title={MLAlgo-Bench: Can Machines Implement Machine Learning Algorithms?},
  author={Wang, Yunfei and Zhang, Yeqin and Wu, Yuyang and Lu, Liang and Le Nguyen, Phi and Wang, Xiaoliang and Nguyen, Cam-Tu},
  booktitle={Findings of the Association for Computational Linguistics: EMNLP 2025},
  pages={14298--14329},
  year={2025}
}

@article{chen2025mlrbench,
  title     = {MLR-Bench: Evaluating AI Agents on Open-Ended Machine Learning Research},
  author    = {Hui Chen and Miao Xiong and Yujie Lu and Wei Han and Ailin Deng and Yufei He and Jiaying Wu and Yibo Li and Yue Liu and Bryan Hooi},
  year      = {2025},
  journal   = {arXiv preprint arXiv:2505.19955},
}

@article{lupidi2026airsbench,
  title     = {AIRS-Bench: a Suite of Tasks for Frontier AI Research Science Agents},
  author    = {Alisia Maria Lupidi and Bhavul Gauri and Thomas Foster and Bassel Al Omari and Despoina Magka and Alberto Pepe and Alexis Audran-Reiss and Muna Aghamelu and Nicolas Mario Baldwin and Lucia Cipolina-Kun and Jean-Christophe Gagnon-Audet and Chee Hau Leow and Sandra Lefdal and Hossam Mossalam and Abhinav Moudgil and Saba Nazir and Emanuel Tewolde and Isabel Urrego and Jordi Armengol-Estapé and Amar Budhiraja and Gaurav Chaurasia and Abhishek Charnalia and Derek Dunfield and Karen Hambardzumyan and Daniel Izcovich and Martin Josifoski and Ishita Mediratta and Kelvin Niu and Parth Pathak and Michael Shvartsman and Edan Toledo and Anton Protopopov and Roberta Raileanu and Alexander H. Miller and Tatiana Shavrina and Jakob N. Foerster and Yoram Bachrach},
  year      = {2026},
  journal   = {arXiv preprint arXiv:2602.06855},
}

@article{garikaparthi2026researchgym,
  title     = {ResearchGym: Evaluating Language Model Agents on Real-World AI Research},
  author    = {Aniketh Garikaparthi and Manasi Patwardhan and Arman Cohan},
  year      = {2026},
  journal   = {arXiv preprint arXiv:2602.15112},
}

@article{wang2026firebench,
  title     = {FIRE-Bench: Evaluating Agents on the Rediscovery of Scientific Insights},
  author    = {Zhen Wang and Fan Bai and Zhongyan Luo and Jinyan Su and Kaiser Sun and Xinle Yu and Jieyuan Liu and Kun Zhou and Claire Cardie and Mark Dredze and Eric P. Xing and Zhiting Hu},
  year      = {2026},
  journal   = {arXiv preprint arXiv:2602.02905},
}

@article{mang2025frontiercs,
  title     = {FrontierCS: Evolving Challenges for Evolving Intelligence},
  author    = {Qiuyang Mang and Wenhao Chai and Zhifei Li and Huanzhi Mao and Shang Zhou and Alexander Du and Hanchen Li and Shu Liu and Edwin Chen and Yichuan Wang and Xieting Chu and Zerui Cheng and Yuan Xu and Tian Xia and Zirui Wang and Tianneng Shi and Jianzhu Yao and Yilong Zhao and Qizheng Zhang and Charlie Ruan and Zeyu Shen and Kaiyuan Liu and Runyuan He and Dong Xing and Zerui Li and Zirong Zeng and Yige Jiang and Lufeng Cheng and Ziyi Zhao and Youran Sun and Wesley Zheng and Meiyuwang Zhang and Ruyi Ji and Xuechang Tu and Zihan Zheng and Zexing Chen and Kangyang Zhou and Zhaozi Wang and Jingbang Chen and Aleksandra Korolova and Peter Henderson and Pramod Viswanath and Vijay Ganesh and Saining Xie and Zhuang Liu and Dawn Song and Sewon Min and Ion Stoica and Joseph E. Gonzalez and Jingbo Shang and Alvin Cheung},
  year      = {2025},
  journal   = {arXiv preprint arXiv:2512.15699},
}

@inproceedings{majumder2024discoverybench,
  title     = {DiscoveryBench: Towards Data-Driven Discovery with Large Language Models},
  author    = {Bodhisattwa Prasad Majumder and Harshit Surana and Dhruv Agarwal and Bhavana Dalvi Mishra and Abhijeetsingh Meena and Aryan Prakhar and Tirth Vora and Tushar Khot and Ashish Sabharwal and Peter Clark},
  year      = {2025},
  booktitle = {The Thirteenth International Conference on Learning Representations}
}

@inproceedings{chen2024scienceagentbench,
  title     = {ScienceAgentBench: Toward Rigorous Assessment of Language Agents for Data-Driven Scientific Discovery},
  author    = {Ziru Chen and Shijie Chen and Yuting Ning and Qianheng Zhang and Boshi Wang and Botao Yu and Yifei Li and Zeyi Liao and Chen Wei and Zitong Lu and Vishal Dey and Mingyi Xue and Frazier N. Baker and Benjamin Burns and Daniel Adu-Ampratwum and Xuhui Huang and Xia Ning and Song Gao and Yu Su and Huan Sun},
  year      = {2025},
  booktitle = {The Thirteenth International Conference on Learning Representations}
}

@article{liu2025researchbench,
  title     = {ResearchBench: Benchmarking LLMs in Scientific Discovery via Inspiration-Based Task Decomposition},
  author    = {Yujie Liu and Zonglin Yang and Tong Xie and Jinjie Ni and Ben Gao and Yuqiang Li and Shixiang Tang and Wanli Ouyang and Erik Cambria and Dongzhan Zhou},
  year      = {2025},
  journal   = {arXiv preprint arXiv:2503.21248},
}

@article{panigrahi2026heurekabench,
  title     = {HeurekaBench: A Benchmarking Framework for AI Co-scientist},
  author    = {Siba Smarak Panigrahi and Jovana Videnovic and Maria Brbic},
  year      = {2026},
  journal   = {arXiv preprint arXiv:2601.01678},
}

@inproceedings{schmidgall2025agentlab,
  title     = {Agent Laboratory: Using LLM Agents as Research Assistants},
  author    = {Samuel Schmidgall and Yusheng Su and Ze Wang and Ximeng Sun and Jialian Wu and Xiaodong Yu and Jiang Liu and Michael Moor and Zicheng Liu and Emad Barsoum},
  year      = {2025},
  booktitle = {Findings of the Association for Computational Linguistics},
  pages     = {5977--6043}
}

@article{lu2024aiscientist,
  title     = {The AI Scientist: Towards Fully Automated Open-Ended Scientific Discovery},
  author    = {Chris Lu and Cong Lu and Robert Tjarko Lange and Jakob N. Foerster and Jeff Clune and David Ha},
  year      = {2024},
  journal   = {arXiv preprint arXiv:2408.06292},
}

@article{yamada2025aiscientistv2,
  title     = {The AI Scientist-v2: Workshop-Level Automated Scientific Discovery via Agentic Tree Search},
  author    = {Yutaro Yamada and Robert Tjarko Lange and Cong Lu and Shengran Hu and Chris Lu and Jakob N. Foerster and Jeff Clune and David Ha},
  year      = {2025},
  journal   = {arXiv preprint arXiv:2504.08066},
}

@article{tang2025airesearcher,
  title     = {AI-Researcher: Autonomous Scientific Innovation},
  author    = {Jiabin Tang and Lianghao Xia and Zhonghang Li and Chao Huang},
  year      = {2025},
  journal   = {arXiv preprint arXiv:2505.18705},
}

@article{shao2025drtulu,
  title     = {DR Tulu: Reinforcement Learning with Evolving Rubrics for Deep Research},
  author    = {Rulin Shao and Akari Asai and Shannon Zejiang Shen and Hamish Ivison and Varsha Kishore and Jingming Zhuo and Xinran Zhao and Molly Park and Samuel G. Finlayson and David A. Sontag and Tyler Murray and Sewon Min and Pradeep Dasigi and Luca Soldaini and Faeze Brahman and Wen-tau Yih and Tongshuang Wu and Luke Zettlemoyer and Yoon Kim and Hannaneh Hajishirzi and Pang Wei Koh},
  year      = {2025},
  journal   = {arXiv preprint arXiv:2511.19399},
}

@article{gottweis2025coscientist,
  title     = {Towards an AI co-scientist},
  author    = {Juraj Gottweis and Wei-Hung Weng and Alexander N. Daryin and Tao Tu and Anil Palepu and Petar Sirkovic and Artiom Myaskovsky and Felix Weissenberger and Keran Rong and Ryutaro Tanno and Khaled Saab and Dan Popovici and Jacob Blum and Fan Zhang and Katherine Chou and Avinatan Hassidim and Burak Gokturk and Amin Vahdat and Pushmeet Kohli and Yossi Matias and Andrew Carroll and Kavita Kulkarni and Nenad Tomasev and Yuan Guan and Vikram Dhillon and Eeshit Dhaval Vaishnav and Byron Lee and Tiago R. D. Costa and José R. Penadés and Gary Peltz and Yunhan Xu and Annalisa Pawlosky and Alan Karthikesalingam and Vivek Natarajan},
  year      = {2025},
  journal   = {arXiv preprint arXiv:2502.18864},
}

@article{novikov2025alphaevolve,
  title     = {AlphaEvolve: A coding agent for scientific and algorithmic discovery},
  author    = {Alexander Novikov and Ngân Vu and Marvin Eisenberger and Emilien Dupont and Po-Sen Huang and Adam Zsolt Wagner and Sergey Shirobokov and Borislav Kozlovskii and Francisco J. R. Ruiz and Abbas Mehrabian and M. Pawan Kumar and Abigail See and Swarat Chaudhuri and George Holland and Alex Davies and Sebastian Nowozin and Pushmeet Kohli and Matej Balog},
  year      = {2025},
  journal   = {arXiv preprint arXiv:2506.13131},
}

@article{yuksekgonul2026learning,
  title     = {Learning to Discover at Test Time},
  author    = {Mert Yuksekgonul and D. M. Koceja and Xinhao Li and Federico Bianchi and Jed McCaleb and Xiaolong Wang and Jan Kautz and Yejin Choi and James Zou and Carlos Guestrin and Yu Sun},
  year      = {2026},
  journal   = {arXiv preprint arXiv:2601.16175}
}

@article{vitvitskyi2026mining,
  title     = {Mining Generalizable Activation Functions},
  author    = {Alex Vitvitskyi and Michael Boratko and Matej Grcic and Razvan Pascanu and Deep Shah and Petar Velickovic},
  year      = {2026},
  journal   = {arXiv preprint arXiv:2602.05688},
}

@article{udrescu2020ai,
  title     = {AI Feynman: A physics-inspired method for symbolic regression},
  author    = {Silviu‐Marian Udrescu and Max Tegmark},
  year      = {2020},
  journal   = {Science Advances},
  volume    = {6},
  number    = {16},
  pages     = {eaay2631--eaay2631}
}

@article{jumper2021highly,
  title     = {Highly accurate protein structure prediction with AlphaFold},
  author    = {John Jumper and Richard Evans and Alexander Pritzel and Tim Green and Michael Figurnov and Olaf Ronneberger and Kathryn Tunyasuvunakool and Russ Bates and Augustin Žídek and Anna Potapenko and Alex Bridgland and Clemens Meyer and Simon Köhl and Andrew J. Ballard and Andrew Cowie and Bernardino Romera‐Paredes and Stanislav Nikolov and Rishub Jain and Jonas Adler and Trevor Back and Stig Petersen and David Reiman and Ellen Clancy and Michał Zieliński and Martin Steinegger and Michalina Pacholska and Tamas Berghammer and Sebastian W. Bodenstein and David Silver and Oriol Vinyals and Andrew Senior and Koray Kavukcuoglu and Pushmeet Kohli and Demis Hassabis},
  year      = {2021},
  journal   = {Nature},
  volume    = {596},
  number    = {7873},
  pages     = {583--589}
}

@inproceedings{finn2017model,
  title     = {Model-Agnostic Meta-Learning for Fast Adaptation of Deep Networks},
  author    = {Chelsea Finn and Pieter Abbeel and Sergey Levine},
  year      = {2017},
  booktitle = {Proceedings of the 34th International Conference on Machine Learning},
  pages     = {1126--1135}
}

@article{andrychowicz2016learning,
  title={Learning to learn by gradient descent by gradient descent},
  author={Andrychowicz, Marcin and Denil, Misha and Gomez, Sergio and Hoffman, Matthew W and Pfau, David and Schaul, Tom and Shillingford, Brendan and De Freitas, Nando},
  journal={Advances in neural information processing systems},
  volume={29},
  year={2016}
}

@article{romera2024funsearch,
  title     = {Mathematical discoveries from program search with large language models},
  author    = {Bernardino Romera-Paredes and Mohammadamin Barekatain and Alexander Novikov and Matej Balog and M. Pawan Kumar and Emilien Dupont and Francisco J. R. Ruiz and Jordan S. Ellenberg and Pengming Wang and Omar Fawzi and Pushmeet Kohli and Alhussein Fawzi},
  year      = {2024},
  journal   = {Nature},
  volume    = {625},
  number    = {7995},
  pages     = {468--475}
}

@article{hubert2025olympiad,
  title={Olympiad-level formal mathematical reasoning with reinforcement learning},
  author={Hubert, Thomas and Mehta, Rishi and Sartran, Laurent and Horv{\'a}th, Mikl{\'o}s Z and {\v{Z}}u{\v{z}}i{\'c}, Goran and Wieser, Eric and Huang, Aja and Schrittwieser, Julian and Schroecker, Yannick and Masoom, Hussain and others},
  journal={Nature},
  pages={1--3},
  year={2025},
  publisher={Nature Publishing Group UK London}
}

@article{swanson2025virtual,
  title     = {The Virtual Lab: AI Agents Design New SARS-CoV-2 Nanobodies with Experimental Validation},
  author    = {Swanson, Kyle and Wu, Wesley and Bulaong, Nash L. and Pak, John E. and Zou, James},
  year      = {2024}
}

@article{phan2025migrate,
  title     = {MiGrATe: Mixed-Policy GRPO for Adaptation at Test-Time},
  author    = {Peter Phan and Dhruv Agarwal and Kavitha Srinivas and Horst Samulowitz and Pavan Kapanipathi and Andrew McCallum},
  year      = {2025},
  journal   = {arXiv preprint arXiv:2508.08641},
}

@article{wang2025thetaevolve,
  title     = {ThetaEvolve: Test-time Learning on Open Problems},
  author    = {Yiping Wang and Shao-Rong Su and Zhiyuan Zeng and Eva Xu and Liliang Ren and Xinyu Yang and Zeyi Huang and Xuehai He and Luyao Ma and Baolin Peng and Hao Cheng and Pengcheng He and Weizhu Chen and Shuohang Wang and Simon Shaolei Du and Yelong Shen},
  year      = {2025},
  journal   = {arXiv preprint arXiv:2511.23473},
}

@article{surina2025algorithm,
  title     = {Algorithm Discovery With LLMs: Evolutionary Search Meets Reinforcement Learning},
  author    = {Anja Surina and Amin Mansouri and Lars Quaedvlieg and Amal Seddas and Maryna Viazovska and Emmanuel Abbe and Caglar Gulcehre},
  year      = {2025},
  journal   = {arXiv preprint arXiv:2504.05108},
}

@article{zuo2025ttrl,
  title={Ttrl: Test-time reinforcement learning},
  author={Zuo, Yuxin and Zhang, Kaiyan and Sheng, Li and Qu, Shang and Cui, Ganqu and Zhu, Xuekai and Li, Haozhan and Zhang, Yuchen and Long, Xinwei and Hua, Ermo and others},
  journal={arXiv preprint arXiv:2504.16084},
  year={2025}
}

@article{chen2021evaluating,
  title     = {Evaluating Large Language Models Trained on Code},
  author    = {Mark Chen and Jerry Tworek and Heewoo Jun and Qiming Yuan and Henrique Pondé de Oliveira Pinto and Jared Kaplan and Harri Edwards and Yuri Burda and Nicholas Joseph and Greg Brockman and Alex Ray and Raul Puri and Gretchen Krueger and Michael Petrov and Heidy Khlaaf and Girish Sastry and Pamela Mishkin and Brooke Chan and Scott Gray and Nick Ryder and Mikhail Pavlov and Alethea Power and Lukasz Kaiser and Mohammad Bavarian and Clemens Winter and Philippe Tillet and Felipe Petroski Such and Dave Cummings and Matthias Plappert and Fotios Chantzis and Elizabeth Barnes and Ariel Herbert-Voss and William Hebgen Guss and Alex Nichol and Alex Paino and Nikolas Tezak and Jie Tang and Igor Babuschkin and Suchir Balaji and Shantanu Jain and William Saunders and Christopher Hesse and Andrew N. Carr and Jan Leike and Joshua Achiam and Vedant Misra and Evan Morikawa and Alec Radford and Matthew Knight and Miles Brundage and Mira Murati and Katie Mayer and Peter Welinder and Bob McGrew and Dario Amodei and Sam McCandlish and Ilya Sutskever and Wojciech Zaremba},
  year      = {2021},
  journal   = {arXiv preprint arXiv:2107.03374},
}

@inproceedings{jain2024livecodebench,
  title     = {LiveCodeBench: Holistic and Contamination Free Evaluation of Large Language Models for Code},
  author    = {Naman Jain and King Han and Alex Gu and Wen-Ding Li and Fanjia Yan and Tianjun Zhang and Sida Wang and Armando Solar-Lezama and Koushik Sen and Ion Stoica},
  year      = {2025},
  booktitle = {The Thirteenth International Conference on Learning Representations}
}

@article{imajuku2025ale,
  title     = {ALE-Bench: A Benchmark for Long-Horizon Objective-Driven Algorithm Engineering},
  author    = {Yuki Imajuku and Kohki Horie and Yoichi Iwata and Kensho Aoki and Naohiro Takahashi and Takuya Akiba},
  year      = {2025},
  journal   = {arXiv preprint arXiv:2506.09050},
}

@inproceedings{ouyang2025kernelbench,
  title     = {KernelBench: Can LLMs Write Efficient GPU Kernels?},
  author    = {Anne Ouyang and Simon Guo and Simran Arora and Alex L. Zhang and William Hu and Christopher Ré and Azalia Mirhoseini},
  year      = {2025},
  booktitle = {Forty-second International Conference on Machine Learning}
}

@article{mirhoseini2021graph,
  title={A graph placement methodology for fast chip design},
  author={Mirhoseini, Azalia and Goldie, Anna and Yazgan, Mustafa and Jiang, Joe Wenjie and Songhori, Ebrahim and Wang, Shen and Lee, Young-Joon and Johnson, Eric and Pathak, Omkar and Nova, Azade and others},
  journal={Nature},
  volume={594},
  number={7862},
  pages={207--212},
  year={2021},
  publisher={Nature Publishing Group UK London}
}

@article{chen2018learning,
  title     = {Learning to Optimize Tensor Programs},
  author    = {Tianqi Chen and Lianmin Zheng and Eddie Q. Yan and Ziheng Jiang and Thierry Moreau and Luis Ceze and Carlos Guestrin and Arvind Krishnamurthy},
  year      = {2018}
}

@article{appel1977solution,
  title     = {The Solution of the Four-Color-Map Problem},
  author    = {K. I. Appel and Wolfgang Haken},
  year      = {1977},
  journal   = {Scientific American},
  volume    = {237},
  number    = {4},
  pages     = {108--121}
}

@inproceedings{zoph2017neural,
  title     = {Neural Architecture Search with Reinforcement Learning},
  author    = {Barret Zoph and Quoc V. Le},
  year      = {2017},
  booktitle = {5th International Conference on Learning Representations}
}

@inproceedings{cubuk2019autoaugment,
  title     = {AutoAugment: Learning Augmentation Strategies From Data},
  author    = {Ekin D. Cubuk and Barret Zoph and Dandelion Mané and Vijay Vasudevan and Quoc V. Le},
  year      = {2019},
  booktitle = {{IEEE} Conference on Computer Vision and Pattern Recognition},
  pages     = {113--123}
}

@article{lam2023graphcast,
  title     = {Learning skillful medium-range global weather forecasting},
  author    = {Rémi Lam and Álvaro Sánchez‐González and Matthew Willson and Peter Wirnsberger and Meire Fortunato and Ferran Alet and Suman Ravuri and Timo Ewalds and Zach Eaton-Rosen and Weihua Hu and Alexander Merose and Stephan Hoyer and George Holland and Oriol Vinyals and Jacklynn Stott and Alexander Pritzel and Shakir Mohamed and Peter Battaglia},
  year      = {2023},
  journal   = {Science},
  volume    = {382},
  number    = {6677},
  pages     = {1416--1421}
}

@article{merchant2023gnome,
  title     = {Scaling deep learning for materials discovery},
  author    = {Amil Merchant and Simon L. Batzner and Samuel S. Schoenholz and Muratahan Aykol and Gowoon Cheon and Ekin Dogus Cubuk},
  year      = {2023},
  journal   = {Nature},
  volume    = {624},
  number    = {7990},
  pages     = {80--85}
}

@article{watson2023rfdiffusion,
  title     = {De novo design of protein structure and function with RFdiffusion},
  author    = {Joseph L. Watson and David Juergens and Nathaniel R. Bennett and Brian L. Trippe and Jason Yim and Helen E. Eisenach and Woody Ahern and Andrew J. Borst and Robert J. Ragotte and Lukas F. Milles and Basile I. M. Wicky and Nikita Hanikel and Samuel J. Pellock and Alexis Courbet and William Sheffler and Jue Wang and Preetham Venkatesh and Isaac Sappington and Susana Vázquez Torres and Anna Lauko and Valentin De Bortoli and Émile Mathieu and Sergey Ovchinnikov and Regina Barzilay and Tommi Jaakkola and Frank DiMaio and Minkyung Baek and David Baker},
  year      = {2023},
  journal   = {Nature},
  volume    = {620},
  number    = {7976},
  pages     = {1089--1100}
}

@article{coley2019robotic,
  title     = {A robotic platform for flow synthesis of organic compounds informed by AI planning},
  author    = {Connor W. Coley and Dale A. Thomas and Justin A. M. Lummiss and Jonathan N. Jaworski and C. Breen and Victor Schultz and Travis Hart and Joshua Fishman and Luke Rogers and Hanyu Gao and Robert W. Hicklin and Pieter Plehiers and Joshua Byington and John S. Piotti and William H. Green and A. John Hart and Timothy F. Jamison and Klavs F. Jensen},
  year      = {2019},
  journal   = {Science},
  volume    = {365},
  number    = {6453}
}

@article{mankowitz2023alphadev,
  title     = {Faster sorting algorithms discovered using deep reinforcement learning},
  author    = {Daniel J. Mankowitz and Andrea Michi and Anton Zhernov and Marco Gelmi and Marco Selvi and Cosmin Paduraru and Edouard Leurent and Shariq Iqbal and Jean-Baptiste Lespiau and Alex Ahern and Thomas Köppe and Kevin Millikin and Stephen Gaffney and Sophie Elster and Jackson Broshear and Chris Gamble and Kieran Milan and Robert Tung and Minjae Hwang and A. Taylan Cemgil and Mohammadamin Barekatain and Yujia Li and Amol Mandhane and Thomas Hubert and Julian Schrittwieser and Demis Hassabis and Pushmeet Kohli and Martin A. Riedmiller and Oriol Vinyals and David Silver},
  year      = {2023},
  journal   = {Nature},
  volume    = {618},
  number    = {7964},
  pages     = {257--263}
}

@inproceedings{liu2019darts,
  title     = {DARTS: Differentiable Architecture Search},
  author    = {Hanxiao Liu and Karen Simonyan and Yiming Yang},
  year      = {2019},
  booktitle = {7th International Conference on Learning Representations}
}

@article{brown2020language,
  title     = {Language Models are Few-Shot Learners},
  author    = {T. B. Brown and Benjamin Mann and Nick Ryder and Melanie Subbiah and Jared Kaplan and Prafulla Dhariwal and Arvind Neelakantan and Pranav Shyam and Girish Sastry and Amanda Askell and Sandhini Agarwal and Ariel Herbert-Voss and Gretchen Krueger and Tom Henighan and Rewon Child and Aditya Ramesh and Daniel M. Ziegler and Jeffrey Wu and Clemens Winter and Christopher Hesse and Mark Chen and Eric J. Sigler and Mateusz Litwin and Scott Gray and Benjamin Chess and Jack Clark and Christopher Berner and Sam McCandlish and Alec Radford and Ilya Sutskever and Dario Amodei},
  year      = {2020},
  journal   = {Neural Information Processing Systems},
  volume    = {33},
  pages     = {1877--1901}
}

@article{schick2023toolformer,
  title={Toolformer: Language models can teach themselves to use tools},
  author={Schick, Timo and Dwivedi-Yu, Jane and Dess{\`\i}, Roberto and Raileanu, Roberta and Lomeli, Maria and Hambro, Eric and Zettlemoyer, Luke and Cancedda, Nicola and Scialom, Thomas},
  journal={Advances in neural information processing systems},
  volume={36},
  pages={68539--68551},
  year={2023}
}

@inproceedings{huang2023mlagentbench,
  title     = {MLAgentBench: Evaluating Language Agents on Machine Learning Experimentation},
  author    = {Qian Huang and Jian Vora and Percy Liang and Jure Leskovec},
  year      = {2024},
  booktitle = {Forty-first International Conference on Machine Learning},
  pages     = {20271--20309}
}

@article{si2026executiongrounded,
  title   = {Towards Execution-Grounded Automated AI Research},
  author  = {Chenglei Si and Zitong Yang and Yejin Choi and Emmanuel Cand{\`e}s and Diyi Yang and Tatsunori Hashimoto},
  year    = {2026},
  journal = {arXiv preprint arXiv:2601.14525}
}

@article{press2025algotune,
  title     = {AlgoTune: Can Language Models Speed Up General-Purpose Numerical Programs?},
  author    = {Ori Press and Brandon Amos and Haoyu Zhao and Yikai Wu and Samuel K. Ainsworth and Dominik Krupke and Patrick Kidger and Touqir Sajed and Bartolomeo Stellato and Jisun Park and Nathanael Bosch and Eli Meril and Albert Steppi and Arman Zharmagambetov and Fangzhao Zhang and David Perez-Pineiro and Alberto Mercurio and Ni Zhan and Talor Abramovich and Kilian Lieret and Hanlin Zhang and Shirley Huang and Matthias Bethge and Ofir Press},
  year      = {2025},
  journal   = {arXiv preprint arXiv:2507.15887},
}

@article{chen2026agent2rlbench,
  title     = {Agent$^2$ RL-Bench: Can LLM Agents Engineer Agentic RL Post-Training?},
  author    = {Wanyi Chen and Xiao Yang and Xu Yang and Tianming Sha and Qizheng Li and Zhuo Wang and Bowen Xian and Fang Kong and Weiqing Liu and Jiang Bian},
  year      = {2026},
  journal   = {arXiv preprint arXiv:2604.10547}
}

@article{panfilov2026claudini,
  title     = {Claudini: Autoresearch Discovers State-of-the-Art Adversarial Attack Algorithms for LLMs},
  author    = {Alexander Panfilov and Peter Romov and Igor Shilov and Yves-Alexandre de Montjoye and Jonas Geiping and Maksym Andriushchenko},
  year      = {2026},
  journal   = {arXiv preprint arXiv:2603.24511},
}

@article{chen2026avo,
  title     = {AVO: Agentic Variation Operators for Autonomous Evolutionary Search},
  author    = {Terry Chen and Zhifan Ye and Bing Xu and Zihao Ye and Timmy Liu and Ali Hassani and Tianqi Chen and Andrew Kerr and Haicheng Wu and Yang Xu and Yu-Jung Chen and Hanfeng Chen and Aditya Kane and Ronny Krashinsky and Ming-Yu Liu and Vinod Grover and Luis Ceze and Roger A. Bringmann and John Tran and Wei Liu and Fung Xie and Michael Lightstone and Humphrey Shi},
  year      = {2026},
  journal   = {arXiv preprint arXiv:2603.24517},
}

@article{chen2026sweci,
  title     = {SWE-CI: Evaluating Agent Capabilities in Maintaining Codebases via Continuous Integration},
  author    = {Jialong Chen and Xander Xu and Hu Wei and Chuan Chen and Bing Zhao},
  year      = {2026},
  journal   = {arXiv preprint arXiv:2603.03823},
}

@article{rank2026posttrainbench,
  title     = {PostTrainBench: Can LLM Agents Automate LLM Post-Training?},
  author    = {Ben Rank and Hardik Bhatnagar and Ameya Prabhu and Shira Eisenberg and Karina Nguyen and Matthias Bethge and Maksym Andriushchenko},
  year      = {2026},
  journal   = {arXiv preprint arXiv:2603.08640},
}

@article{simon1996sciences,
  title={The science of design: Creating the artificial},
  author={Simon, Herbert A},
  journal={Design issues},
  pages={67--82},
  year={1988},
  publisher={JSTOR}
}

@book{vincenti1990what,
  title={What engineers know and how they know it},
  author={Vincenti, Walter G and others},
  volume={141},
  year={1990},
  publisher={Baltimore: Johns Hopkins University Press}
}

@article{krenn2022scientific,
  title     = {On scientific understanding with artificial intelligence},
  author    = {Mario Krenn and Robert Pollice and Si Yue Guo and Matteo Aldeghi and Alba Cervera-Lierta and Pascal Friederich and Gabriel dos Passos Gomes and Florian Häse and Adrián Jinich and AkshatKumar Nigam and Zhenpeng Yao and Alán Aspuru‐Guzik},
  year      = {2022},
  journal   = {Nature Reviews Physics},
  volume    = {4},
  number    = {12},
  pages     = {761--769}
}

@article{openai2023gpt4,
  title     = {GPT-4 Technical Report},
  author    = {OpenAI},
  year      = {2023},
  journal   = {arXiv preprint arXiv:2303.08774},
}

@article{bai2023qwen,
  title     = {Qwen Technical Report},
  author    = {Jinze Bai and Shuai Bai and Yunfei Chu and Zeyu Cui and Kai Dang and Xiaodong Deng and Yang Fan and Wenbin Ge and Yu Han and Fei Huang and Binyuan Hui and Luo Ji and Mei Li and Junyang Lin and Runji Lin and Dayiheng Liu and Gao Liu and Chengqiang Lu and Keming Lu and Jianxin Ma and Rui Men and Xingzhang Ren and Xuancheng Ren and Chuanqi Tan and Sinan Tan and Jianhong Tu and Peng Wang and Shijie Wang and Wei Wang and Shengguang Wu and Benfeng Xu and Jin Xu and An Yang and Hao Yang and Jian Yang and Shusheng Yang and Yang Yao and Bowen Yu and Hongyi Yuan and Zheng Yuan and Jianwei Zhang and Xingxuan Zhang and Yichang Zhang and Zhenru Zhang and Chang Zhou and Jingren Zhou and Xiaohuan Zhou and Tianhang Zhu},
  year      = {2023},
  journal   = {arXiv preprint arXiv:2309.16609},
}

@article{touvron2023llama2,
  title     = {Llama 2: Open Foundation and Fine-Tuned Chat Models},
  author    = {Hugo Touvron and Louis Martin and Kevin Stone and Peter Albert and Amjad Almahairi and Yasmine Babaei and Nikolay Bashlykov and Soumya Batra and Prajjwal Bhargava and Shruti Bhosale and Dan Bikel and Lukas Blecher and Cristian Canton-Ferrer and Moya Chen and Guillem Cucurull and David Esiobu and Jude Fernandes and Jeremy Fu and Wenyin Fu and Brian Fuller and Cynthia Gao and Vedanuj Goswami and Naman Goyal and Anthony Hartshorn and Saghar Hosseini and Rui Hou and Hakan Inan and Marcin Kardas and Viktor Kerkez and Madian Khabsa and Isabel Kloumann and Artem Korenev and Punit Singh Koura and Marie-Anne Lachaux and Thibaut Lavril and Jenya Lee and Diana Liskovich and Yinghai Lu and Yuning Mao and Xavier Martinet and Todor Mihaylov and Pushkar Mishra and Igor Molybog and Yixin Nie and Andrew Poulton and Jeremy Reizenstein and Rashi Rungta and Kalyan Saladi and Alan Schelten and Ruan Silva and Eric Michael Smith and Ranjan Subramanian and Xiaoqing Ellen Tan and Binh Tang and Ross Taylor and Adina Williams and Jian Xiang Kuan and Puxin Xu and Zheng Yan and Iliyan Zarov and Yuchen Zhang and Angela Fan and Melanie Kambadur and Sharan Narang and Aurélien Rodriguez and Robert Stojnic and Sergey Edunov and Thomas Scialom},
  year      = {2023},
  journal   = {arXiv preprint arXiv:2307.09288},
}

@article{kimiteam2025kimi15,
  title     = {Kimi k1.5: Scaling Reinforcement Learning with LLMs},
  author    = {Kimi Team},
  year      = {2025},
  journal   = {arXiv preprint arXiv:2501.12599},
}

@inproceedings{mialon2023gaia,
  title     = {GAIA: a benchmark for General AI Assistants},
  author    = {Grégoire Mialon and Clémentine Fourrier and Thomas Wolf and Yann LeCun and Thomas Scialom},
  year      = {2024},
  booktitle = {The Twelfth International Conference on Learning Representations}
}

@article{an2025qwen3,
  title     = {Qwen3 Technical Report},
  author    = {An Yang and Anfeng Li and Baosong Yang and Beichen Zhang and Binyuan Hui and Bo Zheng and B. X. Yu and Chang Gao and C. Huang and Chenxu Lv and Chujie Zheng and Dayiheng Liu and Fan Zhou and Fei Huang and H Feng and Hao Ge and Haoran Wei and Lin Huan and Jialong Tang and Jian Yang and Jianhong Tu and Jianwei Zhang and Jianxin Yang and Jiaxi Yang and Jing Zhou and Jingren Zhou and Junyang Lin and Kai Dang and Keqin Bao and Kexin Yang and Le Yu and Linyu Deng and Mei Li and Mingfeng Xue and Mingze Li and Pei Zhang and Peng Wang and Qin Zhu and Rui Men and Ruize Gao and Shixuan Liu and Shuang Luo and Tianhao Li and Tianyi Tang and Wenbiao Yin and Xingzhang Ren and Xinyu Wang and Xinyu Zhang and Xuancheng Ren and Fan Yang and Su Yang and Yichang Zhang and Yinger Zhang and Yu Wan and Yuqiong Liu and Zekun Wang and Zeyu Cui and Zhenru Zhang and Zhipeng Zhou and Zihan Qiu},
  year      = {2025},
  journal   = {arXiv preprint arXiv:2505.09388}
}

@article{lin2025goedelprover,
  title     = {Goedel-Prover: A Frontier Model for Open-Source Automated Theorem Proving},
  author    = {Yong Lin and Shange Tang and Bohan Lyu and Jiayun Wu and Hongzhou Lin and Kaiyu Yang and Jia Li and Mengzhou Xia and Danqi Chen and Sanjeev Arora and Chi Jin},
  year      = {2025},
  journal   = {arXiv preprint arXiv:2502.07640},
}

@article{chen2025seed,
  title     = {Seed-Prover 1.5: Mastering Undergraduate-Level Theorem Proving via Learning from Experience},
  author    = {Jiangjie Chen and Wenxiang Chen and Jiacheng Du and Jinyi Hu and Zhicheng Jiang and Allan Jie and Xiaoran Jin and Xing Jin and Chenggang Li and Wenlei Shi and Zhihong Wang and Mingxuan Wang and Chenrui Wei and Shufa Wei and Huajian Xin and Fan Yang and Weihao Gao and Zheng Yuan and Tianyang Zhan and Zeyu Zheng and Tianxi Zhou and Thomas Hanwen Zhu},
  year      = {2025},
  journal   = {arXiv preprint arXiv:2512.17260},
}

@article{liang2026swe,
  title     = {SWE-Next: Scalable Real-World Software Engineering Tasks for Agents},
  author    = {Jiarong Liang and Zhiheng Lyu and Zijie Liu and Xiangchao Chen and Ping Nie and Kai Zou and Wenhu Chen},
  year      = {2026},
  journal   = {arXiv preprint arXiv:2603.20691},
}

@article{wei2025browsecomp,
  title     = {BrowseComp: A Simple Yet Challenging Benchmark for Browsing Agents},
  author    = {Jason Wei and Zhiqing Sun and Spencer Papay and Scott McKinney and Jeffrey Han and Isa Fulford and Hyung Won Chung and Alex Tachard Passos and William Fedus and Amelia Glaese},
  year      = {2025},
  journal   = {arXiv preprint arXiv:2504.12516},
}

@article{team2023gemini,
  title     = {Gemini: A Family of Highly Capable Multimodal Models},
  author    = {Gemini Team},
  year      = {2023},
  journal   = {arXiv preprint arXiv:2312.11805},
}

@article{yang2025reinforcement,
  title     = {Reinforcement Learning for Machine Learning Engineering Agents},
  author    = {Sherry Yang and Joy He-Yueya and Percy Liang},
  year      = {2025},
  journal   = {arXiv preprint arXiv:2509.01684},
}

@inproceedings{starace2025paperbench,
  title     = {PaperBench: Evaluating AI's Ability to Replicate AI Research},
  author    = {Giulio Starace and Oliver Jaffe and Dane Sherburn and James Aung and Jun Shern Chan and Leon Maksin and Rachel Dias and Evan Mays and Benjamin Kinsella and Wyatt Thompson and Johannes Heidecke and Amelia Glaese and Tejal Patwardhan},
  year      = {2025},
  booktitle = {Forty-second International Conference on Machine Learning}
}

@article{bragg2025astabench,
  title     = {AstaBench: Rigorous Benchmarking of AI Agents with a Scientific Research Suite},
  author    = {Jonathan Bragg and Mike D'Arcy and Nishant Balepur and Dan Bareket and Bhavana Dalvi and Sergey Feldman and Dany Haddad and Jena D. Hwang and Peter A. Jansen and Varsha Kishore and Bodhisattwa Prasad Majumder and Aakanksha Naik and Sigal Rahamimov and Kyle Richardson and Amanpreet Singh and Harshit Surana and Aryeh Tiktinsky and Rosni Vasu and Guy Wiener and Chloe Anastasiades and Stefan Candra and Jason Dunkelberger and Dan Emery and Rob Evans and Malachi Hamada and Regan Huff and Rodney Kinney and Matt Latzke and Jaron Lochner and Ruben Lozano-Aguilera and Cecile Nguyen and Smita Rao and Amber Tanaka and Brooke Vlahos and Peter Clark and Doug Downey and Yoav Goldberg and Ashish Sabharwal and Daniel S. Weld},
  year      = {2025},
  journal   = {arXiv preprint arXiv:2510.21652},
}

@article{lange2025shinkaevolve,
  title     = {ShinkaEvolve: Towards Open-Ended And Sample-Efficient Program Evolution},
  author    = {Robert Tjarko Lange and Yuki Imajuku and Edoardo Cetin},
  year      = {2025},
  journal   = {arXiv preprint arXiv:2509.19349},
}

@article{hutter2019automl,
  title     = {Automated Machine Learning - Methods, Systems, Challenges},
  author    = {Frank Hutter and Lars Kotthoff and Joaquin Vanschoren},
  year      = {2019}
}

@inproceedings{gonzalez2020improved,
  title     = {Improved Training Speed, Accuracy, and Data Utilization Through Loss Function Optimization},
  author    = {Santiago Gonzalez and Risto Miikkulainen},
  year      = {2020},
  booktitle = {{IEEE} Congress on Evolutionary Computation},
  pages     = {1--8}
}

@inproceedings{yao2023react,
  title     = {ReAct: Synergizing Reasoning and Acting in Language Models},
  author    = {Shunyu Yao and Jeffrey Zhao and Dian Yu and Nan Du and Izhak Shafran and Karthik R. Narasimhan and Yuan Cao},
  year      = {2023},
  booktitle = {The Eleventh International Conference on Learning Representations}
}

@article{chen2026autolab,
  title     = {AutoLabs: Cognitive Multi-Agent Systems with Self-Correction for Autonomous Chemical Experimentation},
  author    = {Gihan Panapitiya and Emily Saldanha and Heather Job and Olivia Hess},
  year      = {2025},
  journal   = {arXiv preprint arXiv:2509.25651},
}

@article{lin2025goedel,
  title     = {Goedel-Prover-V2: Scaling Formal Theorem Proving with Scaffolded Data Synthesis and Self-Correction},
  author    = {Yong Lin and Shange Tang and Bohan Lyu and Ziran Yang and Jui-Hui Chung and Haoyu Zhao and Lai Jiang and Yihan Geng and Jiawei Ge and Jingruo Sun and Jiayun Wu and Jiri Gesi and Ximing Lu and David Acuna and Kaiyu Yang and Hongzhou Lin and Yejin Choi and Danqi Chen and Sanjeev Arora and Chi Jin},
  year      = {2025},
  journal   = {arXiv preprint arXiv:2508.03613},
}

@article{wang2025kimina,
  title     = {Kimina-Prover Preview: Towards Large Formal Reasoning Models with Reinforcement Learning},
  author    = {Haiming Wang and Mert Unsal and Xiaohan Lin and Mantas Baksys and Junqi Liu and Marco Dos Santos and Flood Sung and Marina Vinyes and Zhenzhe Ying and Zekai Zhu and Jianqiao Lu and Hugues de Saxcé and Bolton Bailey and Chendong Song and Chenjun Xiao and Dehao Zhang and Ebony Zhang and Frederick Pu and Han Zhu and Jiawei Liu and Jonas Bayer and Julien Michel and Longhui Yu and Léo Dreyfus-Schmidt and Lewis Tunstall and Luigi Pagani and Moreira Machado and Pauline Bourigault and Ran Wang and Stanislas Polu and Thibaut Barroyer and Wen-Ding Li and Yazhe Niu and Yann Fleureau and Yangyang Hu and Zhouliang Yu and Zihan Wang and Zhilin Yang and Zhengying Liu and Jia Li},
  year      = {2025},
  journal   = {arXiv preprint arXiv:2504.11354},
}

@inproceedings{chiang2024chatbot,
  title     = {Chatbot Arena: An Open Platform for Evaluating LLMs by Human Preference},
  author    = {Wei-Lin Chiang and Lianmin Zheng and Ying Sheng and Anastasios Nikolas Angelopoulos and Tianle Li and Dacheng Li and Banghua Zhu and Hao Zhang and Michael I. Jordan and Joseph E. Gonzalez and Ion Stoica},
  year      = {2024},
  booktitle = {Forty-first International Conference on Machine Learning},
  pages     = {8359--8388}
}

@inproceedings{zhou2023webarena,
  title     = {WebArena: A Realistic Web Environment for Building Autonomous Agents},
  author    = {Shuyan Zhou and Frank F. Xu and Hao Zhu and Xuhui Zhou and Robert Lo and Abishek Sridhar and Xianyi Cheng and Tianyue Ou and Yonatan Bisk and Daniel Fried and Uri Alon and Graham Neubig},
  year      = {2024},
  booktitle = {The Twelfth International Conference on Learning Representations}
}

@article{deepseekai2025r1,
  title     = {DeepSeek-R1 incentivizes reasoning in LLMs through reinforcement learning},
  author    = {Daya Guo and Dejian Yang and Haowei Zhang and Junxiao Song and Peiyi Wang and Qihao Zhu and Runxin Xu and Ruoyu Zhang and Shirong Ma and Xiao Bi and Xiaokang Zhang and Xingkai Yu and Yu Wu and Z. F. Wu and Zhibin Gou and Zhihong Shao and Zhuoshu Li and Ziyi Gao and Aixin Liu and Bing Xue and Bingxuan Wang and Bochao Wu and Bei Feng and Chengda Lu and Chenggang Zhao and Chengqi Deng and Chong Ruan and Damai Dai and Deli Chen and Dongjie Ji and Erhang Li and Fangyun Lin and Fucong Dai and Fuli Luo and Guangbo Hao and Guanting Chen and Guowei Li and Hao Zhang and Hanwei Xu and Honghui Ding and Huazuo Gao and Hui Qu and Hui Li and Jianzhong Guo and Jiashi Li and Jingchang Chen and Jingyang Yuan and Jinhao Tu and Junjie Qiu and Junlong Li and J. L. Cai and Jiaqi Ni and Jian Liang and Jin Chen and Kai Dong and Kai Hu and Kaichao You and Kaige Gao and Kang Guan and Kexin Huang and Kuai Yu and Lean Wang and Lecong Zhang and Liang Zhao and Litong Wang and Liyue Zhang and Lei Xu and Leyi Xia and Mingchuan Zhang and Minghua Zhang and Minghui Tang and Mingxu Zhou and Meng Li and Miaojun Wang and Mingming Li and Ning Tian and Panpan Huang and Peng Zhang and Qiancheng Wang and Qinyu Chen and Qiushi Du and Ruiqi Ge and Ruisong Zhang and Ruizhe Pan and Runji Wang and R. J. Chen and R. L. Jin and Ruyi Chen and Shanghao Lu and Shangyan Zhou and Shanhuang Chen and Shengfeng Ye and Shiyu Wang and Shuiping Yu and Shunfeng Zhou and Shuting Pan and S. S. Li and Shuang Zhou and Shaoqing Wu and Tao Yun and Tian Pei and Tianyu Sun and Tao Wang and Wangding Zeng and Wen Liu and Wenfeng Liang and Wenjun Gao and Wenqin Yu and Wentao Zhang and W. L. Xiao and Wei An and Xiaodong Liu and Xiaohan Wang and Xiaokang Chen and Xiaotao Nie and Xin Cheng and Xin Liu and Xin Xie and Xingchao Liu and Xinyu Yang and Xinyuan Li and Xuecheng Su and Xuheng Lin and X. Q. Li and Xiangyue Jin and Xiaojin Shen and Xiaosha Chen and Xiaowen Sun and Xiaoxiang Wang and Xinnan Song and Xinyi Zhou and Xianzu Wang and Xinxia Shan and Y. K. Li and Y. Q. Wang and Y. X. Wei and Yang Zhang and Yanhong Xu and Yao Li and Yao Zhao and Yaofeng Sun and Yaohui Wang and Yi Yu and Yichao Zhang and Yifan Shi and Yiliang Xiong and Ying He and Yishi Piao and Yisong Wang and Yixuan Tan and Yiyang Ma and Yiyuan Liu and Yongqiang Guo and Yuan Ou and Yuduan Wang and Yue Gong and Yuheng Zou and Yujia He and Yunfan Xiong and Yuxiang Luo and Yuxiang You and Yuxuan Liu and Yuyang Zhou and Y. X. Zhu and Yanping Huang and Yaohui Li and Yi Zheng and Yuchen Zhu and Yunxian Ma and Ying Tang and Yukun Zha and Yuting Yan and Z. Z. Ren and Zehui Ren and Zhangli Sha and Zhe Fu and Zhean Xu and Zhenda Xie and Zhengyan Zhang and Zhewen Hao and Zhicheng Ma and Zhigang Yan and Zhiyu Wu and Zihui Gu and Zijia Zhu and Zijun Liu and Zilin Li and Ziwei Xie and Ziyang Song and Zizheng Pan and Zhen Huang and Zhipeng Xu and Zhongyu Zhang and Zhen Zhang},
  year      = {2025},
  journal   = {Nature},
  volume    = {645},
  number    = {8081},
  pages     = {633--638}
}

@article{kimiteam2025kimik2,
  title     = {Kimi K2: Open Agentic Intelligence},
  author    = {Kimi Team},
  year      = {2025},
  journal   = {arXiv preprint arXiv:2507.20534},
}

@article{assumpcao2025codeevolve,
  title={Codeevolve: An open source evolutionary coding agent for algorithm discovery and optimization},
  author={Assump{\c{c}}{\~a}o, Henrique and Ferreira, Diego and Campos, Leandro and Murai, Fabricio},
  journal={arXiv preprint arXiv:2510.14150},
  year={2025}
}

@misc{greenblatt2026aar,
  title        = {Automated Alignment Researchers},
  author       = {Anthropic},
  year         = {2026},
  howpublished = {\url{https://www.anthropic.com/research/automated-alignment-researchers}}
}

@misc{karpathy2025autoresearch,
  title        = {autoresearch},
  author       = {Karpathy, Andrej},
  year         = {2025},
  howpublished = {\url{https://github.com/karpathy/autoresearch}}
}

@misc{openai2025o3o4systemcard,
  title        = {OpenAI o3 and o4-mini System Card},
  author       = {{OpenAI}},
  year         = {2025},
  howpublished = {\url{https://openai.com/index/o3-o4-mini-system-card/}},
  note         = {System card, published April 16, 2025}
}

@article{comanici2025gemini,
  title={Gemini 2.5: Pushing the frontier with advanced reasoning, multimodality, long context, and next generation agentic capabilities},
  author={Comanici, Gheorghe and Bieber, Eric and Schaekermann, Mike and Pasupat, Ice and Sachdeva, Noveen and Dhillon, Inderjit and Blistein, Marcel and Ram, Ori and Zhang, Dan and Rosen, Evan and others},
  journal={arXiv preprint arXiv:2507.06261},
  year={2025}
}

@software{openevolve,
  title = {OpenEvolve: an open-source evolutionary coding agent},
  author = {Asankhaya Sharma},
  year = {2025},
  publisher = {GitHub},
  url = {https://github.com/algorithmicsuperintelligence/openevolve}
}

@misc{zhang2024mleagent,
  title = {MLE-Agent: Your Intelligent Companion for Seamless AI Engineering and Research},
  author = {Huaizheng Zhang and Yizheng Huang and Lei Zhang},
  year = {2024},
  note = {\url{https://github.com/MLSysOps/MLE-agent}},
}

@misc{anthropic2025claude4systemcard,
  title        = {System Card: Claude Opus 4 \& Claude Sonnet 4},
  author       = {{Anthropic}},
  year         = {2025},
  howpublished = {\url{https://www-cdn.anthropic.com/07b2a3f9902ee19fe39a36ca638e5ae987bc64dd.pdf}}
}

@misc{jordan2024muon,
  author       = {Keller Jordan and Yuchen Jin and Vlado Boza and Jiacheng You and
                  Franz Cesista and Laker Newhouse and Jeremy Bernstein},
  title        = {Muon: An optimizer for hidden layers in neural networks},
  year         = {2024},
  url          = {https://kellerjordan.github.io/posts/muon/}
}

@inproceedings{he2016deepresidual,
  title     = {Deep Residual Learning for Image Recognition},
  author    = {He, Kaiming and Zhang, Xiangyu and Ren, Shaoqing and Sun, Jian},
  year      = {2016},
  booktitle = {2016 IEEE Conference on Computer Vision and Pattern Recognition (CVPR)},
  pages     = {770--778}
}

@article{mnih2015humanlevel,
  title     = {Human-level control through deep reinforcement learning},
  author    = {Volodymyr Mnih and Koray Kavukcuoglu and David Silver and Andrei A. Rusu and Joel Veness and Marc G. Bellemare and Alex Graves and Martin Riedmiller and Andreas Fidjeland and Georg Ostrovski and Stig Petersen and Charles Beattie and Amir Sadik and Ioannis Antonoglou and Helen King and Dharshan Kumaran and Daan Wierstra and Shane Legg and Demis Hassabis},
  year      = {2015},
  journal   = {Nature},
  volume    = {518},
  number    = {7540},
  pages     = {529--533}
}

@inproceedings{radford2021learningtransferable,
  title={Learning transferable visual models from natural language supervision},
  author={Radford, Alec and Kim, Jong Wook and Hallacy, Chris and Ramesh, Aditya and Goh, Gabriel and Agarwal, Sandhini and Sastry, Girish and Askell, Amanda and Mishkin, Pamela and Clark, Jack and others},
  booktitle={International conference on machine learning},
  pages={8748--8763},
  year={2021},
  organization={PmLR}
}

@article{kingma2014adam,
  title     = {Adam: A Method for Stochastic Optimization},
  author    = {Diederik P. Kingma and Jimmy Ba},
  year      = {2014},
  journal   = {arXiv preprint arXiv:1412.6980}
}

@article{kaplan2020scalinglaws,
  title     = {Scaling Laws for Neural Language Models},
  author    = {Jared Kaplan and Sam McCandlish and Tom Henighan and T. B. Brown and Benjamin Chess and Rewon Child and Scott Gray and Alec Radford and Jeffrey Wu and Dario Amodei},
  year      = {2020},
  journal   = {arXiv preprint arXiv:2001.08361}
}

@article{zhang2019rmsnorm,
  title     = {Root Mean Square Layer Normalization},
  author    = {Biao Zhang and Rico Sennrich},
  year      = {2019},
  journal   = {Neural Information Processing Systems},
  volume    = {32},
  pages     = {12360--12371}
}

@article{liu2025muon,
  title={Muon is scalable for llm training},
  author={Liu, Jingyuan and Su, Jianlin and Yao, Xingcheng and Jiang, Zhejun and Lai, Guokun and Du, Yulun and Qin, Yidao and Xu, Weixin and Lu, Enzhe and Yan, Junjie and others},
  journal={arXiv preprint arXiv:2502.16982},
  year={2025}
}

@article{dao2022flashattention,
  title={Flashattention: Fast and memory-efficient exact attention with io-awareness},
  author={Dao, Tri and Fu, Dan and Ermon, Stefano and Rudra, Atri and R{\'e}, Christopher},
  journal={Advances in neural information processing systems},
  volume={35},
  pages={16344--16359},
  year={2022}
}

@article{vaswani2017attention,
  title={Attention is all you need},
  author={Vaswani, Ashish and Shazeer, Noam and Parmar, Niki and Uszkoreit, Jakob and Jones, Llion and Gomez, Aidan N and Kaiser, {\L}ukasz and Polosukhin, Illia},
  journal={Advances in neural information processing systems},
  volume={30},
  year={2017}
}

@article{ho2020denoisingdiffusion,
  title={Denoising diffusion probabilistic models},
  author={Ho, Jonathan and Jain, Ajay and Abbeel, Pieter},
  journal={Advances in neural information processing systems},
  volume={33},
  pages={6840--6851},
  year={2020}
}

@article{schulman2017ppo,
  title={Proximal policy optimization algorithms},
  author={Schulman, John and Wolski, Filip and Dhariwal, Prafulla and Radford, Alec and Klimov, Oleg},
  journal={arXiv preprint arXiv:1707.06347},
  year={2017}
}

@article{hoffmann2022chinchilla,
  title   = {Training Compute-Optimal Large Language Models},
  author  = {Hoffmann, Jordan and Borgeaud, Sebastian and Mensch, Arthur and Buchatskaya, Elena and Cai, Trevor and Rutherford, Eliza and de Las Casas, Diego and Hendricks, Lisa Anne and Welbl, Johannes and Clark, Aidan and Hennigan, Tom and Noland, Eric and Millican, Katie and van den Driessche, George and Damoc, Bogdan and Guy, Aurelia and Osindero, Simon and Simonyan, Karen and Elsen, Erich and Rae, Jack W. and Vinyals, Oriol and Sifre, Laurent},
  journal = {arXiv preprint arXiv:2203.15556},
  year    = {2022},
  doi     = {10.48550/arXiv.2203.15556},
  url     = {https://arxiv.org/abs/2203.15556}
}

@inproceedings{yang2022mup,
  title     = {Tensor Programs V: Tuning Large Neural Networks via Zero-Shot Hyperparameter Transfer},
  author    = {Yang, Greg and Hu, Edward J. and Babuschkin, Igor and Sidor, Szymon and Liu, Xiaodong and Farhi, David and Ryder, Nick and Pachocki, Jakub and Chen, Weizhu and Gao, Jianfeng},
  booktitle = {Advances in Neural Information Processing Systems},
  volume    = {34},
  year      = {2021},
  url       = {https://proceedings.neurips.cc/paper/2021/hash/8df7c2e3c3c3be098ef7b382bd2c37ba-Abstract.html}
}

@inproceedings{everett2024scalingexponents,
  title     = {Scaling Exponents Across Parameterizations and Optimizers},
  author    = {Everett, Katie E. and Xiao, Lechao and Wortsman, Mitchell and Alemi, Alexander A. and Novak, Roman and Liu, Peter J. and Gur, Izzeddin and Sohl-Dickstein, Jascha and Kaelbling, Leslie Pack and Lee, Jaehoon and Pennington, Jeffrey},
  booktitle = {Proceedings of the 41st International Conference on Machine Learning},
  series    = {Proceedings of Machine Learning Research},
  volume    = {235},
  pages     = {12666--12700},
  publisher = {PMLR},
  year      = {2024},
  url       = {https://proceedings.mlr.press/v235/everett24a.html}
}

@inproceedings{kaddour2023nogain,
  title     = {No Train No Gain: Revisiting Efficient Training Algorithms For Transformer-based Language Models},
  author    = {Kaddour, Jean and Key, Oscar and Nawrot, Piotr and Minervini, Pasquale and Kusner, Matt J.},
  booktitle = {Advances in Neural Information Processing Systems},
  volume    = {36},
  year      = {2023},
  url       = {https://proceedings.neurips.cc/paper_files/paper/2023/hash/51f3d6252706100325ddc435ba0ade0e-Abstract-Conference.html}
}

@article{wen2025fantasticoptimizers,
  title   = {Fantastic Pretraining Optimizers and Where to Find Them},
  author  = {Wen, Kaiyue and Hall, David and Ma, Tengyu and Liang, Percy},
  journal = {arXiv preprint arXiv:2509.02046},
  year    = {2025},
  doi     = {10.48550/arXiv.2509.02046},
  url     = {https://arxiv.org/abs/2509.02046}
}

@inproceedings{choshen2024misfitting,
  title     = {(Mis)Fitting Scaling Laws: A Survey of Scaling Law Fitting Techniques in Deep Learning},
  author    = {Li, Margaret and Kudugunta, Sneha and Zettlemoyer, Luke},
  booktitle = {International Conference on Learning Representations},
  year      = {2025},
  url       = {https://openreview.net/forum?id=xI71dsS3o4}
}

@inproceedings{caballero2023broken,
  title     = {Broken Neural Scaling Laws},
  author    = {Caballero, Ethan and Gupta, Kshitij and Rish, Irina and Krueger, David},
  booktitle = {International Conference on Learning Representations},
  year      = {2023},
  url       = {https://openreview.net/forum?id=sckjveqlCZ}
}

@article{wei2022emergent,
  title   = {Emergent Abilities of Large Language Models},
  author  = {Wei, Jason and Tay, Yi and Bommasani, Rishi and Raffel, Colin and Zoph, Barret and Borgeaud, Sebastian and Yogatama, Dani and Bosma, Maarten and Zhou, Denny and Metzler, Donald and Chi, Ed H. and Hashimoto, Tatsunori and Vinyals, Oriol and Liang, Percy and Dean, Jeff and Fedus, William},
  journal = {Transactions on Machine Learning Research},
  year    = {2022},
  url     = {https://openreview.net/forum?id=yzkSU5zdwD}
}

@inproceedings{schaeffer2023mirage,
  title     = {Are Emergent Abilities of Large Language Models a Mirage?},
  author    = {Schaeffer, Rylan and Miranda, Brando and Koyejo, Sanmi},
  booktitle = {Advances in Neural Information Processing Systems},
  volume    = {36},
  year      = {2023},
  url       = {https://proceedings.neurips.cc/paper_files/paper/2023/hash/adc98a266f45005c403b8311ca7e8bd7-Abstract-Conference.html}
}

@article{ye2026evaluationdriven,
  title   = {Evaluation-driven Scaling for Scientific Discovery},
  author  = {Ye, Haotian and Lin, Haowei and Tang, Jingyi and Luo, Yizhen and Yang, Caiyin and Su, Chang and Thapa, Rahul and Yang, Rui and Liu, Ruihua and Li, Zeyu and Gao, Chong and Ding, Dachao and He, Guangrong and Zhang, Miaolei and Sun, Lina and Wang, Wenyang and Zhong, Yuchen and Shen, Zhuohao and He, Di and Ma, Jianzhu and Ermon, Stefano and Li, Tongyang and Chu, Xiaowen and Zou, James and Xu, Yuzhi},
  journal = {arXiv preprint arXiv:2604.19341},
  year    = {2026},
  url     = {https://arxiv.org/abs/2604.19341}
}

@misc{autolab2026benchmark,
  title  = {AutoLab: Can Models Begin to Participate in the Loops That Drive Scientific and Engineering Progress?},
  author = {{AutoLab Team}},
  year   = {2026},
  url    = {https://github.com/autolabhq/autolab}
}

@misc{Harbor_Framework,
  author = {{Harbor Framework Team}},
  month  = jan,
  title  = {{Harbor: A framework for evaluating and optimizing agents and models in container environments}},
  url    = {https://github.com/harbor-framework/harbor},
  year   = {2026}
}


\clearpage
\appendix



\section{Full Task Catalog}
\label{app:task_catalog}

Table~\ref{tab:tasks-appendix} lists the full \bench task catalog, grouped by research area, with each task's research question, external package(s), baselines, and evaluation settings.

\begingroup
\footnotesize
\renewcommand{\arraystretch}{1.05}
\setlength{\tabcolsep}{3pt}
\sloppy
\begin{xltabular}{\textwidth}{
  >{\raggedright\arraybackslash\hsize=0.75\hsize}X
  >{\raggedright\arraybackslash\hsize=1.45\hsize}X
  >{\raggedright\arraybackslash\hsize=0.80\hsize}X
  >{\raggedright\arraybackslash\hsize=0.90\hsize}X
  >{\raggedright\arraybackslash\hsize=1.10\hsize}X}
\caption{The full MLS-Bench task catalog grouped by research area. Each row gives the formal task name, a one-sentence research question, the external package(s) that supply the training and evaluation pipeline (\texttt{author/repo} for upstream packages, \emph{custom} for in-house scaffolds), the registered baselines, and the model/dataset evaluation settings.}\label{tab:tasks-appendix}\\
\toprule
\textbf{Name} & \textbf{Description} & \textbf{External Package(s)} & \textbf{Baselines} & \textbf{Evaluation Settings} \\
\midrule
\endfirsthead
\multicolumn{5}{l}{\small\textit{(Table~\ref{tab:tasks-appendix} continued from previous page)}}\\
\toprule
\textbf{Name} & \textbf{Description} & \textbf{External Package(s)} & \textbf{Baselines} & \textbf{Evaluation Settings} \\
\midrule
\endhead
\midrule
\multicolumn{5}{r}{\small\textit{(continued on next page)}}\\
\endfoot
\bottomrule
\endlastfoot

\midrule
\multicolumn{5}{l}{\textbf{Language Models (\textsc{LM})}} \\
\midrule
LLM Agent Tool-Use Reasoning Strategy &
  Studies how tool-use search, backtracking, and stopping policies affect answer validity and query efficiency. &
  \texttt{\seqsplit{zhichengg}}/\allowbreak \texttt{\seqsplit{StableToolBench}} &
  Greedy Chain (CoT) \newline DFS with LLM Ranking \newline DFSDT &
  StableToolBench I1-instruction 50q / deepseek-chat \newline StableToolBench I1-instruction 50q / qwen2.5-72b-instruct \newline StableToolBench I1-instruction 50q / qwen2.5-7b-instruct \\
\arrayrulecolor{black!15}\hline\arrayrulecolor{black}
Masked Diffusion LM: Demasking Strategy &
  Studies how demasking schedules, position selection, and token assignment affect diffusion language-model quality and decoding efficiency. &
  \texttt{\seqsplit{ML-GSAI}}/\allowbreak \texttt{\seqsplit{LLaDA}} &
  Top-K Margin \newline Confidence Greedy \newline KLASS &
  LLaDA / MATH-500 \newline LLaDA / HumanEval \newline Dream / C4 prefix continuation \\
\arrayrulecolor{black!15}\hline\arrayrulecolor{black}
Autoregressive Attention Mechanism &
  Studies how self-attention computation and positional handling affect autoregressive pretraining loss and downstream accuracy. &
  \texttt{\seqsplit{karpathy}}/\allowbreak \texttt{\seqsplit{nanoGPT}} \newline \texttt{\seqsplit{EleutherAI}}/\allowbreak \texttt{\seqsplit{lm-evaluation-harness}} &
  QK-Norm \newline RoPE \newline RoPE + QK-Norm &
  ClimbMix val loss + WikiText-2/LAMBADA PPL \newline HellaSwag, ARC-Easy, PIQA, WinoGrande 0-shot accuracy \\
\arrayrulecolor{black!15}\hline\arrayrulecolor{black}
Low-Bit Linear Pretraining Layer &
  Studies how low-bit linear layers and quantization functions affect pretraining loss under discrete weight constraints. &
  \texttt{\seqsplit{karpathy}}/\allowbreak \texttt{\seqsplit{nanoGPT}} \newline \texttt{\seqsplit{EleutherAI}}/\allowbreak \texttt{\seqsplit{lm-evaluation-harness}} &
  Binary Sign (BitNet) \newline Ternary 1.58-bit (BitNet b1.58) \newline INT2 Uniform &
  ClimbMix val loss + WikiText-2/LAMBADA PPL \newline HellaSwag, ARC-Easy, PIQA, WinoGrande 0-shot accuracy \\
\arrayrulecolor{black!15}\hline\arrayrulecolor{black}
Autoregressive Embedding Strategy &
  Studies how token embeddings, position embeddings, value embeddings, and weight tying affect autoregressive pretraining loss and downstream accuracy. &
  \texttt{\seqsplit{karpathy}}/\allowbreak \texttt{\seqsplit{nanoGPT}} \newline \texttt{\seqsplit{EleutherAI}}/\allowbreak \texttt{\seqsplit{lm-evaluation-harness}} &
  Untied Embeddings \newline Value Embeddings \newline Bigram Hash Embeddings &
  ClimbMix val loss + WikiText-2/LAMBADA PPL \newline HellaSwag, ARC-Easy, PIQA, WinoGrande 0-shot accuracy \\
\arrayrulecolor{black!15}\hline\arrayrulecolor{black}
Subquadratic Attention Mechanism &
  Studies whether linear or subquadratic attention can reduce autoregressive validation loss while preserving downstream performance. &
  \texttt{\seqsplit{karpathy}}/\allowbreak \texttt{\seqsplit{nanoGPT}} \newline \texttt{\seqsplit{EleutherAI}}/\allowbreak \texttt{\seqsplit{lm-evaluation-harness}} &
  RetNet \newline DeltaNet \newline GLA &
  ClimbMix val loss + WikiText-2/LAMBADA PPL \newline HellaSwag, ARC-Easy, PIQA, WinoGrande 0-shot accuracy \\
\arrayrulecolor{black!15}\hline\arrayrulecolor{black}
Autoregressive Pretraining Loss &
  Studies how alternative next-token training losses affect autoregressive validation cross-entropy. &
  \texttt{\seqsplit{karpathy}}/\allowbreak \texttt{\seqsplit{nanoGPT}} \newline \texttt{\seqsplit{EleutherAI}}/\allowbreak \texttt{\seqsplit{lm-evaluation-harness}} &
  Label Smoothing \newline Softcap Cross-Entropy \newline Z-Loss &
  ClimbMix val loss + WikiText-2/LAMBADA PPL \newline HellaSwag, ARC-Easy, PIQA, WinoGrande 0-shot accuracy \\
\arrayrulecolor{black!15}\hline\arrayrulecolor{black}
Pretraining Learning-Rate Schedule &
  Studies how warmup, decay shape, and schedule horizon affect autoregressive pretraining validation loss. &
  \texttt{\seqsplit{karpathy}}/\allowbreak \texttt{\seqsplit{nanoGPT}} \newline \texttt{\seqsplit{EleutherAI}}/\allowbreak \texttt{\seqsplit{lm-evaluation-harness}} &
  WSD (Warmup-Stable-Decay) \newline Trapezoidal \newline WSD with Inverse-Sqrt Decay &
  ClimbMix val loss + WikiText-2/LAMBADA PPL \newline HellaSwag, ARC-Easy, PIQA, WinoGrande 0-shot accuracy \\
\arrayrulecolor{black!15}\hline\arrayrulecolor{black}
Transformer Feed-Forward Block &
  Studies how activation, gating, and expansion choices in the feed-forward sublayer affect language-model validation loss. &
  \texttt{\seqsplit{karpathy}}/\allowbreak \texttt{\seqsplit{nanoGPT}} \newline \texttt{\seqsplit{EleutherAI}}/\allowbreak \texttt{\seqsplit{lm-evaluation-harness}} &
  ReLU-Squared \newline SwiGLU \newline GeGLU &
  ClimbMix val loss + WikiText-2/LAMBADA PPL \newline HellaSwag, ARC-Easy, PIQA, WinoGrande 0-shot accuracy \\
\arrayrulecolor{black!15}\hline\arrayrulecolor{black}
Normalization and Block Layout &
  Studies how normalization placement, affine behavior, and transformer block layout affect pretraining stability and validation loss. &
  \texttt{\seqsplit{karpathy}}/\allowbreak \texttt{\seqsplit{nanoGPT}} \newline \texttt{\seqsplit{EleutherAI}}/\allowbreak \texttt{\seqsplit{lm-evaluation-harness}} &
  RMSNorm \newline RMSNorm + Sandwich-Norm \newline RMSNorm (Parallel Block) &
  ClimbMix val loss + WikiText-2/LAMBADA PPL \newline HellaSwag, ARC-Easy, PIQA, WinoGrande 0-shot accuracy \\
\arrayrulecolor{black!15}\hline\arrayrulecolor{black}
Pretraining Optimizer Design &
  Studies how optimizer choice, parameter grouping, and schedule coupling affect autoregressive pretraining validation loss. &
  \texttt{\seqsplit{karpathy}}/\allowbreak \texttt{\seqsplit{nanoGPT}} \newline \texttt{\seqsplit{EleutherAI}}/\allowbreak \texttt{\seqsplit{lm-evaluation-harness}} &
  AdamW + Nesterov \newline Lion \newline Muon &
  ClimbMix val loss + WikiText-2/LAMBADA PPL \newline HellaSwag, ARC-Easy, PIQA, WinoGrande 0-shot accuracy \\
\arrayrulecolor{black!15}\hline\arrayrulecolor{black}
Transformer Residual Stream Strategy &
  Studies how residual connections and information flow across transformer layers affect validation loss, perplexity, and accuracy metrics. &
  \texttt{\seqsplit{karpathy}}/\allowbreak \texttt{\seqsplit{nanoGPT}} \newline \texttt{\seqsplit{EleutherAI}}/\allowbreak \texttt{\seqsplit{lm-evaluation-harness}} &
  Vanilla (Pre-LN) \newline ProRes \newline Learned Scaling \newline Block Attention Residuals &
  ClimbMix val loss + WikiText-2/LAMBADA PPL \newline HellaSwag, ARC-Easy, PIQA, WinoGrande 0-shot accuracy \\
\arrayrulecolor{black!15}\hline\arrayrulecolor{black}
Reasoning RL Advantage Estimation &
  Studies how advantage estimates for online language-model reinforcement learning affect mathematical reasoning accuracy. &
  \texttt{\seqsplit{volcengine}}/\allowbreak \texttt{\seqsplit{verl}} &
  GRPO \newline Dr. GRPO \newline Reinforce++ Baseline &
  GSM8K \newline MATH-500 \newline AMC \\
\arrayrulecolor{black!15}\hline\arrayrulecolor{black}
Reasoning RL Importance-Sampling Granularity &
  Studies how importance-sampling ratio granularity and clipping affect online language-model reinforcement learning for reasoning. &
  \texttt{\seqsplit{volcengine}}/\allowbreak \texttt{\seqsplit{verl}} &
  Token-Level (Vanilla PPO) \newline Sequence-Level (GSPO) \newline First-K Tokens &
  GSM8K \newline MATH-500 \newline AMC \\
\arrayrulecolor{black!15}\hline\arrayrulecolor{black}
Actor Divergence Estimator for Reasoning RL &
  Studies how per-token actor KL estimation controls reference-policy drift while preserving reasoning accuracy during online RL. &
  \texttt{\seqsplit{volcengine}}/\allowbreak \texttt{\seqsplit{verl}} &
  K1 (Unbiased Log-Ratio) \newline K2 (Squared Log-Ratio) \newline K3 (Low-Variance KL) \newline Absolute Log-Ratio &
  GSM8K \newline MATH-500 \newline AMC \\
\arrayrulecolor{black!15}\hline\arrayrulecolor{black}
Pre-Advantage Reward Normalization &
  Studies how reward normalization before advantage estimation affects reasoning accuracy in online language-model RL. &
  \texttt{\seqsplit{volcengine}}/\allowbreak \texttt{\seqsplit{verl}} &
  Outcome-Only (Raw) \newline Group-Std Normalization \newline Batch-Std Whitening \newline Length-Aware Normalization &
  GSM8K \newline MATH-500 \newline AMC \\
\arrayrulecolor{black!15}\hline\arrayrulecolor{black}
Symbolic Scaling-Law Discovery &
  Studies how symbolic functional forms and group-specific coefficients capture held-out scaling behavior. &
  \texttt{\seqsplit{linhaowei1}}/\allowbreak \texttt{\seqsplit{SLD}} &
  Human Exact Form \newline SLDAgent-Style \newline Kernel Ridge Regression \newline XGBoost &
  SLDBench Vocabulary Scaling \newline SLDBench LR x Batch-Size Scaling \newline SLDBench Data-Constrained Scaling \\
\arrayrulecolor{black!15}\hline\arrayrulecolor{black}
Language-Agent Collaboration Topology &
  Studies how deterministic collaboration topology affects multi-agent code-generation quality and execution success. &
  \texttt{\seqsplit{OpenBMB}}/\allowbreak \texttt{\seqsplit{ChatDev}} &
  Chain \newline Star \newline Layered &
  HumanEval-33 (deepseek-chat, 4 agents) \newline HumanEval-33 (qwen2.5-72b-instruct, 4 agents) \newline SRDD-20 (deepseek-chat, 4 agents) \\

\midrule
\multicolumn{5}{l}{\textbf{Robotics (\textsc{Rob})}} \\
\midrule
Latent World-Model Planner &
  Studies how goal-conditioned planning should exploit a fixed latent world model to improve navigation success. &
  \texttt{\seqsplit{facebookresearch}}/\allowbreak \texttt{\seqsplit{eb\_jepa}} &
  Random \newline CEM \newline MPPI \newline iCEM &
  Two Rooms (Horizon 30) \newline Two Rooms (Horizon 60) \newline Two Rooms (Horizon 90) \\
\arrayrulecolor{black!15}\hline\arrayrulecolor{black}
Temporal Latent Prediction Loss &
  Studies how latent prediction objectives affect multi-step video representation quality. &
  \texttt{\seqsplit{facebookresearch}}/\allowbreak \texttt{\seqsplit{eb\_jepa}} &
  MSE \newline Smooth L1 \newline Cosine &
  Moving MNIST AP (small: henc=16, dstc=8, hpre=16) \newline Moving MNIST AP (base: henc=32, dstc=16, hpre=32) \newline Moving MNIST AP (large: henc=64, dstc=32, hpre=64) \\
\arrayrulecolor{black!15}\hline\arrayrulecolor{black}
Anti-Collapse Representation Regularizer &
  Studies how self-supervised regularization prevents representation collapse and improves linear-probe accuracy. &
  \texttt{\seqsplit{facebookresearch}}/\allowbreak \texttt{\seqsplit{eb\_jepa}} &
  Naive \newline VICReg \newline SigReg \newline Barlow Twins &
  ResNet-18 Probe \newline ResNet-34 Probe \newline ResNet-50 Probe \\
\arrayrulecolor{black!15}\hline\arrayrulecolor{black}
Diffusion Guidance for Robot Trajectory Planning &
  Studies guidance mechanisms for a fixed trajectory-level diffusion planner on D4RL MuJoCo, optimizing normalized score across hopper-medium-v2, walker2d-medium-v2, and halfcheetah-medium-v2. &
  \texttt{\seqsplit{CleanDiffuserTeam}}/\allowbreak \texttt{\seqsplit{CleanDiffuser}} &
  Diffuser (Classifier Guidance) \newline Classifier-Free Guidance \newline No Guidance \newline Decision Diffuser &
  D4RL Hopper-Medium-v2 \newline D4RL Walker2d-Medium-v2 \newline D4RL HalfCheetah-Medium-v2 \\
\arrayrulecolor{black!15}\hline\arrayrulecolor{black}
Diffusion Policy Learning for Robot Control &
  Studies how diffusion policy training, value guidance, and action generation affect robot-control episode reward. &
  \texttt{\seqsplit{CleanDiffuserTeam}}/\allowbreak \texttt{\seqsplit{CleanDiffuser}} &
  DQL (Diffusion Q-Learning) \newline IDQL \newline Diffusion Policy &
  D4RL Hopper-Medium-v2 \newline D4RL Walker2d-Medium-v2 \newline D4RL HalfCheetah-Medium-v2 \\
\arrayrulecolor{black!15}\hline\arrayrulecolor{black}
Efficient Diffusion Sampling for Robot Actions &
  Studies how solver choice and sampling\_steps affect DQL-style diffusion-policy normalized score at low NFE on D4RL MuJoCo. &
  \texttt{\seqsplit{CleanDiffuserTeam}}/\allowbreak \texttt{\seqsplit{CleanDiffuser}} &
  DDPM (100-Step Ancestral Sampling) \newline DDIM (20-Step Deterministic Sampling) \newline DPM-Solver++ 2M (10-Step) &
  D4RL Hopper-Medium-v2 \newline D4RL Walker2d-Medium-v2 \newline D4RL HalfCheetah-Medium-v2 \\
\arrayrulecolor{black!15}\hline\arrayrulecolor{black}
Humanoid Transfer Policy Learning &
  Studies how actor-critic architecture, policy optimization, and rollout processing affect humanoid command-following transfer. &
  \texttt{\seqsplit{roboterax}}/\allowbreak \texttt{\seqsplit{humanoid-gym}} &
  Default PPO \newline PPO with Adaptive KL \newline PPO with LayerNorm &
  RobotEra XBot-L Training \newline RobotEra XBot-L / Diverse Commands \newline RobotEra XBot-L / Forward-Only \newline RobotEra XBot-L / High Speed \\
\arrayrulecolor{black!15}\hline\arrayrulecolor{black}
Behavioral Cloning Loss for Manipulation &
  Studies how imitation-learning loss design affects rollout success for low-dimensional robot manipulation tasks. &
  \texttt{\seqsplit{ARISE-Initiative}}/\allowbreak \texttt{\seqsplit{robomimic}} &
  NLL with Entropy \newline Weighted NLL \newline Default (NLL) &
  Tool Hang (PH) \newline Can (PH) \newline Square (PH) \\
\arrayrulecolor{black!15}\hline\arrayrulecolor{black}
Offline Value Loss for Manipulation &
  Studies how asymmetric value regression loss design affects offline robot manipulation policy success. &
  \texttt{\seqsplit{ARISE-Initiative}}/\allowbreak \texttt{\seqsplit{robomimic}} &
  Quantile Regression \newline Huber Pinball \newline Default (Expectile) &
  Tool Hang (PH) \newline Can (PH) \newline Square (PH) \\
\arrayrulecolor{black!15}\hline\arrayrulecolor{black}
Observation Fusion Encoder for Imitation Learning &
  Designs a multimodal robot state encoder for behavioral cloning to improve rollout success rate on manipulation tasks. &
  \texttt{\seqsplit{ARISE-Initiative}}/\allowbreak \texttt{\seqsplit{robomimic}} &
  Attention Fusion \newline Gated Fusion \newline Default (Concatenation) &
  Tool Hang (PH) \newline Can (PH) \newline Square (PH) \\
\arrayrulecolor{black!15}\hline\arrayrulecolor{black}
Trajectory Optimization for Model-Based Planning &
  An online planning algorithm selects actions through learned-world-model trajectory optimization to improve episode reward. &
  \texttt{\seqsplit{nicklashansen}}/\allowbreak \texttt{\seqsplit{tdmpc2}} &
  CEM \newline iCEM \newline MPPI &
  Walker Walk \newline Cheetah Run \newline Cartpole Swingup \\
\arrayrulecolor{black!15}\hline\arrayrulecolor{black}
Latent Representation Normalization for Model-Based RL &
  Designs latent-state normalization for the TD-MPC2 encoder and dynamics world-model networks, evaluated by DMControl episode reward. &
  \texttt{\seqsplit{nicklashansen}}/\allowbreak \texttt{\seqsplit{tdmpc2}} &
  SimNorm \newline L2 normalization \newline RMSNorm \newline Identity (no normalization) &
  DMControl walker-walk \newline DMControl cheetah-run \newline DMControl cartpole-swingup \\

\midrule
\multicolumn{5}{l}{\textbf{Vision \& Generation (\textsc{V\&G})}} \\
\midrule
3D Gaussian Splatting Densification Strategy Design &
  Designs a 3D Gaussian Splatting densification strategy controlling clone, split, prune, reset, relocation, and sample-add behavior to improve held-out novel-view quality on Mip-NeRF 360 scenes. &
  \texttt{\seqsplit{nerfstudio-project}}/\allowbreak \texttt{\seqsplit{gsplat}} &
  Original 3DGS densification \newline AbsGS + Taming-3DGS + New Split \newline EDC-TamingGS-Abs &
  Mip-NeRF 360 garden (8x, best PSNR) \newline Mip-NeRF 360 bicycle (8x, best PSNR) \newline Mip-NeRF 360 bonsai (8x, best PSNR) \newline Mip-NeRF 360 stump (8x, best PSNR) \\
\arrayrulecolor{black!15}\hline\arrayrulecolor{black}
3D Gaussian Splatting Regularizer Design &
  Designs a scalar regularizer added to the 3DGS photometric loss during 30k-step Mip-NeRF 360 reconstruction, evaluated on held-out novel views and scored by best PSNR. &
  \texttt{\seqsplit{nerfstudio-project}}/\allowbreak \texttt{\seqsplit{gsplat}} &
  No regularization \newline Scale + opacity L1 \newline Effective-rank + scale/opacity L1 &
  Mip-NeRF 360 garden (8x, best PSNR) \newline Mip-NeRF 360 bicycle (8x, best PSNR) \newline Mip-NeRF 360 bonsai (8x, best PSNR) \newline Mip-NeRF 360 stump (8x, best PSNR) \\
\arrayrulecolor{black!15}\hline\arrayrulecolor{black}
Custom Sampler for Diffusion Bridge Models &
  Designs a low-NFE sampler for Diffusion Bridge Models on image-to-image translation, ImageNet center-inpainting, and DIODE depth, evaluated by FID at NFE=5. &
  \texttt{\seqsplit{thu-ml}}/\allowbreak \texttt{\seqsplit{DiffusionBridge}} &
  DBIM \newline DBIM-HO (high-order) \newline DDBM (50 NFE reference) \newline ECSI &
  Edges2Handbags / e2h (FID, NFE=5) \newline ImageNet center-inpaint (FID, NFE=5) \newline DIODE depth (FID, NFE=5) \\
\arrayrulecolor{black!15}\hline\arrayrulecolor{black}
Time Scheduler for Diffusion Bridge Models (NFE=5) &
  Designs a monotone low-step time schedule for Diffusion Bridge Models, evaluated by FID on Edges2Handbags, ImageNet center-inpainting, and DIODE depth at NFE=5. &
  \texttt{\seqsplit{thu-ml}}/\allowbreak \texttt{\seqsplit{DiffusionBridge}} &
  Karras EDM (rho=7) \newline Uniform (linear) \newline Cosine (Nichol-Dhariwal) \newline Log-linear (geometric) &
  Edges2Handbags / e2h (FID, NFE=5) \newline ImageNet center-inpaint (FID, NFE=5) \newline DIODE depth (FID, NFE=5) \\
\arrayrulecolor{black!15}\hline\arrayrulecolor{black}
Diffusion Model Architecture Design &
  Design a denoising UNet backbone for unconditional CIFAR-10 DDPM training, optimizing best FID with fixed epsilon prediction and 50-step DDIM sampling. &
  \texttt{\seqsplit{huggingface}}/\allowbreak \texttt{\seqsplit{diffusers}} &
  Standard DDPM U-Net \newline Full-Attention U-Net \newline No-Attention U-Net &
  CIFAR-10 DDPM Small \newline CIFAR-10 DDPM Medium \newline CIFAR-10 DDPM Large \\
\arrayrulecolor{black!15}\hline\arrayrulecolor{black}
Diffusion Model: Classifier-Free Guidance Optimization &
  Design a classifier-free guidance method for Stable Diffusion text-to-image generation across SD v1.5, Stable Diffusion 2 Base, and Stable Diffusion XL; evaluation generates COCO-caption images and official scoring uses per-model FID. &
  \texttt{\seqsplit{CFGpp-diffusion}}/\allowbreak \texttt{\seqsplit{CFGpp}} &
  Standard CFG \newline CFG++ \newline Zero-Init CFG++ &
  Stable Diffusion v1.5 / COCO captions / NFE=10 \newline Stable Diffusion 2 Base / COCO captions / NFE=10 \newline Stable Diffusion XL Base 1.0 / COCO captions / NFE=10 \\
\arrayrulecolor{black!15}\hline\arrayrulecolor{black}
Class-Conditional Diffusion: Conditioning Injection Methods &
  Design class-conditioning injection for a CIFAR-10 class-conditional UNet2DModel/DDPM, optimizing best FID with 50-step DDIM sampling. &
  \texttt{\seqsplit{huggingface}}/\allowbreak \texttt{\seqsplit{diffusers}} &
  Concat-FiLM \newline Cross-Attention \newline AdaLN-Zero &
  CIFAR-10 Class-Conditional Small UNet2DModel \newline CIFAR-10 Class-Conditional Medium UNet2DModel \newline CIFAR-10 Class-Conditional Large UNet2DModel \\
\arrayrulecolor{black!15}\hline\arrayrulecolor{black}
Diffusion Model: Sampler Efficiency Optimization &
  Design a Stable Diffusion sampler update rule for COCO-caption text-to-image generation at a fixed NFE=20 budget; official scoring uses per-model FID. &
  \texttt{\seqsplit{CFGpp-diffusion}}/\allowbreak \texttt{\seqsplit{CFGpp}} &
  DDIM \newline DPM++ 3M \newline DPM++ 2S &
  Stable Diffusion v1.5 / COCO captions / NFE=20 \newline Stable Diffusion 2 Base / COCO captions / NFE=20 \newline Stable Diffusion XL Base 1.0 / COCO captions / NFE=20 \\
\arrayrulecolor{black!15}\hline\arrayrulecolor{black}
Diffusion Prediction Parameterization &
  Design a prediction target and consistent x0 inversion for unconditional CIFAR-10 UNet2DModel diffusion, optimizing best FID with 50-step DDIM sampling. &
  \texttt{\seqsplit{huggingface}}/\allowbreak \texttt{\seqsplit{diffusers}} &
  Epsilon Prediction \newline V-Prediction \newline X0 Prediction &
  CIFAR-10 Unconditional Small UNet2DModel \newline CIFAR-10 Unconditional Medium UNet2DModel \newline CIFAR-10 Unconditional Large UNet2DModel \\
\arrayrulecolor{black!15}\hline\arrayrulecolor{black}
Flow Map with Perceptual Loss &
  Studies whether auxiliary perceptual losses on denoised images improve CIFAR-10 FID for MeanFlow flow-map training with DiT backbones. &
  \texttt{\seqsplit{snap-research}}/\allowbreak \texttt{\seqsplit{alphaflow}} &
  Pure MSE Velocity \newline MSE + Charbonnier + LPIPS + Gradient + Multiscale \newline MSE + LPIPS + Gradient + Multiscale + FFT &
  CIFAR-10 Small DiT \newline CIFAR-10 Medium DiT \newline CIFAR-10 Large DiT \\
\arrayrulecolor{black!15}\hline\arrayrulecolor{black}
VAE Loss Function Design for Image Reconstruction &
  Studies how VAE loss components affect CIFAR-10 AutoencoderKL reconstruction quality, scored primarily by rFID on the full test set. &
  \texttt{\seqsplit{huggingface}}/\allowbreak \texttt{\seqsplit{diffusers}} &
  L1 + KL \newline L1 + LPIPS + KL \newline L1 + LPIPS + KL + PatchGAN &
  CIFAR-10 AutoencoderKL Small \newline CIFAR-10 AutoencoderKL Medium \newline CIFAR-10 AutoencoderKL Large \\

\midrule
\multicolumn{5}{l}{\textbf{Reinforcement Learning (\textsc{RL})}} \\
\midrule
Cooperative MARL Centralized Critic Architecture for MAPPO &
  Studies centralized critic architectures for MAPPO on SMACLite cooperative MARL maps, scored by greedy-policy test win rate and return. &
  \texttt{\seqsplit{uoe-agents}}/\allowbreak \texttt{\seqsplit{epymarl}} &
  IPPO Decentralized Critic \newline MAPPO Centralized Critic \newline MAT-Style Attention Critic &
  SMACLite MMM (10-agent heterogeneous) \newline SMACLite 2s3z (5-agent heterogeneous) \newline SMACLite 3s5z (8-agent heterogeneous) \\
\arrayrulecolor{black!15}\hline\arrayrulecolor{black}
Meta-RL: Context Encoder for PEARL Task Inference &
  Studies PEARL context encoders that map transition tuples to latent task representations for fast adaptation, evaluated by meta\_test\_return after 20 meta-training iterations. &
  \texttt{\seqsplit{katerakelly}}/\allowbreak \texttt{\seqsplit{oyster}} &
  PEARL MLP Context Encoder \newline PEARL Recurrent Context Encoder \newline PEARL Attention Context Encoder &
  Half-Cheetah Velocity (30 train/10 test tasks) \newline Sparse Point Robot (40 train/10 test tasks) \newline Point Robot (40 train/10 test tasks) \\
\arrayrulecolor{black!15}\hline\arrayrulecolor{black}
Meta-RL Algorithm Design &
  Studies complete meta-RL algorithm design across task inference, policy conditioning, and meta-training, scored by meta\_test\_return on held-out tasks after the fixed short-budget protocol. &
  \texttt{\seqsplit{katerakelly}}/\allowbreak \texttt{\seqsplit{oyster}} &
  PEARL \newline FOCAL \newline VariBAD &
  Half-Cheetah Velocity (30 train/10 test tasks) \newline Sparse Point Robot (40 train/10 test tasks) \newline Point Robot (40 train/10 test tasks) \\
\arrayrulecolor{black!15}\hline\arrayrulecolor{black}
Intrinsic Exploration for Sparse Rewards &
  Studies how intrinsic rewards and advantage mixing affect exploration and return in sparse-reward Atari environments. &
  \texttt{\seqsplit{vwxyzjn}}/\allowbreak \texttt{\seqsplit{cleanrl}} &
  PPO \newline RND \newline ICM &
  Tutankham-v5 \newline Frostbite-v5 \newline PrivateEye-v5 \\
\arrayrulecolor{black!15}\hline\arrayrulecolor{black}
Offline Dexterous Manipulation from Narrow Demonstrations &
  Studies how offline RL algorithms learn dexterous manipulation from narrow human demonstration datasets. &
  \texttt{\seqsplit{corl-team}}/\allowbreak \texttt{\seqsplit{CORL}} &
  IQL \newline AWAC \newline ReBRAC &
  Pen-Human-v1 \newline Hammer-Human-v1 \newline Door-Cloned-v1 \\
\arrayrulecolor{black!15}\hline\arrayrulecolor{black}
Q-Overestimation Suppression for Offline Continuous Control &
  Studies how offline continuous-control algorithms suppress out-of-distribution Q-value overestimation. &
  \texttt{\seqsplit{corl-team}}/\allowbreak \texttt{\seqsplit{CORL}} &
  ReBRAC \newline TD3-BC \newline IQL &
  HalfCheetah-Medium-v2 \newline Maze2D-Medium-v1 \newline Walker2d-Medium-v2 \\
\arrayrulecolor{black!15}\hline\arrayrulecolor{black}
Offline-to-Online Fine-Tuning Without Forgetting &
  Studies how offline-to-online reinforcement learning prevents forgetting and value collapse during continued interaction. &
  \texttt{\seqsplit{corl-team}}/\allowbreak \texttt{\seqsplit{CORL}} &
  IQL \newline AWAC \newline SPOT &
  Pen-Cloned-v1 \newline Hammer-Cloned-v1 \newline Hammer-Expert-v1 \\
\arrayrulecolor{black!15}\hline\arrayrulecolor{black}
Off-Policy Actor-Critic for Continuous Control &
  Changes off-policy actor-critic update rules, losses, or exploration strategies to improve mean episodic return on continuous-control tasks. &
  \texttt{\seqsplit{vwxyzjn}}/\allowbreak \texttt{\seqsplit{cleanrl}} &
  DDPG \newline TD3 \newline SAC &
  HalfCheetah-v4 \newline Reacher-v4 \newline Ant-v4 \\
\arrayrulecolor{black!15}\hline\arrayrulecolor{black}
On-Policy Actor-Critic for Continuous Control &
  Changes on-policy actor-critic objectives, update rules, or exploration mechanisms to improve mean episodic return on continuous-control tasks. &
  \texttt{\seqsplit{vwxyzjn}}/\allowbreak \texttt{\seqsplit{cleanrl}} &
  PPO \newline AWR \newline PPO (KL Penalty) &
  HalfCheetah-v4 \newline Swimmer-v4 \newline InvertedDoublePendulum-v4 \\
\arrayrulecolor{black!15}\hline\arrayrulecolor{black}
Inverse RL Reward Learning from Demonstrations &
  Studies how reward models learned from expert demonstrations affect downstream policy return in continuous-control locomotion. &
  \texttt{\seqsplit{HumanCompatibleAI}}/\allowbreak \texttt{\seqsplit{imitation}} &
  GAIL \newline AIRL \newline BC &
  HalfCheetah-v4 \newline Hopper-v4 \newline Walker2d-v4 \\
\arrayrulecolor{black!15}\hline\arrayrulecolor{black}
Value-Based Visual Control &
  Studies how value-based RL losses, update rules, and exploration strategies affect visual-control episodic return. &
  \texttt{\seqsplit{vwxyzjn}}/\allowbreak \texttt{\seqsplit{cleanrl}} &
  QR-DQN \newline C51 \newline Double-DQN &
  BreakoutNoFrameskip-v4 \newline SeaquestNoFrameskip-v4 \newline PongNoFrameskip-v4 \\
\arrayrulecolor{black!15}\hline\arrayrulecolor{black}
Value-Based Discrete Control &
  Changes value estimation, uncertainty handling, or replay-based update rules to improve episodic return on discrete-action control tasks. &
  \texttt{\seqsplit{vwxyzjn}}/\allowbreak \texttt{\seqsplit{cleanrl}} &
  QR-DQN \newline Dueling-DQN \newline C51 &
  CartPole-v1 \newline LunarLander-v2 \newline Acrobot-v1 \\
\arrayrulecolor{black!15}\hline\arrayrulecolor{black}
Constraint Handling for Safe RL &
  Changes Lagrangian or controller-style multiplier updates and cost-reward advantage mixing to improve reward while keeping episode cost below target. &
  \texttt{\seqsplit{PKU-Alignment}}/\allowbreak \texttt{\seqsplit{omnisafe}} &
  Naive PPO \newline Lagrangian PPO \newline PID Lagrangian &
  SafetyPointGoal1-v0 \newline SafetyCarGoal1-v0 \newline SafetyPointButton1-v0 \\

\midrule
\multicolumn{5}{l}{\textbf{ML Systems \& Efficient ML (\textsc{Sys})}} \\
\midrule
Diffusion LM KV Cache Policy &
  Studies how token-state refresh intervals, masks, transfer ratios, and fallbacks affect denoising quality and cache reuse. &
  \texttt{\seqsplit{maomaocun}}/\allowbreak \texttt{\seqsplit{dLLM-Cache}} &
  Vanilla (Uncached) \newline dLLM-Cache \newline d2Cache \newline Elastic-Cache &
  MATH-500 \newline HumanEval \newline ARC-Challenge \\
\arrayrulecolor{black!15}\hline\arrayrulecolor{black}
LLM KV Cache: Adaptive Quantization Policy &
  Studies adaptive 4-bit KV-cache quantization for instruction-tuned long-context inference, trading benchmark final-score quality against effective KV bits and compression. &
  \texttt{\seqsplit{huggingface}}/\allowbreak \texttt{\seqsplit{transformers}} &
  KIVI Overlap (4-bit) \newline KVTuner-4 Per-Token \newline KVTuner-4 KIVI \newline SQuat Subspace (4-bit) &
  LongBench-E hotpotqa\_e QA F1 \newline LongBench-E passage\_retrieval\_en\_e retrieval score \newline LongBench-E repobench-p\_e code-similarity score \newline NeedleBench NIAH exact phrase retrieval \newline GSM8K exact final-answer accuracy \\
\arrayrulecolor{black!15}\hline\arrayrulecolor{black}
LLM KV Cache Selection Budgeting &
  Studies how selection and eviction controllers allocate layer budgets and recent windows for quality, latency, and memory tradeoffs. &
  \texttt{\seqsplit{huggingface}}/\allowbreak \texttt{\seqsplit{transformers}} &
  Full Attention \newline StreamingLLM \newline Expected Attention \newline LagKV &
  LongBench-E hotpotqa\_e QA F1 \newline LongBench-E passage\_retrieval\_en\_e retrieval score \newline LongBench-E repobench-p\_e code-similarity score \newline LongBench v2 train split multiple-choice accuracy \newline GSM8K exact final-answer accuracy \\
\arrayrulecolor{black!15}\hline\arrayrulecolor{black}
LLM Pretraining: KV-Structural Reduction &
  Studies GPT-style KV-state structural reduction through MHA, MQA, GQA, and MLA-style latent KV compression under fixed nanoGPT pretraining. &
  \texttt{\seqsplit{karpathy}}/\allowbreak \texttt{\seqsplit{nanoGPT}} \newline \texttt{\seqsplit{EleutherAI}}/\allowbreak \texttt{\seqsplit{lm-evaluation-harness}} &
  MHA \newline MQA \newline GQA \newline MLA &
  ClimbMix val loss + KV bytes/token + WikiText-2/WikiText-103/LAMBADA heldout loss \newline HellaSwag, ARC-Easy, PIQA, WinoGrande 0-shot accuracy \\
\arrayrulecolor{black!15}\hline\arrayrulecolor{black}
LLM Pretraining: Custom GPU Kernel Optimization &
  Studies custom/fused MLP kernels for nanoGPT pretraining while preserving ClimbMix validation, held-out perplexity, and downstream lm-eval quality. &
  \texttt{\seqsplit{karpathy}}/\allowbreak \texttt{\seqsplit{nanoGPT}} \newline \texttt{\seqsplit{EleutherAI}}/\allowbreak \texttt{\seqsplit{lm-evaluation-harness}} &
  ReLU-Squared (Torch) \newline Triton GELU \newline Triton ReLU-Squared (Fused) &
  ClimbMix val loss + WikiText-2/LAMBADA PPL \newline HellaSwag, ARC-Easy, PIQA, WinoGrande 0-shot accuracy \\
\arrayrulecolor{black!15}\hline\arrayrulecolor{black}
LLM Post-Training Quantization (PTQ) Algorithm &
  Design a post-training quantization algorithm for a pretrained LLM that minimizes WikiText-2 perplexity degradation under INT4/INT3 group quantization without retraining. &
  \texttt{\seqsplit{IST-DASLab}}/\allowbreak \texttt{\seqsplit{gptq}} &
  Round-to-Nearest (RTN) \newline GPTQ \newline AWQ &
  PTQ INT4 \newline PTQ INT3 \newline PTQ INT4 (g64) \\
\arrayrulecolor{black!15}\hline\arrayrulecolor{black}
LLM Quantization-Aware Training (QAT) Algorithm &
  Design a quantization-aware training algorithm for a pretrained LLM that minimizes WikiText-2 perplexity after INT4/INT3/INT2 quantization at inference time. &
  custom &
  No QAT \newline STE \newline LSQ \newline Finetune + PTQ &
  QAT INT4 \newline QAT INT3 \newline QAT INT2 \\
\arrayrulecolor{black!15}\hline\arrayrulecolor{black}
Fused Attention Kernel Design for H100 GPUs &
  Design an OpenAI Triton fused self-attention forward kernel for H100 GPUs that maximizes throughput (TFLOPs/s) while preserving numerical correctness. &
  \texttt{\seqsplit{Dao-AILab}}/\allowbreak \texttt{\seqsplit{flash-attention}} &
  FlashAttention \newline FlashAttention-2 \newline FlashAttention-3 &
  Head Dim 64 / Seq 4K \newline Head Dim 128 / Seq 8K \newline Head Dim 256 / Seq 16K \\
\arrayrulecolor{black!15}\hline\arrayrulecolor{black}
MoE Expert Parallelism Load Balancing &
  Design an efficient MoE expert-replica placement algorithm that minimizes GPU/node load imbalance while preserving inter-node locality and low runtime. &
  \texttt{\seqsplit{deepseek-ai}}/\allowbreak \texttt{\seqsplit{eplb}} &
  Greedy \newline Zigzag \newline Flat Zigzag &
  DeepSeek-V3 \newline Qwen3-MoE \newline DeepSeek-V2 \newline Stress-Skew \\
\arrayrulecolor{black!15}\hline\arrayrulecolor{black}
Long-Context Inference-Time Sparse Attention &
  Design an inference-time sparse attention module for a pretrained instruction-tuned causal LLM that preserves NIAH and LongBench quality under a 25\% density budget without retraining. &
  custom &
  Dense \newline StreamingLLM \newline BigBird \newline Block Top-K &
  NIAH (8K) \newline LongBench Qasper \newline LongBench MultiFieldQA-EN \\

\midrule
\multicolumn{5}{l}{\textbf{AI for Science (\textsc{Sci})}} \\
\midrule
Mutation Fitness Predictor &
  Studies how mutant and wild-type protein representations can predict functional effects of sequence mutations. &
  \texttt{\seqsplit{OATML-Markslab}}/\allowbreak \texttt{\seqsplit{ProteinGym}} &
  Ridge Regression \newline MLP \newline Reshape CNN &
  BLAT\_ECOLX \newline ESTA\_BACSU \newline RASH\_HUMAN \\
\arrayrulecolor{black!15}\hline\arrayrulecolor{black}
Backbone-to-Sequence Inverse Folding &
  Studies how geometric structure encoding and sequence decoding recover amino-acid sequences from protein backbones. &
  \texttt{\seqsplit{A4Bio}}/\allowbreak \texttt{\seqsplit{ProteinInvBench}} &
  ProteinMPNN \newline PiFold \newline GVP &
  CATH 4.2 \newline CATH 4.3 \newline TS50 \\
\arrayrulecolor{black!15}\hline\arrayrulecolor{black}
Geometric Protein Structure Encoder &
  Studies how local and global geometric protein representations transfer to structure-aware function prediction. &
  \texttt{\seqsplit{a-r-j}}/\allowbreak \texttt{\seqsplit{ProteinWorkshop}} &
  SchNet \newline EGNN \newline GearNet &
  EC \newline GO-BP \newline Fold \\
\arrayrulecolor{black!15}\hline\arrayrulecolor{black}
Atmospheric Column Emulator Architecture &
  Studies how neural emulator architecture maps vertical atmospheric states to sub-grid physics tendencies across training budgets. &
  \texttt{\seqsplit{leap-stc}}/\allowbreak \texttt{\seqsplit{ClimSim}} &
  CNN \newline Encoder-Decoder \newline U-Net \newline HSR &
  Short Budget \newline Medium Budget \newline Long Budget \\
\arrayrulecolor{black!15}\hline\arrayrulecolor{black}
Diffusion-Prior Inverse Solver &
  Studies how diffusion priors and measurement guidance can be combined for inverse-problem reconstruction. &
  \texttt{\seqsplit{devzhk}}/\allowbreak \texttt{\seqsplit{InverseBench}} &
  DPS \newline REDDiff \newline LGD &
  Inverse Scattering \newline Black Hole Imaging \newline Inpainting \\
\arrayrulecolor{black!15}\hline\arrayrulecolor{black}
Molecular Representation Predictor &
  Studies how molecular graph and geometric representations improve property prediction under scaffold-based generalization. &
  \texttt{\seqsplit{deepmodeling}}/\allowbreak \texttt{\seqsplit{Uni-Mol}} &
  D-MPNN \newline Uni-Mol \newline GIN &
  BBBP \newline BACE \newline Tox21 \\
\arrayrulecolor{black!15}\hline\arrayrulecolor{black}
Protein-Ligand Interaction Model &
  Studies how intra- and inter-molecular geometric interactions should be represented to predict binding affinity. &
  \texttt{\seqsplit{guaguabujianle}}/\allowbreak \texttt{\seqsplit{EHIGN\_PLA}} &
  EHIGN \newline GIGN \newline SchNet \newline EGNN &
  PDBbind 2013 \newline PDBbind 2016 \newline PDBbind 2019 \\
\arrayrulecolor{black!15}\hline\arrayrulecolor{black}
Contrastive Virtual-Screening Objective &
  Studies how projection geometry and contrastive losses affect zero-shot protein-ligand screening quality. &
  \texttt{\seqsplit{jianhuiwemi}}/\allowbreak \texttt{\seqsplit{HypSeek}} &
  Vanilla CLIP \newline HCC \newline HCC + Hyperbolic Cone &
  HypSeek Training \newline DUD-E \newline LIT-PCBA \newline DEKOIS 2.0 \\
\arrayrulecolor{black!15}\hline\arrayrulecolor{black}
Weather Forecast Variable Aggregation &
  Studies how weather forecasting models aggregate information across heterogeneous meteorological variables for optimal prediction. &
  \texttt{\seqsplit{microsoft}}/\allowbreak \texttt{\seqsplit{ClimaX}} &
  Cross-Attention \newline Mean Pooling \newline Learned Weighted Sum &
  Z500 3-Day \newline T850 5-Day \newline 10m-Wind 7-Day \\
\arrayrulecolor{black!15}\hline\arrayrulecolor{black}
Industrial CFD Design: Custom Neural Operator Design &
  Designs and implements a custom neural operator for industrial aerodynamic design prediction on 3D unstructured point clouds. &
  \texttt{\seqsplit{thuml}}/\allowbreak \texttt{\seqsplit{Neural-Solver-Library}} &
  PointNet \newline GraphSAGE \newline Graph U-Net \newline Transolver &
  Car Design \newline AirfRANS \newline Aircraft Design \\

\midrule
\multicolumn{5}{l}{\textbf{Optimization \& Theory (\textsc{Opt})}} \\
\midrule
Optimization Bilevel &
  Studies a fixed bilevel-optimization benchmark based on Shen and Chen's penalty-based bilevel gradient descent experiments, selecting supported methods and tuning paper-style strategy hyperparameters. &
  \texttt{\seqsplit{hanshen95}}/\allowbreak \texttt{\seqsplit{penalized-bilevel-gradient-descent}} &
  V-PBGD \newline G-PBGD \newline RHG \newline T-RHG &
  Toy Convergence \newline HyperClean (Linear) \newline HyperClean (MLP) \\
\arrayrulecolor{black!15}\hline\arrayrulecolor{black}
RAIN Convex-Concave &
  Studies gradient-norm convergence on the exact convex-concave benchmark instances used by the official RAIN bilinear and delta-function scripts. &
  \texttt{\seqsplit{TrueNobility303}}/\allowbreak \texttt{\seqsplit{RAIN}} &
  SEG \newline R-SEG \newline SEAG \newline RAIN &
  Default Noise \newline Low Noise \newline High Noise \\
\arrayrulecolor{black!15}\hline\arrayrulecolor{black}
Optimizer Design for Diagonal-Net Sparse Recovery &
  Designs an optimizer that recovers a sparse linear predictor from fewer training samples under a diagonal-net parameterization with noisy labels. &
  \texttt{\seqsplit{TrueNobility303}}/\allowbreak \texttt{\seqsplit{RAIN}} &
  SGD \newline AdaGrad \newline Adam \newline Adam (Alt.) &
  d=200, k=5, s=0.1 \newline d=500, k=10, s=0.1 \newline d=500, k=10, s=0.2 \newline d=10000, k=50 \\
\arrayrulecolor{black!15}\hline\arrayrulecolor{black}
Differentially Private SGD: Privacy-Utility Optimization &
  Design an improved DP-SGD variant that achieves higher test accuracy under the same (epsilon, delta)-differential privacy budget. &
  custom &
  Standard DP-SGD \newline Automatic Clipping (AUTO-S) \newline Adaptive Quantile Clipping \newline Step-Decay Noise Schedule &
  MNIST \newline Fashion-MNIST \newline CIFAR-10 \\
\arrayrulecolor{black!15}\hline\arrayrulecolor{black}
Evolutionary Optimization Strategy Design &
  Design a novel combination of selection, crossover, mutation operators and/or evolutionary loop for continuous black-box optimization across multiple benchmark functions. &
  \texttt{\seqsplit{DEAP}}/\allowbreak \texttt{\seqsplit{deap}} &
  GA (SBX) \newline CMA-ES \newline Differential Evolution \newline L-SHADE &
  Rastrigin (30D) \newline Rosenbrock (30D) \newline Ackley (30D) \newline Rastrigin (100D) \\
\arrayrulecolor{black!15}\hline\arrayrulecolor{black}
Gradient Compression for Communication-Efficient Distributed Training &
  Design a gradient compression operator that reduces communication cost in distributed training while maintaining convergence quality. &
  custom &
  TopK Sparsification with Error Feedback \newline QSGD (Quantized SGD) \newline SignSGD &
  ResNet-20 / CIFAR-10 \newline VGG-11-BN / CIFAR-100 \newline ResNet-56 / CIFAR-10 \\
\arrayrulecolor{black!15}\hline\arrayrulecolor{black}
Hyperparameter Optimization: Custom Search Strategy Design &
  Design a custom HPO strategy that improves final validation score and convergence under limited multi-fidelity evaluation budgets. &
  custom &
  Random Search \newline TPE \newline Hyperband \newline DEHB \newline BOHB \newline Optuna CMA-ES &
  XGBoost \newline SVM \newline Neural Net \\
\arrayrulecolor{black!15}\hline\arrayrulecolor{black}
Multi-Objective Optimization: Custom Evolutionary Strategy Design &
  Design a custom multi-objective evolutionary strategy that improves convergence, diversity, and spread on standard benchmark problems. &
  \texttt{\seqsplit{DEAP}}/\allowbreak \texttt{\seqsplit{deap}} &
  NSGA-II \newline MOEA/D \newline SPEA2 \newline NSGA-III \newline RVEA \newline AGE-MOEA &
  ZDT1 \newline ZDT3 \newline DTLZ2 \newline DTLZ1 \\
\arrayrulecolor{black!15}\hline\arrayrulecolor{black}
Sample-Efficient Neural Architecture Search &
  Design and implement a sample-efficient NAS optimizer that discovers high-performing architectures in the NAS-Bench-201 search space under a strict query budget. &
  \texttt{\seqsplit{automl}}/\allowbreak \texttt{\seqsplit{naslib}} &
  Random Search \newline REA \newline BANANAS &
  CIFAR-10 \newline CIFAR-100 \newline ImageNet16-120 \\
\arrayrulecolor{black!15}\hline\arrayrulecolor{black}
Online Bandits: Exploration-Exploitation Strategy Design &
  Design and implement a bandit policy that minimizes cumulative regret across diverse multi-armed bandit settings. &
  \texttt{\seqsplit{SMPyBandits}}/\allowbreak \texttt{\seqsplit{SMPyBandits}} &
  UCB1 \newline Thompson Sampling \newline KL-UCB &
  Stochastic MAB \newline Contextual Bandit \newline Non-Stationary Bandit \\
\arrayrulecolor{black!15}\hline\arrayrulecolor{black}
PAC-Bayes Generalization Bound Optimization &
  Design a tighter PAC-Bayes generalization bound by optimizing the bound formulation, prior/posterior parameterization, and KL divergence estimation for stochastic neural networks. &
  \texttt{\seqsplit{mperezortiz}}/\allowbreak \texttt{\seqsplit{PBB}} &
  McAllester \newline Catoni \newline Quadratic &
  MNIST (FCN) \newline MNIST (CNN) \newline FashionMNIST (CNN) \\
\arrayrulecolor{black!15}\hline\arrayrulecolor{black}
Optimization Parity &
  Improve a fixed two-layer MLP's ability to learn sparse parity by designing only its initialization, training dataset, and AdamW hyperparameters. &
  \texttt{\seqsplit{pytorch}}/\allowbreak \texttt{\seqsplit{examples}} &
  Default \newline Multi-Epoch \newline No Weight Decay &
  n=32, k=8 \newline n=50, k=8 \newline n=64, k=8 \\
\arrayrulecolor{black!15}\hline\arrayrulecolor{black}
Variance Reduction for Stochastic Optimization &
  Design an improved variance reduction strategy for stochastic gradient descent on finite-sum optimization problems. &
  custom &
  SVRG \newline STORM \newline STORM+ &
  Logistic Regression \newline MLP \newline Ill-Conditioned \\

\midrule
\multicolumn{5}{l}{\textbf{Classical \& Adaptive Learning (\textsc{CAL})}} \\
\midrule
Few-Shot Image Classification Method &
  Studies how support encoding, query comparison, and loss design affect episodic few-shot image-classification accuracy. &
  \texttt{\seqsplit{sicara}}/\allowbreak \texttt{\seqsplit{easy-few-shot-learning}} &
  ProtoNet \newline MatchingNet \newline RelationNet &
  Mini-ImageNet 5w-5s \newline CIFAR-FS \newline CUB \\
\arrayrulecolor{black!15}\hline\arrayrulecolor{black}
Meta-Learning Inner-Loop Optimizer &
  Studies how differentiable inner-loop adaptation rules affect few-shot classification accuracy in gradient-based meta-learning. &
  \texttt{\seqsplit{learnables}}/\allowbreak \texttt{\seqsplit{learn2learn}} &
  MAML \newline Meta-SGD \newline ANIL &
  Mini-ImageNet 5w-1s \newline Mini-ImageNet 5w-5s \newline CIFAR-FS 5w-5s \\
\arrayrulecolor{black!15}\hline\arrayrulecolor{black}
Pool-Based Active Learning Query Strategy &
  Studies how unlabeled-sample query rules affect accuracy under a fixed labeling budget. &
  \texttt{\seqsplit{JordanAsh}}/\allowbreak \texttt{\seqsplit{badge}} &
  BADGE \newline BAIT \newline BALD \newline Least Confidence \newline Random &
  Letter \newline Spambase \newline Splice \\
\arrayrulecolor{black!15}\hline\arrayrulecolor{black}
Unsupervised Tabular Anomaly Detector &
  Studies how unlabeled anomaly scoring algorithms identify outliers across tabular data distributions. &
  custom &
  IF (Isolation Forest) \newline LOF \newline OCSVM \newline ECOD \newline COPOD &
  Cardio \newline Thyroid \newline Satellite \newline Shuttle \\
\arrayrulecolor{black!15}\hline\arrayrulecolor{black}
Post-Hoc Probability Calibration Mapping &
  Studies how post-hoc probability transforms improve classifier confidence calibration. &
  custom &
  Platt \newline Temperature Scaling \newline Isotonic Regression &
  RF / MNIST \newline MLP / Fashion-MNIST \newline GBM / Madelon \newline SVM / Breast Cancer \\
\arrayrulecolor{black!15}\hline\arrayrulecolor{black}
Geometry-Robust Clustering Algorithm &
  Studies how clustering objectives and distance metrics handle convex blobs, non-convex moons, and high-dimensional digit data. &
  custom &
  K-Means \newline DBSCAN \newline HDBSCAN &
  Blobs \newline Moons \newline Digits \\
\arrayrulecolor{black!15}\hline\arrayrulecolor{black}
Continual Learning Importance Regularizer &
  Changes parameter-importance estimation and regularization loss to reduce catastrophic forgetting and improve final average accuracy across contexts. &
  \texttt{\seqsplit{GMvandeVen}}/\allowbreak \texttt{\seqsplit{continual-learning}} &
  EWC \newline SI \newline Online EWC &
  Split-MNIST \newline Permuted-MNIST \newline Split-CIFAR100 \\
\arrayrulecolor{black!15}\hline\arrayrulecolor{black}
Nonlinear 2D Structure-Preserving Embedding &
  Studies how nonlinear dimensionality reduction preserves neighborhood structure in low-dimensional embeddings. &
  custom &
  PCA \newline t-SNE \newline UMAP \newline TriMap \newline PaCMAP &
  MNIST \newline Fashion-MNIST \newline 20 Newsgroups \\
\arrayrulecolor{black!15}\hline\arrayrulecolor{black}
Adaptive Boosting Weight and Target Strategy &
  Studies how pseudo-targets, learner weights, and sample reweighting affect boosted ensemble performance. &
  custom &
  AdaBoost \newline Gradient Boosting \newline XGBoost-style &
  Breast Cancer \newline Diabetes \newline California Housing \\
\arrayrulecolor{black!15}\hline\arrayrulecolor{black}
Heterogeneous Federated Server Aggregation &
  Changes server-side client selection and model aggregation to improve federated test accuracy under heterogeneous client data. &
  \texttt{\seqsplit{adap}}/\allowbreak \texttt{\seqsplit{flower}} &
  FedAvg \newline FedProx \newline SCAFFOLD &
  CIFAR-10 (Non-IID alpha=0.1) \newline FEMNIST \newline Shakespeare \\
\arrayrulecolor{black!15}\hline\arrayrulecolor{black}
Correlation-Aware Tabular Imputation &
  Studies how feature correlations and predictive structure guide missing-value imputation in tabular data. &
  custom &
  Mean Imputation \newline KNN Imputation \newline MICE \newline MissForest \newline GAIN &
  Breast Cancer Wisconsin \newline Wine \newline California Housing \\
\arrayrulecolor{black!15}\hline\arrayrulecolor{black}
Selective Deferral Under Subgroup Shift &
  Studies how acceptance and deferral rules trade off selective risk, subgroup robustness, and coverage on AIF360 tabular datasets. &
  custom &
  Confidence Thresholding \newline Conformal Abstention \newline Learned Deferral \newline Group-wise Thresholding &
  Adult \newline COMPAS \newline Law School GPA \\
\arrayrulecolor{black!15}\hline\arrayrulecolor{black}
Shift-Robust Subgroup Calibration &
  Studies how post-hoc calibration behaves under subgroup distribution shift and worst-group reliability constraints on AIF360 tabular datasets. &
  custom &
  Temperature Scaling \newline Isotonic Regression \newline Beta Calibration \newline Group-wise Temperature Scaling &
  Adult \newline COMPAS \newline Law School GPA \\
\arrayrulecolor{black!15}\hline\arrayrulecolor{black}
Genetic Programming Search for Symbolic Regression &
  Studies how symbolic-regression search strategies recover generalizable analytical expressions. &
  \texttt{\seqsplit{trevorstephens}}/\allowbreak \texttt{\seqsplit{gplearn}} &
  Standard GP \newline Parsimony GP \newline Lexicase GP &
  Nguyen-7 \newline Nguyen-10 \newline Koza-3 \\

\midrule
\multicolumn{5}{l}{\textbf{Deep Learning (\textsc{DL})}} \\
\midrule
Adaptive Classification Loss &
  Modify the training loss over logits and labels to improve classification accuracy across image-model families. &
  custom &
  Label Smoothing \newline Focal Loss \newline PolyLoss &
  ResNet-56 / CIFAR-100 \newline VGG-16-BN / CIFAR-100 \newline MobileNet-V2 / Fashion-MNIST \\
\arrayrulecolor{black!15}\hline\arrayrulecolor{black}
Image Augmentation Policy &
  Design the training transform pipeline combining geometric, photometric, and erasing operations to improve image-classification generalization. &
  custom &
  Cutout \newline RandAugment \newline TrivialAugmentWide &
  ResNet-20 / CIFAR-10 \newline ResNet-56 / CIFAR-100 \newline MobileNet-V2 / Fashion-MNIST \\
\arrayrulecolor{black!15}\hline\arrayrulecolor{black}
Hierarchical Classification Loss Weighting &
  Studies how fine-label and coarse-label objectives should be combined to improve hierarchical image classification. &
  custom &
  Uncertainty Weighting \newline DWA \newline PCGrad &
  ResNet-20 / CIFAR-100-MT \newline ResNet-56 / CIFAR-100-MT \newline VGG-16-BN / CIFAR-100-MT \\
\arrayrulecolor{black!15}\hline\arrayrulecolor{black}
Spatial Feature Aggregation &
  Studies how global spatial features should be aggregated to improve image-classification accuracy across convolutional architectures. &
  custom &
  Global Max \newline GeM \newline Avg + Max &
  ResNet-56 / CIFAR-100 \newline VGG-16-BN / CIFAR-100 \newline MobileNet-V2 / Fashion-MNIST \\
\arrayrulecolor{black!15}\hline\arrayrulecolor{black}
Long-Tail Class Reweighting &
  Studies how class-count statistics should be mapped to loss weights to improve test accuracy on balanced test sets for long-tailed image classification. &
  custom &
  Inverse Frequency \newline Class-Balanced (Effective Number) \newline Balanced Softmax &
  ResNet-32 / CIFAR-10-LT \newline ResNet-32 / CIFAR-100-LT \newline VGG-16-BN / CIFAR-100-LT \\
\arrayrulecolor{black!15}\hline\arrayrulecolor{black}
Convolutional Activation Nonlinearity &
  Studies how drop-in activation functions affect accuracy across convolutional image classifiers. &
  custom &
  GELU \newline SiLU \newline Mish &
  ResNet-20 / CIFAR-10 \newline VGG-16-BN / CIFAR-100 \newline MobileNet-V2 / Fashion-MNIST \\
\arrayrulecolor{black!15}\hline\arrayrulecolor{black}
Architecture-Aware Learning-Rate Scheduling &
  Designs an epoch-level learning-rate curve conditioned on architecture and dataset to improve convergence and final classification accuracy. &
  custom &
  Cosine \newline WarmupCosine \newline OneCycle &
  ResNet-20 / CIFAR-10 \newline ResNet-56 / CIFAR-100 \newline MobileNet-V2 / Fashion-MNIST \\
\arrayrulecolor{black!15}\hline\arrayrulecolor{black}
Normalization Statistics and Affine Design &
  Studies how normalization statistics and affine behavior affect convolutional training stability and test accuracy. &
  custom &
  GroupNorm \newline Batch-Instance Norm \newline Switchable Norm &
  ResNet-56 / CIFAR-100 \newline ResNet-110 / CIFAR-100 \newline MobileNet-V2 / Fashion-MNIST \\
\arrayrulecolor{black!15}\hline\arrayrulecolor{black}
Adaptive Regularization Loss &
  Adds a model-, output-, input-, or epoch-dependent regularization term to improve classification generalization beyond standard weight decay. &
  custom &
  DropBlock \newline Confidence Penalty \newline Orthogonal Regularization &
  ResNet-56 / CIFAR-100 \newline VGG-16-BN / CIFAR-100 \newline MobileNet-V2 / Fashion-MNIST \\
\arrayrulecolor{black!15}\hline\arrayrulecolor{black}
Residual Block Skip Design &
  Studies how shortcut transformations and residual branch computation affect optimization and generalization across network depths. &
  custom &
  Pre-Activation \newline Gated Residual \newline Stochastic Depth &
  ResNet-20 / CIFAR-10 \newline ResNet-56 / CIFAR-100 \newline ResNet-110 / CIFAR-100 \\
\arrayrulecolor{black!15}\hline\arrayrulecolor{black}
DL Weight Initialization Strategy Design &
  Designs data-independent initialization for convolutional, normalization, and classifier layers to improve convergence and final accuracy. &
  custom &
  Kaiming Normal \newline Fixup \newline Orthogonal &
  ResNet-56 / CIFAR-100 \newline VGG-16-BN / CIFAR-100 \newline MobileNet-V2 / Fashion-MNIST \\

\midrule
\multicolumn{5}{l}{\textbf{Time Series \& Forecasting (\textsc{TS})}} \\
\midrule
Concept-Drift-Aware Quantitative Forecasting &
  The stock prediction model and data pipeline are redesigned to handle temporal distribution shift and improve signal quality and portfolio metrics. &
  \texttt{\seqsplit{microsoft}}/\allowbreak \texttt{\seqsplit{qlib}} &
  TRA \newline AdaRNN \newline LightGBM &
  CSI 300 \newline CSI 300 (Shifted) \newline CSI 300 (Recent) \\
\arrayrulecolor{black!15}\hline\arrayrulecolor{black}
Graph-Based Quantitative Forecasting &
  Studies how inter-asset graph relationships affect return signal quality and portfolio performance. &
  \texttt{\seqsplit{microsoft}}/\allowbreak \texttt{\seqsplit{qlib}} &
  HIST \newline GATs \newline LightGBM &
  CSI 300 \newline CSI 100 \newline CSI 300 (Recent) \\
\arrayrulecolor{black!15}\hline\arrayrulecolor{black}
Quantitative Return Forecasting &
  Studies how predictive models and input processing affect next-period return signals and portfolio performance. &
  \texttt{\seqsplit{microsoft}}/\allowbreak \texttt{\seqsplit{qlib}} &
  LightGBM \newline LSTM \newline Transformer &
  CSI 300 \newline CSI 100 \newline CSI 300 (Recent) \\
\arrayrulecolor{black!15}\hline\arrayrulecolor{black}
Spatial-Temporal Traffic Forecasting Model &
  Studies how spatial-temporal models capture sensor-network dependencies for traffic forecasting. &
  \texttt{\seqsplit{GestaltCogTeam}}/\allowbreak \texttt{\seqsplit{BasicTS}} &
  STID \newline DLinear \newline StemGNN \newline iTransformer \newline TimesNet \newline SOFTS \newline TimeMixer &
  METR-LA \newline PEMS-BAY \newline PEMS04 \\
\arrayrulecolor{black!15}\hline\arrayrulecolor{black}
Reconstruction Model for Time-Series Anomaly Detection &
  An unsupervised reconstruction model detects anomalous multivariate time-series segments to improve F-score. &
  \texttt{\seqsplit{thuml}}/\allowbreak \texttt{\seqsplit{Time-Series-Library}} &
  DLinear \newline TimesNet \newline PatchTST &
  PSM \newline MSL \newline SMAP \\
\arrayrulecolor{black!15}\hline\arrayrulecolor{black}
Multivariate Time-Series Classification Model &
  Studies how representation learning improves classification of multivariate time-series signals. &
  \texttt{\seqsplit{thuml}}/\allowbreak \texttt{\seqsplit{Time-Series-Library}} &
  DLinear \newline TimesNet \newline PatchTST &
  EthanolConcentration \newline FaceDetection \newline Handwriting \\
\arrayrulecolor{black!15}\hline\arrayrulecolor{black}
Exogenous-Variable Target Forecasting Model &
  Studies how exogenous variables improve target-channel forecasting. &
  \texttt{\seqsplit{thuml}}/\allowbreak \texttt{\seqsplit{Time-Series-Library}} &
  DLinear \newline PatchTST \newline iTransformer \newline TimeXer &
  ETTh1 \newline Weather \newline ECL \\
\arrayrulecolor{black!15}\hline\arrayrulecolor{black}
Masked Multivariate Time-Series Imputation &
  Studies how imputation models reconstruct missing regions in multivariate time series. &
  \texttt{\seqsplit{thuml}}/\allowbreak \texttt{\seqsplit{Time-Series-Library}} &
  DLinear \newline TimesNet \newline PatchTST &
  ETTh1 (25\% missing) \newline Weather (25\% missing) \newline ECL (25\% missing) \\
\arrayrulecolor{black!15}\hline\arrayrulecolor{black}
Multivariate Long-Horizon Forecasting Model &
  Studies how long-horizon forecasting models predict future multivariate sequences. &
  \texttt{\seqsplit{thuml}}/\allowbreak \texttt{\seqsplit{Time-Series-Library}} &
  DLinear \newline PatchTST \newline iTransformer \newline TimeMixer \newline TimeXer &
  ETTh1 \newline Weather \newline ECL \\
\arrayrulecolor{black!15}\hline\arrayrulecolor{black}
Univariate Short-Horizon Forecasting Model &
  Studies how short-horizon forecasting models predict seasonal univariate series. &
  \texttt{\seqsplit{thuml}}/\allowbreak \texttt{\seqsplit{Time-Series-Library}} &
  DLinear \newline TimesNet \newline PatchTST \newline TimeMixer &
  M4 Monthly \newline M4 Quarterly \newline M4 Yearly \\

\midrule
\multicolumn{5}{l}{\textbf{Structured \& Causal Reasoning (\textsc{SCR})}} \\
\midrule
Discrete Causal Graph Discovery &
  Studies how causal discovery algorithms recover equivalence-class graph structure from discrete observational data. &
  \texttt{\seqsplit{py-why}}/\allowbreak \texttt{\seqsplit{causal-learn}} &
  PC \newline GES \newline GRaSP \newline BOSS \newline Hill Climbing &
  Cancer \newline Child \newline ALARM \newline HAILFINDER \newline Win95pts \\
\arrayrulecolor{black!15}\hline\arrayrulecolor{black}
Linear Gaussian Causal Discovery &
  Studies how observational algorithms recover causal graph structure under linear Gaussian assumptions. &
  \texttt{\seqsplit{py-why}}/\allowbreak \texttt{\seqsplit{causal-learn}} &
  PC \newline GRaSP \newline BOSS &
  ER (n=10) \newline ER (n=20) \newline SF (n=50) \newline SF (n=50, Hard) \newline ER (n=20, Noisy) \\
\arrayrulecolor{black!15}\hline\arrayrulecolor{black}
Non-Gaussian Causal Discovery &
  Studies how non-Gaussian structure can identify directed causal relationships from observational data. &
  \texttt{\seqsplit{py-why}}/\allowbreak \texttt{\seqsplit{causal-learn}} &
  ICA-LiNGAM \newline DirectLiNGAM \newline NOTEARS &
  ER (n=30) \newline ER (n=50) \newline SF (n=100) \\
\arrayrulecolor{black!15}\hline\arrayrulecolor{black}
Nonlinear Causal Discovery &
  Studies how nonlinear additive-noise assumptions support directed causal graph recovery from observations. &
  \texttt{\seqsplit{py-why}}/\allowbreak \texttt{\seqsplit{causal-learn}} &
  CAM \newline NOTEARS-MLP \newline DirectLiNGAM \newline GraN-DAG &
  SF (n=20, GP) \newline ER (n=20, Gauss) \newline ER (n=12, Low-Sample) \\
\arrayrulecolor{black!15}\hline\arrayrulecolor{black}
Heterogeneous Treatment Effect Estimation &
  Studies how observational estimators recover individual and average treatment effects on synthetic CATE benchmark families. &
  custom &
  S-Learner \newline T-Learner \newline IPW \newline Causal Forest \newline DR-Learner \newline R-Learner &
  IHDP-inspired Synth \newline Jobs/LaLonde-inspired Synth \newline ACIC-inspired Synth \\
\arrayrulecolor{black!15}\hline\arrayrulecolor{black}
Unconditional Graph Generator Architecture &
  Studies how graph generator architecture affects distributional match to target graph statistics. &
  \texttt{\seqsplit{pyg-team}}/\allowbreak \texttt{\seqsplit{pytorch\_geometric}} &
  GraphVAE \newline GRAN \newline DiGress &
  Community-Small \newline Ego-Small \newline ENZYMES \\
\arrayrulecolor{black!15}\hline\arrayrulecolor{black}
Structure-Aware Graph Readout Pooling &
  Studies how graph-level readout mechanisms affect graph classification accuracy and macro F1 under a fixed message-passing backbone. &
  \texttt{\seqsplit{pyg-team}}/\allowbreak \texttt{\seqsplit{pytorch\_geometric}} &
  GIN + Sum \newline SAGPool \newline DiffPool &
  MUTAG \newline PROTEINS \newline NCI1 \\
\arrayrulecolor{black!15}\hline\arrayrulecolor{black}
Graph Link Encoder-Decoder &
  Studies how node encoders and edge decoders affect missing-link prediction quality. &
  custom &
  GCN + MLP Decoder \newline VGAE \newline SEAL &
  Cora \newline CiteSeer \newline ogbl-collab \\
\arrayrulecolor{black!15}\hline\arrayrulecolor{black}
Graph Node Message Passing &
  Studies how message-passing layers affect node classification across citation network benchmarks. &
  \texttt{\seqsplit{pyg-team}}/\allowbreak \texttt{\seqsplit{pytorch\_geometric}} &
  GCN \newline GAT \newline GraphSAGE &
  Cora \newline CiteSeer \newline PubMed \\
\arrayrulecolor{black!15}\hline\arrayrulecolor{black}
Homophily-Heterophily Graph Filter &
  The graph signal propagation filter is changed to improve node classification accuracy across homophilic and heterophilic graphs. &
  \texttt{\seqsplit{ivam-he}}/\allowbreak \texttt{\seqsplit{ChebNetII}} &
  GPR-GNN \newline BernNet \newline ChebNetII &
  Cora \newline CiteSeer \newline Texas \newline Cornell \\

\midrule
\multicolumn{5}{l}{\textbf{Trustworthy Learning (\textsc{TL})}} \\
\midrule
Score-Based Black-Box Linf Attack &
  Designs a query-efficient black-box Linf evasion attack to improve attack success rate under a fixed per-sample query budget. &
  \texttt{\seqsplit{Harry24k}}/\allowbreak \texttt{\seqsplit{adversarial-attacks-pytorch}} &
  Square Attack \newline SPSA \newline Random Search &
  ResNet-20 / CIFAR-10 \newline VGG-11-BN / CIFAR-10 \newline MobileNet-V2 / CIFAR-10 \newline ResNet-20 / CIFAR-100 \newline MobileNet-V2 / CIFAR-100 \\
\arrayrulecolor{black!15}\hline\arrayrulecolor{black}
Sparse L0 Adversarial Attack &
  Studies how sparse perturbation strategies improve attack success while respecting a strict pixel budget. &
  \texttt{\seqsplit{Harry24k}}/\allowbreak \texttt{\seqsplit{adversarial-attacks-pytorch}} &
  OnePixel \newline SparseFool \newline JSMA \newline Pixle \newline Sparse-RS &
  Rebuffi-R18 (l2-AT) / CIFAR-10 \newline Augustin (l2-robust) / CIFAR-10 \newline Engstrom (l2-robust) / CIFAR-10 \\
\arrayrulecolor{black!15}\hline\arrayrulecolor{black}
White-Box Linf Evasion Attack &
  Designs a gradient-based white-box Linf attack to improve attack success rate while respecting the perturbation budget. &
  \texttt{\seqsplit{Harry24k}}/\allowbreak \texttt{\seqsplit{adversarial-attacks-pytorch}} &
  FGSM \newline PGD \newline MI-FGSM \newline AutoAttack &
  ResNet-20 / CIFAR-10 \newline VGG-11-BN / CIFAR-10 \newline ResNet-20 / CIFAR-100 \newline VGG-11-BN / CIFAR-100 \newline MobileNet-V2 / CIFAR-100 \\
\arrayrulecolor{black!15}\hline\arrayrulecolor{black}
Linf Adversarial Training for Robust Accuracy &
  Studies how adversarial training procedures improve robust accuracy while maintaining clean accuracy. &
  \texttt{\seqsplit{Harry24k}}/\allowbreak \texttt{\seqsplit{adversarial-attacks-pytorch}} &
  Standard Training \newline PGD-AT \newline TRADES \newline MART \newline AWP + TRADES &
  SmallCNN / MNIST \newline PreAct ResNet-18 / CIFAR-10 \newline VGG-11-BN / CIFAR-10 \newline PreAct ResNet-18 / CIFAR-100 \\
\arrayrulecolor{black!15}\hline\arrayrulecolor{black}
Poisoned-Sample Scoring for Backdoor Filtering &
  A suspicion scoring rule identifies and filters backdoored training examples to reduce attack success rate while preserving clean accuracy. &
  custom &
  Confidence Filter \newline Spectral Signatures \newline Activation Clustering \newline Z-Score Outlier &
  ResNet-20 / CIFAR-10 (BadNets) \newline VGG-16-BN / CIFAR-100 (Blend) \newline MobileNet-V2 / Fashion-MNIST (BadNets) \\
\arrayrulecolor{black!15}\hline\arrayrulecolor{black}
Targeted Update Rules for Class Unlearning &
  An unlearning update rule removes forget-class information while improving retained accuracy and reducing forget-set membership leakage. &
  custom &
  Retain Fine-Tune \newline Negative Gradient \newline Bad Teacher \newline SCRUB &
  ResNet-20 / CIFAR-10 (Class 0) \newline VGG-16-BN / CIFAR-100 (Class 0) \newline MobileNet-V2 / Fashion-MNIST (Class 0) \\
\arrayrulecolor{black!15}\hline\arrayrulecolor{black}
Training Regularization for Membership Privacy &
  Studies how privacy-preserving training losses reduce membership leakage while maintaining accuracy. &
  custom &
  ERM \newline Label Smoothing \newline Confidence Penalty \newline RelaxLoss &
  ResNet-20 / CIFAR-10 \newline VGG-16-BN / CIFAR-100 \newline MobileNet-V2 / Fashion-MNIST \\
\arrayrulecolor{black!15}\hline\arrayrulecolor{black}
Robust Losses for Label-Flip Poisoning &
  A robust loss or sample-weighting rule improves clean accuracy under label-flip poisoning and reduces poisoned-label memorization. &
  custom &
  Cross-Entropy \newline Generalized Cross-Entropy \newline Symmetric Cross-Entropy \newline Bootstrap &
  ResNet-20 / CIFAR-10 (Label-Flip) \newline VGG-16-BN / CIFAR-100 (Label-Flip) \newline MobileNet-V2 / Fashion-MNIST (Label-Flip) \\

\end{xltabular}
\endgroup


\clearpage
\section{\bench-Lite: 30-Task Subset}
\label{app:lite_tasks}
\noindent
The 30 tasks of \bench-Lite, grouped by the 12 \bench domains, are:

\paragraph{AI for Science.}
Mutation Fitness Predictor; Diffusion-Prior Inverse Solver; Protein-Ligand Interaction Model.

\paragraph{Structured \& Causal Reasoning.}
Discrete Causal Graph Discovery; Unconditional Graph Generator Architecture.

\paragraph{Vision \& Generation.}
3D Scene Densification Strategy; Low-Step Diffusion Bridge Sampling; Frequency-Aware Autoencoding Loss.

\paragraph{Deep Learning.}
Spatial Feature Aggregation; Convolutional Activation Nonlinearity.

\paragraph{Robotics.}
Latent World-Model Planner; Guided Diffusion Sampling for Robot Actions; Diffusion Policy Learning for Robot Control; Humanoid Transfer Policy Learning; Behavioral Cloning Loss for Manipulation.

\paragraph{Language Models.}
Masked Diffusion Demasking Policy; Pretraining Optimizer Design; Reasoning RL Importance-Sampling Granularity.

\paragraph{ML Systems \& Efficient ML.}
Post-Training Weight Quantization; Quantization-Aware Language-Model Training; Long-Context Inference-Time Sparse Attention.

\paragraph{Classical \& Adaptive Learning.}
Geometry-Robust Clustering Algorithm; Nonlinear 2D Structure-Preserving Embedding.

\paragraph{Optimization \& Theory.}
Multi-Objective Evolutionary Survival and Variation; Variance-Reduced Stochastic Optimization.

\paragraph{Time Series \& Forecasting.}
Concept-Drift-Aware Quantitative Forecasting; Exogenous-Variable Target Forecasting Model; Masked Multivariate Time-Series Imputation.

\paragraph{Reinforcement Learning.}
Value-Based Discrete Control.

\paragraph{Trustworthy Learning.}
Training Regularization for Membership Privacy.


\clearpage
\section{Agent Prompts and Tool Schemas}
\label{app:prompts_and_tools}

\subsection{System Prompt}
\label{app:system_prompt}
The default scientific-innovation system prompt used in our main agent runs is reproduced below.
\begin{lstlisting}[style=prompt]
You are an ML scientist. Your goal is to propose and implement a novel algorithmic contribution that improves performance on the given task.

What counts as a good contribution:
- A new loss function or objective formulation
- A new policy update rule or gradient estimation method
- A novel exploration or regularization strategy
- A new way to parameterize or combine components, with clear motivation

What does NOT count:
- Trivially increasing network capacity to brute-force a metric (a per-task parameter cap is enforced before each test)
- Hyperparameter tuning (learning rates, batch sizes, etc.)
- Copying a reference baseline with cosmetic changes
- Pure engineering tricks without algorithmic novelty

Parameter count is capped (enforced before each test); architectural changes within that budget are encouraged.

IMPORTANT workflow:
1. FIRST call edit() to implement your improved algorithm. Do NOT call test() before making edits.
2. THEN call test() to run the experiment. Each run is numbered (#1, #2, ...).
3. Review the metrics, then edit() to improve your solution based on the feedback.
4. Call test() again to verify the improvement. You MUST iterate at least once (edit → test → review → edit → test) before submitting, unless only 1 test is allowed.
5. When satisfied, call submit(n=N) to submit your best test #N as final.
You have a limited number of test() calls, so make each one count by editing first.

Available tools:
- edit(op, filename, content, ...): Modify files in the workspace.
  - op='replace': replace lines start_line..end_line with content
  - op='insert': insert content after after_line
  - op='create': create a new file (only if allow_create=true)
- test(): Run a new experiment. Executes training and evaluation. Each run is
  numbered #1, #2, etc. The first test runs all seeds; intermediate tests run one seed.
  You have a limited budget of test() calls, so make each one count by editing first.
  If max tests is reached, the last test is auto-submitted.
- submit(n=N): Submit the result from test #N as your final answer (1-indexed).
  This does NOT re-run anything — it picks a previous result. Use n=-1 for the latest.
  You must call test() at least once before you can submit.
- undo(n=1): Revert the last n edit operations.

Constraints:
- Each file shown in the prompt is labeled [READ-ONLY] or [EDITABLE — lines X–Y only].
  Only edit files and line ranges marked EDITABLE. Do not touch READ-ONLY files.
- When a file has multiple editable regions, editing one region may shift line numbers in subsequent regions. Edit from the last (bottom-most) region first, or check the updated editable ranges shown after each edit.
- You MUST call test() at least once before you can call submit().
- When you are done, call submit(n=N) to submit your best test result.
- If your algorithm requires new hyperparameters (e.g., cql_alpha, expectile_tau) that are not
  in the existing config, hardcode their values directly in your code (e.g., in __init__).
  You cannot modify the training loop or config to pass them via command line.
\end{lstlisting}
The no-budget ablation replaces the novelty and parameter-budget preamble with the shorter \texttt{SYSTEM\_PROMPT\_SCI\_PREAMBLE\_NOBUDGET} variant.

\subsection{Initial User Prompt}
\label{app:initial_prompt}
The initial user prompt is assembled from task metadata, annotated files, evaluation commands, baseline results, and budget information as follows.
\begin{lstlisting}[style=prompt]
[Optional extra context block]
# <BASELINE_DERIVATIONS_OR_DEEP_THEORETICAL_CONTEXT> (reference material)

<EXTRA_CONTEXT_TEXT>

# Task: <TASK_NAME>

<TASK_DESCRIPTION>

## <FILENAME>  [READ-ONLY — do not edit]
```<LANGUAGE>
<LINE_NUMBERED_READ_RANGE_OR_FULL_FILE_CONTENT>
```

## <FILENAME>  [EDITABLE — lines <START>–<END> only]
```<LANGUAGE>
Lines <START>-<END>:
<LINE_NUMBERED_READ_RANGE>
```

[If allow_hack=true, editability annotations are omitted. If rigorous_codebase=true and baseline variants are available, read-only non-editable files are skipped.]

## <BASELINE_NAME> baseline — editable region  [READ-ONLY — reference implementation]
```python
Lines <START>–<END>:
<LINE_NUMBERED_BASELINE_EDITABLE_REGION_WITH_THREE_LINES_OF_CONTEXT>
```

## Evaluation Commands
Your algorithm is evaluated by running:
  - `<COMMAND>` → label: `<LABEL>`

## Compute Budget
All evaluation runs on **NVIDIA H100 80GB** GPU(s). Your algorithm must complete within the time limits below. If a command exceeds its time limit, the run is killed and the result is **invalid** (it will not count as a valid test result). Design your model to be efficient enough to train and evaluate within these constraints.

| Command | GPUs | Time Limit |
| --- | --- | --- |
| `<LABEL>` | <GPU_DESCRIPTION> | <TIME_LIMIT> |

## Baseline Results
Beat these with your algorithm:

| baseline | <LABEL_1> | <LABEL_2> |
| --- | --- | --- |
| <BASELINE_NAME> | <METRIC_VALUE> | <METRIC_VALUE> |

## Your Budget
- **Action budget**: <MAX_STEPS> total tool calls (every edit / test / undo / web_search / web_extract counts; submit does not)
- **Test invocations**: at most <MAX_TESTS> (each test() call also consumes one action from the budget above)
- You **must** iterate at least once (edit → test → review → edit → test) before submitting.

[If <MAX_TESTS>=1, the iteration line is replaced by:]
- **CRITICAL — single-test mode (max_tests=1)**: your one and only `test()` call is automatically the FINAL submission. Whatever metrics it returns are written to the leaderboard; whatever bugs it hits are recorded as a failed submission with empty metrics — there is **no second chance**.
- Before you call `test()`, run through this checklist mentally: (1) tensor shapes match between your model layers and the expected input/output, (2) dtypes / device are consistent, (3) any new module is actually used in `forward()`, (4) loss is finite on a tiny dummy input, (5) you handled the corner cases the task description warns about.
- If you are unsure, spend remaining edit budget tightening the code rather than rushing to test. **A crashed test is a wasted submission.** Only call `test()` when you can defend each line.
\end{lstlisting}

\subsection{Tool Schemas}
\label{app:tool_schemas}
The core workspace tool schemas below are reproduced for the three tools \texttt{edit}, \texttt{test}, and \texttt{undo}.
\begin{lstlisting}[style=prompt]
{
  "name": "edit",
  "description": "Edit files in the workspace. Three operations are supported:\n  create: Create a new file with the given content. Only available if allow_create=true.\n  insert: Insert one or more lines immediately after `after_line` (1-indexed).\n  replace: Replace lines `start_line`..`end_line` (inclusive, 1-indexed) with `content`.\nFile paths are relative to the package root (e.g. 'LLaMA-Factory/src/...').\nLines within protected ranges must NOT be modified.",
  "input_schema": {
    "type": "object",
    "properties": {
      "op": {
        "type": "string",
        "enum": [
          "create",
          "insert",
          "replace"
        ],
        "description": "The edit operation to perform."
      },
      "filename": {
        "type": "string",
        "description": "Package-relative path to the file (e.g. 'LLaMA-Factory/src/llamafactory/train/dpo/trainer.py')."
      },
      "content": {
        "type": "string",
        "description": "Content to write (for create/replace) or insert."
      },
      "after_line": {
        "type": "integer",
        "description": "Line number after which to insert (required for op='insert')."
      },
      "start_line": {
        "type": "integer",
        "description": "First line to replace, 1-indexed (required for op='replace')."
      },
      "end_line": {
        "type": "integer",
        "description": "Last line to replace, 1-indexed inclusive (required for op='replace')."
      }
    },
    "required": [
      "op",
      "filename",
      "content"
    ]
  }
}
\end{lstlisting}
\begin{lstlisting}[style=prompt]
{
  "name": "test",
  "description": "Run a new experiment. Executes training and evaluation, then returns metrics. Each run is numbered #1, #2, etc. All runs use all configured seeds. You have a limited test budget.",
  "input_schema": {
    "type": "object",
    "properties": {}
  }
}
\end{lstlisting}
\begin{lstlisting}[style=prompt]
{
  "name": "undo",
  "description": "Revert the last n file modification actions (create/insert/replace) by restoring pre-edit snapshots. Does not undo test calls.",
  "input_schema": {
    "type": "object",
    "properties": {
      "n": {
        "type": "integer",
        "description": "Number of edit actions to undo (default: 1)."
      }
    }
  }
}
\end{lstlisting}


\clearpage
\section{Test-Time Scaling Configurations}
\label{app:test_time_scaling_configs}

\paragraph{Sampling and exploration.} The two ReAct-based setups reuse the
default scaffold and per-call sampling defaults of the underlying provider; we
only vary the action budget. \emph{Sampling} runs 16 independent ReAct chains
of at most 5 actions, each ending in a single \texttt{test} call, and reports
the running best across them. \emph{Exploration} runs a single 50-action
ReAct chain that may issue up to 16 \texttt{test} calls, iteratively refining
one solution. All runs use \(\mathrm{seed}=42\) and a single in-process
container runtime.

\subsection*{OpenEvolve (test-time evolution)}
\label{app:openevolve_hparams}

OpenEvolve is run with a 160-LLM-call budget, two calls per iteration
(mutation followed by a judge that contributes a feedback weight to the
score). Table~\ref{tab:openevolve_hparams} lists the full hyperparameter
set.

\begin{small}
\begin{longtable}{@{}p{0.50\textwidth} l@{}}
\caption{OpenEvolve hyperparameters used for the test-time-evolution setup.}
\label{tab:openevolve_hparams}\\
\toprule
\textbf{Parameter} & \textbf{Value} \\
\midrule
\endfirsthead

\multicolumn{2}{l}{\small\itshape (continued from previous page)}\\
\toprule
\textbf{Parameter} & \textbf{Value} \\
\midrule
\endhead

\midrule
\multicolumn{2}{r}{\small\itshape (continued on next page)}\\
\endfoot

\bottomrule
\endlastfoot

\multicolumn{2}{@{}l}{\textit{Budget}} \\
LLM call budget                 & 160 \\
Max iterations                  & 80 \\
Calls per iteration             & 2 (mutation + judge) \\
LLM feedback (judge)            & enabled \\
\addlinespace[3pt]

\multicolumn{2}{@{}l}{\textit{LLM sampling}} \\
Temperature                     & 0.8 \\
Top-$p$                         & 0.95 \\
Max tokens                      & 16\,000 \\
LLM timeout (s)                 & 240 \\
\addlinespace[3pt]

\multicolumn{2}{@{}l}{\textit{Prompt construction}} \\
Top programs in prompt          & 3 \\
Diverse programs in prompt      & 2 \\
Include execution artifacts     & yes \\
Max artifact bytes              & 20\,480 \\
Meta-prompting                  & enabled \\
Meta-prompt weight              & 0.15 \\
\addlinespace[3pt]

\multicolumn{2}{@{}l}{\textit{Database / population}} \\
Population size                 & 60 \\
Archive size                    & 30 \\
Number of islands               & 2 \\
Migration interval (iters)      & 15 \\
Migration rate                  & 0.15 \\
Elite selection ratio           & 0.20 \\
Exploration ratio               & 0.30 \\
Exploitation ratio              & 0.70 \\
Diversity metric                & feature-based \\
Feature dimensions              & score, complexity \\
Feature bins                    & 10 \\
\addlinespace[3pt]

\multicolumn{2}{@{}l}{\textit{Evaluator}} \\
Evaluation timeout (s)          & 3\,600 \\
Parallel evaluations            & 1 \\
LLM feedback weight             & 0.20 \\
Enable artifacts                & yes \\

\end{longtable}
\end{small}

\subsection*{TTT-Discover (test-time training)}
\label{app:ttt_discover_hparams}

TTT-Discover fine-tunes the underlying policy with LoRA-based RL using each
task's evaluator as the reward signal. It is run on \texttt{Qwen3.5-35B-A3B}
(MoE).

\begin{small}
\begin{longtable}{@{}p{0.50\textwidth} l@{}}
\caption{TTT-Discover hyperparameters used in the test-time training setup.}
\label{tab:ttt_discover_hparams}\\
\toprule
\textbf{Parameter} & \textbf{Value} \\
\midrule
\endfirsthead

\multicolumn{2}{l}{\small\itshape (continued from previous page)}\\
\toprule
\textbf{Parameter} & \textbf{Value} \\
\midrule
\endhead

\midrule
\multicolumn{2}{r}{\small\itshape (continued on next page)}\\
\endfoot

\bottomrule
\endlastfoot

\multicolumn{2}{@{}l}{\textit{Model and runtime}} \\
Base model                          & \texttt{Qwen3.5-35B-A3B} (MoE) \\
Container runtime                   & local (in-process) \\
Compute scale                       & 0.5 \\
Seeds                               & \{42\} \\
Reasoning effort                    & high (thinking enabled) \\
Thinking-token budget               & 10\,000 \\
\addlinespace[3pt]

\multicolumn{2}{@{}l}{\textit{RL training loop}} \\
Number of epochs                    & 50 \\
Group size (rollouts per group)     & 4 \\
Groups per batch                    & 1 \\
Phase-1 max tokens                  & 64\,000 \\
KL penalty coefficient              & 0.30 \\
Learning rate                       & $5\times10^{-6}$ \\
LoRA rank                           & 32 \\
Edit-format penalty                 & 0.0 \\
Checkpoint interval                 & every 2 epochs \\

\end{longtable}
\end{small}


\clearpage
\section{Task Subsets for Ablation and Analysis Experiments}
\label{app:partial_experiment_tasks}

\noindent The ablation and analysis experiments in Sections~\ref{sec:ablations} and~\ref{sec:analysis} are each evaluated on a curated subset, chosen so that the relevant property is well defined on every included task.

\paragraph{Scientific innovation vs.\ engineering optimization (Sec.~\ref{sec:ablations}, Figure~\ref{fig:exp_ablations_panel} left).}
Four tasks where the same editable region admits both a scientific-innovation prompt and an engineering-optimization prompt:
\begin{itemize}
\setlength{\itemsep}{1pt}
\item Quantitative Return Forecasting
\item Graph Node Message Passing
\item Multivariate Long-Horizon Forecasting Model
\item Residual Block Skip Design
\end{itemize}

\paragraph{Capacity-budget validity control (Sec.~\ref{sec:ablations}, Figure~\ref{fig:exp_ablations_panel} middle).}
Four computer-vision and reinforcement-learning tasks where the agent can change model size and the budget check is therefore meaningful:
\begin{itemize}
\setlength{\itemsep}{1pt}
\item Value-Based Discrete Control
\item Off-Policy Actor-Critic for Continuous Control
\item Diffusion Model Architecture Design
\item Class-Conditional Diffusion: Conditioning Injection Methods
\end{itemize}

\paragraph{Test-time scaling (Sec.~\ref{sec:test-time-scaling}).}
Six low-latency tasks for the inference-only setups (Scaling Sampling, Scaling Exploration, OpenEvolve):
\begin{itemize}
\setlength{\itemsep}{1pt}
\item Optimization Bilevel
\item Variance Reduction for Stochastic Optimization
\item Long-Tail Class Reweighting
\item Homophily-Heterophily Graph Filter
\item Reconstruction Model for Time-Series Anomaly Detection
\item Heterogeneous Treatment Effect Estimation
\end{itemize}
The TTT-Discover test-time training setup uses the first two for training.

\paragraph{Verifier-limited compute allocation (Sec.~\ref{sec:optimal-compute}).}
Five LLM pretraining tasks:
\begin{itemize}
\setlength{\itemsep}{1pt}
\item Transformer Feed-Forward Block
\item Pretraining Learning-Rate Schedule
\item Subquadratic Attention Mechanism
\item Pretraining Optimizer Design
\item Normalization and Block Layout
\end{itemize}

\paragraph{Context engineering (Sec.~\ref{sec:context-engineering}).}
Nine tasks across optimization, graph, RL, computer vision, and AI-for-science.
\begin{itemize}
\setlength{\itemsep}{1pt}
\item Homophily-Heterophily Graph Filter
\item PAC-Bayes Generalization Bound Optimization
\item Variance Reduction for Stochastic Optimization
\item Genetic Programming Search for Symbolic Regression
\item Diffusion-Prior Inverse Solver
\item Diffusion Model: Sampler Efficiency Optimization
\item Diffusion Prediction Parameterization
\item Intrinsic Exploration for Sparse Rewards
\item Q-Overestimation Suppression for Offline Continuous Control
\end{itemize}


\clearpage
\section{Human Expert Assessment}
\label{app:expert_assessment}

This appendix accompanies the case study in Section~\ref{sec:analysis} with per-task examples. Each task block first restates the research question, then shows the editable region of the template the agent starts from, then shows one strong human baseline, and finally one or more curated agent submissions. Each agent submission is preceded by the model name and a brief description and followed by the domain-expert authors' written assessment. To keep the appendix readable, every listing is restricted to the editable region or a focused excerpt, since full agent submissions can run to several hundred lines.

The colored bars in the left margin of every code listing mark how each line relates to the editable region of the original template:
\tikz{\fill[color=ealEdit] (0,0) rectangle (1.4pt,2.2ex);}~green --- the line was modified by the agent or baseline shown in this listing;
\tikz{\fill[color=ealEditable] (0,0) rectangle (1.4pt,2.2ex);}~blue --- the line is inside the editable region but unchanged from the template;
no bar --- the line is outside the editable region (read-only context shown for orientation).

\subsection{Fused Causal Attention Kernel}
\label{sec:expert:mlsys-fused-attention}

\begin{mdframed}[backgroundcolor=white,linecolor=gray!50,linewidth=0.6pt,roundcorner=2pt,innertopmargin=6pt,innerbottommargin=6pt]
\textbf{Task.} A fused causal self-attention forward pass in OpenAI Triton is evaluated on NVIDIA H100, maximizing throughput (TFLOPs/s) while keeping the maximum absolute error below \(10^{-2}\) against a reference. Editable: the Triton kernel and its Python wrapper. Read-only: the benchmark harness, FLOP accounting, and correctness check. Provided baselines include a naive Triton kernel, a Flash-Attention v2 style two-pass causal kernel, and a Flash-Attention v3 reference.
\end{mdframed}

\paragraph{Template (editable region).}
\begin{lstlisting}[style=eal/python]
(*@\elnE{  29}@*) @triton.jit
(*@\elnE{  30}@*) def _custom_attn_fwd(
(*@\elnE{  31}@*)     Q, K, V, Out,
(*@\elnE{  32}@*)     sm_scale,
(*@\elnE{  33}@*)     stride_qh, stride_qm, stride_qk,
(*@\elnE{  34}@*)     stride_kh, stride_kn, stride_kk,
(*@\elnE{  35}@*)     stride_vh, stride_vn, stride_vk,
(*@\elnE{  36}@*)     stride_oh, stride_om, stride_ok,
(*@\elnE{  37}@*)     seqlen,
(*@\elnE{  38}@*)     BLOCK_M: tl.constexpr,
(*@\elnE{  39}@*)     BLOCK_N: tl.constexpr,
(*@\elnE{  40}@*)     BLOCK_DMODEL: tl.constexpr,
(*@\ealelide{66}@*)
(*@\elnE{ 107}@*)     grid = (triton.cdiv(seqlen, BLOCK_M), batch * nheads)
(*@\elnE{ 108}@*)     _custom_attn_fwd[grid](
(*@\elnE{ 109}@*)         q, k, v, o, sm_scale,
(*@\elnE{ 110}@*)         q.stride(1), q.stride(2), q.stride(3),
(*@\elnE{ 111}@*)         k.stride(1), k.stride(2), k.stride(3),
(*@\elnE{ 112}@*)         v.stride(1), v.stride(2), v.stride(3),
(*@\elnE{ 113}@*)         o.stride(1), o.stride(2), o.stride(3),
(*@\elnE{ 114}@*)         seqlen,
(*@\elnE{ 115}@*)         BLOCK_M=BLOCK_M, BLOCK_N=BLOCK_N,
(*@\elnE{ 116}@*)         BLOCK_DMODEL=headdim, IS_CAUSAL=causal,
(*@\elnE{ 117}@*)     )
(*@\elnE{ 118}@*)     return o
(*@\elnE{ 119}@*)
\end{lstlisting}

\paragraph{Baseline: \texttt{flash\_v3}.}
\begin{lstlisting}[style=eal/python]
(*@\elnU{  27}@*) # ================================================================
(*@\elnU{  28}@*)
(*@\ealregion{editable region begins at line 29}@*)
(*@\elnM{  29}@*) @triton.autotune(
(*@\elnM{  30}@*)     configs=[
(*@\elnM{  31}@*)         triton.Config({'BLOCK_M': 128, 'BLOCK_N': 128}, num_stages=3, num_warps=8),
(*@\elnM{  32}@*)         triton.Config({'BLOCK_M': 128, 'BLOCK_N': 64}, num_stages=3, num_warps=8),
(*@\elnM{  33}@*)         triton.Config({'BLOCK_M': 128, 'BLOCK_N': 64}, num_stages=4, num_warps=8),
(*@\elnM{  34}@*)         triton.Config({'BLOCK_M': 64, 'BLOCK_N': 64}, num_stages=3, num_warps=4),
(*@\elnM{  35}@*)         triton.Config({'BLOCK_M': 64, 'BLOCK_N': 64}, num_stages=4, num_warps=8),
(*@\elnM{  36}@*)         triton.Config({'BLOCK_M': 64, 'BLOCK_N': 128}, num_stages=3, num_warps=8),
(*@\elnM{  37}@*)         triton.Config({'BLOCK_M': 128, 'BLOCK_N': 32}, num_stages=3, num_warps=4),
(*@\elnM{  38}@*)         triton.Config({'BLOCK_M': 64, 'BLOCK_N': 32}, num_stages=4, num_warps=4),
(*@\elnM{  39}@*)     ],
(*@\elnM{  40}@*)     key=['seqlen', 'BLOCK_DMODEL', 'IS_CAUSAL'],
(*@\elnM{  41}@*) )
(*@\elnE{  42}@*) @triton.jit
(*@\elnM{  43}@*) def _flash_v3_fwd(
(*@\elnE{  44}@*)     Q, K, V, Out,
(*@\elnE{  45}@*)     stride_qh, stride_qm, stride_qk,
(*@\elnE{  46}@*)     stride_kh, stride_kn, stride_kk,
(*@\ealelide{5}@*)
(*@\elnE{  52}@*)     BLOCK_DMODEL: tl.constexpr,
(*@\elnE{  53}@*)     IS_CAUSAL: tl.constexpr,
(*@\elnE{  54}@*) ):
(*@\elnM{  55}@*)     """FA3-inspired: autotuned two-pass causal with software pipelining."""
(*@\elnE{  56}@*)     start_m = tl.program_id(0)
(*@\elnE{  57}@*)     off_hz = tl.program_id(1)
(*@\elnE{  58}@*)
(*@\ealelide{6}@*)
(*@\elnE{  65}@*)     offs_n = tl.arange(0, BLOCK_N)
(*@\elnE{  66}@*)     offs_d = tl.arange(0, BLOCK_DMODEL)
(*@\elnE{  67}@*)
(*@\elnM{  68}@*)     # Load Q with scale already fused (done in wrapper)
(*@\elnE{  69}@*)     q_ptrs = Q + q_offset + offs_m[:, None] * stride_qm + offs_d[None, :] * stride_qk
(*@\ealelide{39}@*)
(*@\elnM{ 109}@*)         k = tl.load(k_ptrs, mask=(start_n + offs_n[:, None]) < seqlen, other=0.0)
(*@\elnM{ 110}@*)         qk = tl.dot(q, tl.trans(k))
(*@\elnM{ 111}@*)         qk = tl.where(offs_m[:, None] >= (start_n + offs_n[None, :]), qk, float("-inf"))
(*@\elnM{ 112}@*)         m_ij = tl.max(qk, axis=1)
(*@\elnM{ 113}@*)         m_new = tl.maximum(m_i, m_ij)
(*@\elnM{ 114}@*)         alpha = tl.math.exp2(m_i - m_new)
(*@\elnM{ 115}@*)         p = tl.math.exp2(qk - m_new[:, None])
(*@\elnM{ 116}@*)         l_i = l_i * alpha + tl.sum(p, axis=1)
(*@\elnM{ 117}@*)         acc = acc * alpha[:, None]
(*@\elnM{ 118}@*)         v_ptrs = V + v_offset + (start_n + offs_n[:, None]) * stride_vn + offs_d[None, :] * stride_vk
(*@\elnM{ 119}@*)         v = tl.load(v_ptrs, mask=(start_n + offs_n[:, None]) < seqlen, other=0.0)
(*@\elnM{ 120}@*)         acc += tl.dot(p.to(v.dtype), v)
(*@\elnM{ 121}@*)         m_i = m_new
(*@\elnM{ 122}@*)
(*@\elnE{ 123}@*)     acc = acc / l_i[:, None]
(*@\elnE{ 124}@*)     o_ptrs = Out + o_offset + offs_m[:, None] * stride_om + offs_d[None, :] * stride_ok
(*@\elnE{ 125}@*)     tl.store(o_ptrs, acc.to(Out.dtype.element_ty), mask=offs_m[:, None] < seqlen)
(*@\elnE{ 126}@*)
(*@\elnE{ 127}@*)
(*@\elnE{ 128}@*) def custom_attention_forward(q, k, v, causal=True, sm_scale=None):
(*@\elnM{ 129}@*)     """FA3-inspired: autotuned pipelining + fused scale + two-pass causal."""
(*@\elnE{ 130}@*)     batch, nheads, seqlen, headdim = q.shape
(*@\elnE{ 131}@*)     q, k, v = q.contiguous(), k.contiguous(), v.contiguous()
(*@\elnE{ 132}@*)     if sm_scale is None:
(*@\elnE{ 133}@*)         sm_scale = 1.0 / math.sqrt(headdim)
(*@\elnM{ 134}@*)     # Fuse scale into Q
(*@\elnM{ 135}@*)     q = (q * (sm_scale * 1.44269504)).contiguous()
(*@\elnE{ 136}@*)     o = torch.empty_like(q)
(*@\elnM{ 137}@*)     grid = lambda META: (triton.cdiv(seqlen, META['BLOCK_M']), batch * nheads)
(*@\elnM{ 138}@*)     _flash_v3_fwd[grid](
(*@\elnM{ 139}@*)         q, k, v, o,
(*@\elnE{ 140}@*)         q.stride(1), q.stride(2), q.stride(3),
(*@\elnE{ 141}@*)         k.stride(1), k.stride(2), k.stride(3),
(*@\elnE{ 142}@*)         v.stride(1), v.stride(2), v.stride(3),
\end{lstlisting}

\paragraph{Agent: Claude Opus 4.6.}\mbox{}\\
Causal/non-causal split with H100-tuned autotune configs.
\begin{lstlisting}[style=eal/python]
(*@\elnU{  27}@*) # ================================================================
(*@\elnU{  28}@*)
(*@\ealregion{editable region begins at line 29}@*)
(*@\elnM{  29}@*) @triton.autotune(
(*@\elnM{  30}@*)     configs=[
(*@\elnM{  31}@*)         # Large BLOCK_M configs (good for small headdim like 64)
(*@\elnM{  32}@*)         triton.Config({'BLOCK_M': 256, 'BLOCK_N': 128}, num_stages=3, num_warps=8),
(*@\elnM{  33}@*)         triton.Config({'BLOCK_M': 256, 'BLOCK_N': 64}, num_stages=3, num_warps=8),
(*@\elnM{  34}@*)         triton.Config({'BLOCK_M': 128, 'BLOCK_N': 128}, num_stages=3, num_warps=8),
(*@\elnM{  35}@*)         triton.Config({'BLOCK_M': 128, 'BLOCK_N': 128}, num_stages=4, num_warps=8),
(*@\elnM{  36}@*)         # Medium configs (good for headdim 128)
(*@\elnM{  37}@*)         triton.Config({'BLOCK_M': 128, 'BLOCK_N': 64}, num_stages=3, num_warps=8),
(*@\elnM{  38}@*)         triton.Config({'BLOCK_M': 128, 'BLOCK_N': 64}, num_stages=4, num_warps=8),
(*@\elnM{  39}@*)         triton.Config({'BLOCK_M': 128, 'BLOCK_N': 64}, num_stages=3, num_warps=4),
(*@\elnM{  40}@*)         triton.Config({'BLOCK_M': 64, 'BLOCK_N': 128}, num_stages=3, num_warps=8),
(*@\elnM{  41}@*)         # Smaller configs (good for large headdim like 256)
(*@\elnM{  42}@*)         triton.Config({'BLOCK_M': 64, 'BLOCK_N': 64}, num_stages=3, num_warps=4),
(*@\elnM{  43}@*)         triton.Config({'BLOCK_M': 64, 'BLOCK_N': 64}, num_stages=4, num_warps=8),
(*@\elnM{  44}@*)         triton.Config({'BLOCK_M': 64, 'BLOCK_N': 64}, num_stages=3, num_warps=8),
(*@\elnM{  45}@*)         triton.Config({'BLOCK_M': 128, 'BLOCK_N': 32}, num_stages=3, num_warps=4),
(*@\elnM{  46}@*)         triton.Config({'BLOCK_M': 64, 'BLOCK_N': 32}, num_stages=4, num_warps=4),
(*@\elnM{  47}@*)         # Additional hdim256 configs - split work more to reduce register pressure
(*@\elnM{  48}@*)         triton.Config({'BLOCK_M': 32, 'BLOCK_N': 64}, num_stages=3, num_warps=4),
(*@\elnM{  49}@*)         triton.Config({'BLOCK_M': 32, 'BLOCK_N': 128}, num_stages=3, num_warps=8),
(*@\elnM{  50}@*)         triton.Config({'BLOCK_M': 16, 'BLOCK_N': 128}, num_stages=3, num_warps=4),
(*@\elnM{  51}@*)         triton.Config({'BLOCK_M': 16, 'BLOCK_N': 64}, num_stages=3, num_warps=4),
(*@\elnM{  52}@*)         triton.Config({'BLOCK_M': 64, 'BLOCK_N': 64}, num_stages=2, num_warps=4),
(*@\elnM{  53}@*)         triton.Config({'BLOCK_M': 64, 'BLOCK_N': 128}, num_stages=2, num_warps=8),
(*@\elnM{  54}@*)         triton.Config({'BLOCK_M': 128, 'BLOCK_N': 128}, num_stages=2, num_warps=8),
(*@\elnM{  55}@*)     ],
(*@\elnM{  56}@*)     key=['seqlen', 'BLOCK_DMODEL', 'IS_CAUSAL'],
(*@\elnM{  57}@*) )
(*@\elnE{  58}@*) @triton.jit
(*@\elnE{  59}@*) def _custom_attn_fwd(
(*@\elnE{  60}@*)     Q, K, V, Out,
(*@\ealelide{61}@*)
(*@\elnM{ 134}@*)     else:
(*@\elnM{ 135}@*)         hi = non_causal_end
(*@\elnM{ 136}@*)
(*@\elnM{ 137}@*)     for start_n in range(non_causal_end, hi, BLOCK_N):
(*@\elnM{ 138}@*)         start_n = tl.multiple_of(start_n, BLOCK_N)
(*@\elnE{ 139}@*)         k_ptrs = K + k_offset + (start_n + offs_n[:, None]) * stride_kn + offs_d[None, :] * stride_kk
(*@\elnE{ 140}@*)         k = tl.load(k_ptrs, mask=(start_n + offs_n[:, None]) < seqlen, other=0.0)
(*@\elnM{ 141}@*)         qk = tl.dot(q, tl.trans(k))
(*@\elnM{ 142}@*)         qk = tl.where(offs_m[:, None] >= (start_n + offs_n[None, :]), qk, float("-inf"))
(*@\elnE{ 143}@*)         m_ij = tl.max(qk, axis=1)
(*@\elnE{ 144}@*)         m_new = tl.maximum(m_i, m_ij)
(*@\elnM{ 145}@*)         alpha = tl.math.exp2(m_i - m_new)
(*@\elnM{ 146}@*)         p = tl.math.exp2(qk - m_new[:, None])
(*@\elnE{ 147}@*)         l_i = l_i * alpha + tl.sum(p, axis=1)
(*@\elnE{ 148}@*)         acc = acc * alpha[:, None]
(*@\elnE{ 149}@*)         v_ptrs = V + v_offset + (start_n + offs_n[:, None]) * stride_vn + offs_d[None, :] * stride_vk
(*@\ealelide{8}@*)
(*@\elnE{ 158}@*)
(*@\elnE{ 159}@*)
(*@\elnE{ 160}@*) def custom_attention_forward(q, k, v, causal=True, sm_scale=None):
(*@\elnM{ 161}@*)     """Optimized wrapper: fuse log2(e)*sm_scale into Q, autotuned kernel launch."""
(*@\elnE{ 162}@*)     batch, nheads, seqlen, headdim = q.shape
(*@\elnE{ 163}@*)     q, k, v = q.contiguous(), k.contiguous(), v.contiguous()
(*@\elnE{ 164}@*)     if sm_scale is None:
(*@\elnE{ 165}@*)         sm_scale = 1.0 / math.sqrt(headdim)
(*@\elnM{ 166}@*)     # Fuse sm_scale * log2(e) into Q to avoid per-element multiply in inner loop
(*@\elnM{ 167}@*)     q = (q * (sm_scale * 1.44269504)).contiguous()
(*@\elnE{ 168}@*)     o = torch.empty_like(q)
(*@\elnM{ 169}@*)     grid = lambda META: (triton.cdiv(seqlen, META['BLOCK_M']), batch * nheads)
(*@\elnE{ 170}@*)     _custom_attn_fwd[grid](
(*@\elnM{ 171}@*)         q, k, v, o,
(*@\elnE{ 172}@*)         q.stride(1), q.stride(2), q.stride(3),
(*@\elnE{ 173}@*)         k.stride(1), k.stride(2), k.stride(3),
(*@\elnE{ 174}@*)         v.stride(1), v.stride(2), v.stride(3),
\end{lstlisting}

\humancomment{Claude Opus 4.6}{claude-opus-4.6}{Reproduces the causal / non-causal split correctly and adds a
fairly comprehensive set of autotune configurations targeted at
H100. Recovers Flash-Attention~v2's delayed block-wise division.
Pre-scales \(Q\) outside the kernel, which is a wash in this
regime. Did not adopt Hopper-specific primitives such as TMA.
Overall an FA2-class implementation, well-tuned for H100 but not
Hopper-native.}

\paragraph{Agent: Gemini 3.1 Pro.}\mbox{}\\
TMA-style attempt via \texttt{tl.make\_block\_ptr}.
\begin{lstlisting}[style=eal/python]
(*@\elnU{  27}@*) # ================================================================
(*@\elnU{  28}@*)
(*@\ealregion{editable region begins at line 29}@*)
(*@\elnM{  29}@*) @triton.autotune(
(*@\elnM{  30}@*)     configs=[
(*@\elnM{  31}@*)         triton.Config({'BLOCK_M': 256, 'BLOCK_N': 128}, num_stages=3, num_warps=8),
(*@\elnM{  32}@*)         triton.Config({'BLOCK_M': 256, 'BLOCK_N': 64}, num_stages=3, num_warps=8),
(*@\elnM{  33}@*)         triton.Config({'BLOCK_M': 128, 'BLOCK_N': 256}, num_stages=3, num_warps=8),
(*@\elnM{  34}@*)         triton.Config({'BLOCK_M': 128, 'BLOCK_N': 128}, num_stages=3, num_warps=8),
(*@\elnM{  35}@*)         triton.Config({'BLOCK_M': 128, 'BLOCK_N': 128}, num_stages=4, num_warps=8),
(*@\elnM{  36}@*)         triton.Config({'BLOCK_M': 128, 'BLOCK_N': 128}, num_stages=5, num_warps=8),
(*@\elnM{  37}@*)         triton.Config({'BLOCK_M': 128, 'BLOCK_N': 64}, num_stages=3, num_warps=8),
(*@\elnM{  38}@*)         triton.Config({'BLOCK_M': 128, 'BLOCK_N': 64}, num_stages=4, num_warps=8),
(*@\elnM{  39}@*)         triton.Config({'BLOCK_M': 128, 'BLOCK_N': 64}, num_stages=5, num_warps=8),
(*@\elnM{  40}@*)         triton.Config({'BLOCK_M': 64, 'BLOCK_N': 256}, num_stages=3, num_warps=8),
(*@\elnM{  41}@*)         triton.Config({'BLOCK_M': 64, 'BLOCK_N': 256}, num_stages=4, num_warps=8),
(*@\elnM{  42}@*)         triton.Config({'BLOCK_M': 64, 'BLOCK_N': 128}, num_stages=3, num_warps=8),
(*@\elnM{  43}@*)         triton.Config({'BLOCK_M': 64, 'BLOCK_N': 128}, num_stages=4, num_warps=8),
(*@\elnM{  44}@*)         triton.Config({'BLOCK_M': 64, 'BLOCK_N': 128}, num_stages=5, num_warps=8),
(*@\elnM{  45}@*)         triton.Config({'BLOCK_M': 64, 'BLOCK_N': 64}, num_stages=3, num_warps=4),
(*@\elnM{  46}@*)         triton.Config({'BLOCK_M': 64, 'BLOCK_N': 64}, num_stages=4, num_warps=4),
(*@\elnM{  47}@*)         triton.Config({'BLOCK_M': 64, 'BLOCK_N': 64}, num_stages=5, num_warps=4),
(*@\elnM{  48}@*)         triton.Config({'BLOCK_M': 64, 'BLOCK_N': 64}, num_stages=4, num_warps=8),
(*@\elnM{  49}@*)         triton.Config({'BLOCK_M': 64, 'BLOCK_N': 64}, num_stages=5, num_warps=8),
(*@\elnM{  50}@*)         triton.Config({'BLOCK_M': 64, 'BLOCK_N': 64}, num_stages=8, num_warps=4),
(*@\elnM{  51}@*)         triton.Config({'BLOCK_M': 128, 'BLOCK_N': 32}, num_stages=4, num_warps=4),
(*@\elnM{  52}@*)         triton.Config({'BLOCK_M': 128, 'BLOCK_N': 32}, num_stages=5, num_warps=8),
(*@\elnM{  53}@*)         # Extra configs specifically for D=256
(*@\elnM{  54}@*)         triton.Config({'BLOCK_M': 128, 'BLOCK_N': 64}, num_stages=2, num_warps=8),
(*@\elnM{  55}@*)         triton.Config({'BLOCK_M': 64, 'BLOCK_N': 64}, num_stages=3, num_warps=8),
(*@\elnM{  56}@*)         triton.Config({'BLOCK_M': 64, 'BLOCK_N': 64}, num_stages=2, num_warps=4),
(*@\elnM{  57}@*)         triton.Config({'BLOCK_M': 64, 'BLOCK_N': 64}, num_stages=2, num_warps=8),
(*@\elnM{  58}@*)         triton.Config({'BLOCK_M': 64, 'BLOCK_N': 32}, num_stages=2, num_warps=4),
(*@\elnM{  59}@*)         triton.Config({'BLOCK_M': 64, 'BLOCK_N': 32}, num_stages=3, num_warps=4),
(*@\elnM{  60}@*)         triton.Config({'BLOCK_M': 64, 'BLOCK_N': 32}, num_stages=4, num_warps=4),
(*@\ealelide{88}@*)
(*@\elnM{ 164}@*)         m_i = m_new
(*@\elnM{ 165}@*)
(*@\elnM{ 166}@*)         k_block_ptr = tl.advance(k_block_ptr, (0, BLOCK_N))
(*@\elnM{ 167}@*)         v_block_ptr = tl.advance(v_block_ptr, (BLOCK_N, 0))
(*@\elnM{ 168}@*)
(*@\elnE{ 169}@*)     acc = acc / l_i[:, None]
(*@\elnM{ 170}@*)
(*@\elnM{ 171}@*)     o_block_ptr = tl.make_block_ptr(
(*@\elnM{ 172}@*)         base=Out + o_offset,
(*@\elnM{ 173}@*)         shape=(seqlen, BLOCK_DMODEL),
(*@\elnM{ 174}@*)         strides=(stride_om, stride_ok),
(*@\elnM{ 175}@*)         offsets=(start_m * BLOCK_M, 0),
(*@\elnM{ 176}@*)         block_shape=(BLOCK_M, BLOCK_DMODEL),
(*@\elnM{ 177}@*)         order=(1, 0),
(*@\elnM{ 178}@*)     )
(*@\elnM{ 179}@*)     tl.store(o_block_ptr, acc.to(Out.dtype.element_ty))
(*@\elnE{ 180}@*)
(*@\elnE{ 181}@*) def custom_attention_forward(q, k, v, causal=True, sm_scale=None):
(*@\elnE{ 182}@*)     batch, nheads, seqlen, headdim = q.shape
(*@\elnE{ 183}@*)     q, k, v = q.contiguous(), k.contiguous(), v.contiguous()
(*@\elnE{ 184}@*)     if sm_scale is None:
(*@\elnE{ 185}@*)         sm_scale = 1.0 / math.sqrt(headdim)
(*@\elnM{ 186}@*)
(*@\elnM{ 187}@*)     scale = sm_scale * 1.44269504
(*@\elnE{ 188}@*)     o = torch.empty_like(q)
(*@\elnM{ 189}@*)
(*@\elnM{ 190}@*)     grid = lambda META: (triton.cdiv(seqlen, META['BLOCK_M']), batch * nheads)
(*@\elnM{ 191}@*)
(*@\elnE{ 192}@*)     _custom_attn_fwd[grid](
(*@\elnM{ 193}@*)         q, k, v, o,
(*@\elnM{ 194}@*)         scale,
(*@\elnE{ 195}@*)         q.stride(1), q.stride(2), q.stride(3),
(*@\elnE{ 196}@*)         k.stride(1), k.stride(2), k.stride(3),
(*@\elnE{ 197}@*)         v.stride(1), v.stride(2), v.stride(3),
\end{lstlisting}

\humancomment{Gemini 3.1 Pro}{gemini-3.1-pro-preview}{The most interesting attempt at Hopper-specific optimisation.
Uses \texttt{tl.make\_block\_ptr} to express the K/V/O loads,
which is the Triton-level idiom that maps onto the H100's
tensor-memory accelerator. Otherwise an FA2 variant: causal
split, autotune configs, delayed denominator division, in-kernel
scaling of \(Q\). A reasonable preliminary step toward
Hopper-native attention.}

\paragraph{Agent: Qwen 3.6 Plus.}\mbox{}\\
A failure case (kept for contrast).
\begin{lstlisting}[style=eal/python]
(*@\elnE{  39}@*)     BLOCK_DMODEL: tl.constexpr,
(*@\elnE{  40}@*)     IS_CAUSAL: tl.constexpr,
(*@\elnE{  41}@*) ):
(*@\elnM{  42}@*)     """Optimized fused self-attention: two-pass causal + fused scale."""
(*@\elnE{  43}@*)     start_m = tl.program_id(0)
(*@\elnE{  44}@*)     off_hz = tl.program_id(1)
(*@\elnE{  45}@*)
(*@\ealelide{6}@*)
(*@\elnE{  52}@*)     offs_n = tl.arange(0, BLOCK_N)
(*@\elnE{  53}@*)     offs_d = tl.arange(0, BLOCK_DMODEL)
(*@\elnE{  54}@*)
(*@\elnM{  55}@*)     # Load Q tile (scale pre-fused in wrapper with log2(e) factor)
(*@\elnE{  56}@*)     q_ptrs = Q + q_offset + offs_m[:, None] * stride_qm + offs_d[None, :] * stride_qk
(*@\elnE{  57}@*)     q = tl.load(q_ptrs, mask=offs_m[:, None] < seqlen, other=0.0)
(*@\elnE{  58}@*)
(*@\ealelide{2}@*)
(*@\elnE{  61}@*)     l_i = tl.zeros([BLOCK_M], dtype=tl.float32)
(*@\elnE{  62}@*)     acc = tl.zeros([BLOCK_M, BLOCK_DMODEL], dtype=tl.float32)
(*@\elnE{  63}@*)
(*@\elnM{  64}@*)     # --- Pass 1: Non-causal blocks (no masking needed, full pipelining) ---
(*@\elnM{  65}@*)     if IS_CAUSAL:
(*@\elnM{  66}@*)         non_causal_end = (start_m * BLOCK_M // BLOCK_N) * BLOCK_N
(*@\elnM{  67}@*)     else:
(*@\elnM{  68}@*)         non_causal_end = seqlen
(*@\elnM{  69}@*)
(*@\elnM{  70}@*)     for start_n in range(0, non_causal_end, BLOCK_N):
(*@\elnE{  71}@*)         start_n = tl.multiple_of(start_n, BLOCK_N)
(*@\elnE{  72}@*)         k_ptrs = K + k_offset + (start_n + offs_n[:, None]) * stride_kn + offs_d[None, :] * stride_kk
(*@\elnM{  73}@*)         k = tl.load(k_ptrs)
(*@\elnM{  74}@*)         qk = tl.dot(q, tl.trans(k))
(*@\elnE{  75}@*)         m_ij = tl.max(qk, axis=1)
(*@\elnE{  76}@*)         m_new = tl.maximum(m_i, m_ij)
(*@\elnM{  77}@*)         alpha = tl.math.exp2(m_i - m_new)
(*@\elnM{  78}@*)         p = tl.math.exp2(qk - m_new[:, None])
(*@\elnE{  79}@*)         l_i = l_i * alpha + tl.sum(p, axis=1)
(*@\ealelide{38}@*)
(*@\elnE{ 121}@*)         sm_scale = 1.0 / math.sqrt(headdim)
(*@\elnM{ 122}@*)     # Fuse softmax scale and log2(e) into Q
(*@\elnM{ 123}@*)     q = (q * (sm_scale * 1.4426950408889634)).contiguous()
(*@\elnE{ 124}@*)     o = torch.empty_like(q)
(*@\elnM{ 125}@*)
(*@\elnM{ 126}@*)     # Per-headdim tuned configurations for H100
(*@\elnM{ 127}@*)     if headdim <= 64:
(*@\elnM{ 128}@*)         # Small head: memory-bound, use large blocks to amortize overhead
(*@\elnM{ 129}@*)         BLOCK_M, BLOCK_N = 128, 64
(*@\elnM{ 130}@*)         num_stages, num_warps = 4, 8
(*@\elnM{ 131}@*)     elif headdim <= 128:
(*@\elnM{ 132}@*)         # Medium head: balanced
(*@\elnM{ 133}@*)         BLOCK_M, BLOCK_N = 128, 64
(*@\elnM{ 134}@*)         num_stages, num_warps = 4, 8
(*@\elnM{ 135}@*)     else:
(*@\elnM{ 136}@*)         # Large head (256): compute-bound, use larger blocks for tensor core efficiency
(*@\elnM{ 137}@*)         BLOCK_M, BLOCK_N = 64, 64
(*@\elnM{ 138}@*)         num_stages, num_warps = 4, 8
(*@\elnM{ 139}@*)
(*@\elnE{ 140}@*)     grid = (triton.cdiv(seqlen, BLOCK_M), batch * nheads)
(*@\elnE{ 141}@*)     _custom_attn_fwd[grid](
(*@\elnM{ 142}@*)         q, k, v, o,
(*@\elnE{ 143}@*)         q.stride(1), q.stride(2), q.stride(3),
(*@\elnE{ 144}@*)         k.stride(1), k.stride(2), k.stride(3),
(*@\elnE{ 145}@*)         v.stride(1), v.stride(2), v.stride(3),
(*@\ealelide{1}@*)
(*@\elnE{ 147}@*)         seqlen,
(*@\elnE{ 148}@*)         BLOCK_M=BLOCK_M, BLOCK_N=BLOCK_N,
(*@\elnE{ 149}@*)         BLOCK_DMODEL=headdim, IS_CAUSAL=causal,
(*@\elnM{ 150}@*)         num_stages=num_stages, num_warps=num_warps,
(*@\elnE{ 151}@*)     )
(*@\elnE{ 152}@*)     return o
(*@\elnE{ 153}@*)
(*@\ealregion{editable region ends at line 153}@*)
\end{lstlisting}

\humancomment{Qwen 3.6 Plus}{qwen3.6-plus}{Effectively a failed attempt: passes the correctness check but
does not realise meaningful speedup over the SDPA baseline.
Useful as a contrast: even when the editable region is small
and the surrounding harness fixes the evaluation protocol, a
weak kernel cannot recover throughput.}

\subsection{\texorpdfstring{$L_\infty$}{L-infinity} Adversarial Training for Robust Accuracy}
\label{sec:expert:security-adversarial-training}

\begin{mdframed}[backgroundcolor=white,linecolor=gray!50,linewidth=0.6pt,roundcorner=2pt,innertopmargin=6pt,innerbottommargin=6pt]
\textbf{Task.} A custom adversarial training procedure is evaluated on robust accuracy under white-box \(\ell_\infty\) attacks while preserving clean accuracy across MNIST, CIFAR-10, and CIFAR-100 settings. Editable: the AdversarialTrainer inner attack and outer training loss in \texttt{custom\_adv\_train.py}. Read-only: the data loaders, model architectures, optimizer, learning-rate schedule, and evaluation attacks. Provided baselines are standard training, PGD-AT, TRADES, MART, and AWP.
\end{mdframed}

\paragraph{Template (editable region).}
\begin{lstlisting}[style=eal/python]
(*@\elnE{  10}@*) class AdversarialTrainer:
(*@\elnE{  11}@*)     """
(*@\elnE{  12}@*)     Adversarial training method.
(*@\elnE{  13}@*)
(*@\elnE{  14}@*)     The agent should modify this class to implement a better adversarial
(*@\elnE{  15}@*)     training procedure that improves model robustness against L_inf attacks.
(*@\elnE{  16}@*)
(*@\elnE{  17}@*)     Args:
(*@\elnE{  18}@*)         model (nn.Module): The model to train.
(*@\elnE{  19}@*)         eps (float): L_inf perturbation budget.
(*@\elnE{  20}@*)         alpha (float): Step size for adversarial perturbation generation.
(*@\elnE{  21}@*)         attack_steps (int): Number of PGD steps for adversarial example generation.
(*@\ealelide{20}@*)
(*@\elnE{  42}@*)             dict: Must contain 'loss' key (float).
(*@\elnE{  43}@*)         """
(*@\elnE{  44}@*)         # Default: standard (non-adversarial) training
(*@\elnE{  45}@*)         self.model.train()
(*@\elnE{  46}@*)         outputs = self.model(images)
(*@\elnE{  47}@*)         loss = F.cross_entropy(outputs, labels)
(*@\elnE{  48}@*)
(*@\elnE{  49}@*)         optimizer.zero_grad()
(*@\elnE{  50}@*)         loss.backward()
(*@\elnE{  51}@*)         optimizer.step()
(*@\elnE{  52}@*)
(*@\elnE{  53}@*)         return {'loss': loss.item()}
(*@\elnE{  54}@*)
\end{lstlisting}

\paragraph{Baseline: \texttt{pgdat}.}
\begin{lstlisting}[style=eal/python]
(*@\elnU{   8}@*) # EDITABLE -- implement AdversarialTrainer below
(*@\elnU{   9}@*) # ===================================================================
(*@\ealregion{editable region begins at line 10}@*)
(*@\elnE{  10}@*) class AdversarialTrainer:
(*@\elnM{  11}@*)     """PGD Adversarial Training (Madry et al., 2018)."""
(*@\elnE{  12}@*)
(*@\elnE{  13}@*)     def __init__(self, model, eps, alpha, attack_steps, num_classes, **kwargs):
(*@\elnE{  14}@*)         self.model = model
(*@\ealelide{3}@*)
(*@\elnE{  18}@*)         self.num_classes = num_classes
(*@\elnE{  19}@*)
(*@\elnE{  20}@*)     def train_step(self, images, labels, optimizer):
(*@\elnM{  21}@*)         # Generate adversarial examples using PGD
(*@\elnM{  22}@*)         self.model.eval()
(*@\elnM{  23}@*)         adv_images = images.clone().detach()
(*@\elnM{  24}@*)         adv_images = adv_images + torch.empty_like(adv_images).uniform_(-self.eps, self.eps)
(*@\elnM{  25}@*)         adv_images = torch.clamp(adv_images, 0.0, 1.0)
(*@\elnE{  26}@*)
(*@\elnM{  27}@*)         for _ in range(self.attack_steps):
(*@\elnM{  28}@*)             adv_images.requires_grad_(True)
(*@\elnM{  29}@*)             outputs = self.model(adv_images)
(*@\elnM{  30}@*)             loss = F.cross_entropy(outputs, labels)
(*@\elnM{  31}@*)             grad = torch.autograd.grad(loss, adv_images)[0]
(*@\elnM{  32}@*)             adv_images = adv_images.detach() + self.alpha * grad.sign()
(*@\elnM{  33}@*)             delta = torch.clamp(adv_images - images, min=-self.eps, max=self.eps)
(*@\elnM{  34}@*)             adv_images = torch.clamp(images + delta, 0.0, 1.0).detach()
(*@\elnE{  35}@*)
(*@\elnM{  36}@*)         # Train on adversarial examples
(*@\elnE{  37}@*)         self.model.train()
(*@\elnM{  38}@*)         outputs = self.model(adv_images)
(*@\elnE{  39}@*)         loss = F.cross_entropy(outputs, labels)
(*@\elnE{  40}@*)
(*@\elnE{  41}@*)         optimizer.zero_grad()
\end{lstlisting}

\paragraph{Agent: GPT-5.4.}\mbox{}\\
PGD with margin-gap inner loss; logit-margin-aware outer weights.
\begin{lstlisting}[style=eal/python]
(*@\elnU{   8}@*) # EDITABLE -- implement AdversarialTrainer below
(*@\elnU{   9}@*) # ===================================================================
(*@\ealregion{editable region begins at line 10}@*)
(*@\elnE{  10}@*) class AdversarialTrainer:
(*@\elnM{  11}@*)     """Adaptive Anchor-Margin Adversarial Training (AAMAT).
(*@\elnE{  12}@*)
(*@\elnM{  13}@*)     Novelty:
(*@\elnM{  14}@*)     1) adversarial examples maximize CE + margin-collapse loss;
(*@\elnM{  15}@*)     2) adversarial supervision uses a confidence-adaptive anchor target that
(*@\elnM{  16}@*)        interpolates between one-hot labels and the clean predictive distribution;
(*@\elnM{  17}@*)     3) robust regularization is upweighted for low-margin / low-confidence samples.
(*@\elnE{  18}@*)     """
(*@\elnE{  19}@*)
(*@\elnE{  20}@*)     def __init__(self, model, eps, alpha, attack_steps, num_classes, **kwargs):
(*@\ealelide{2}@*)
(*@\elnE{  23}@*)         self.alpha = alpha
(*@\elnE{  24}@*)         self.attack_steps = attack_steps
(*@\elnE{  25}@*)         self.num_classes = num_classes
(*@\elnM{  26}@*)         self.beta = 1.2
(*@\elnM{  27}@*)         self.gamma = 0.6
(*@\elnM{  28}@*)         self.lam_attack = 1.0
(*@\elnE{  29}@*)
(*@\elnE{  30}@*)     def train_step(self, images, labels, optimizer):
(*@\elnM{  31}@*)         self.model.train()
(*@\elnM{  32}@*)         logits_clean = self.model(images)
(*@\elnM{  33}@*)         loss_clean = F.cross_entropy(logits_clean, labels)
(*@\elnE{  34}@*)
(*@\elnM{  35}@*)         with torch.no_grad():
(*@\elnM{  36}@*)             probs = F.softmax(logits_clean, dim=1)
(*@\elnM{  37}@*)             p_true = probs.gather(1, labels.unsqueeze(1)).squeeze(1)
(*@\elnM{  38}@*)             masked = logits_clean.detach().clone()
(*@\elnM{  39}@*)             masked.scatter_(1, labels.unsqueeze(1), -1e9)
(*@\elnM{  40}@*)             top_other = masked.max(dim=1).values
(*@\elnM{  41}@*)             true_logit = logits_clean.detach().gather(1, labels.unsqueeze(1)).squeeze(1)
(*@\elnM{  42}@*)             margin = true_logit - top_other
(*@\ealelide{6}@*)
(*@\elnE{  49}@*)
(*@\elnM{  50}@*)         self.model.eval()
(*@\elnM{  51}@*)         adv_images = torch.clamp(images.detach() + 0.001 * torch.randn_like(images), 0.0, 1.0)
(*@\elnM{  52}@*)         for _ in range(self.attack_steps):
(*@\elnM{  53}@*)             adv_images.requires_grad_(True)
(*@\elnM{  54}@*)             logits_adv = self.model(adv_images)
(*@\elnM{  55}@*)             true_adv = logits_adv.gather(1, labels.unsqueeze(1)).squeeze(1)
(*@\elnM{  56}@*)             masked_adv = logits_adv.clone()
(*@\elnM{  57}@*)             masked_adv.scatter_(1, labels.unsqueeze(1), -1e9)
(*@\elnM{  58}@*)             max_other_adv = masked_adv.max(dim=1).values
(*@\elnM{  59}@*)             loss_attack = F.cross_entropy(logits_adv, labels) + self.lam_attack * F.softplus(max_other_adv - true_adv).mean()
(*@\elnM{  60}@*)             grad = torch.autograd.grad(loss_attack, adv_images)[0]
(*@\elnM{  61}@*)             adv_images = adv_images.detach() + self.alpha * grad.sign()
(*@\elnM{  62}@*)             delta = torch.clamp(adv_images - images, min=-self.eps, max=self.eps)
(*@\elnM{  63}@*)             adv_images = torch.clamp(images + delta, 0.0, 1.0).detach()
(*@\elnM{  64}@*)
(*@\elnE{  65}@*)         self.model.train()
(*@\elnM{  66}@*)         logits_adv = self.model(adv_images)
(*@\elnM{  67}@*)         adv_logp = F.log_softmax(logits_adv, dim=1)
(*@\elnM{  68}@*)         loss_anchor = (weight * (-(anchor * adv_logp).sum(dim=1))).mean()
(*@\elnM{  69}@*)         true_adv = logits_adv.gather(1, labels.unsqueeze(1)).squeeze(1)
(*@\elnM{  70}@*)         masked_adv = logits_adv.clone()
(*@\elnM{  71}@*)         masked_adv.scatter_(1, labels.unsqueeze(1), -1e9)
(*@\elnM{  72}@*)         max_other_adv = masked_adv.max(dim=1).values
(*@\elnM{  73}@*)         loss_margin = (weight * F.softplus(max_other_adv - true_adv)).mean()
(*@\elnM{  74}@*)         loss = loss_clean + self.beta * loss_anchor + self.gamma * loss_margin
(*@\elnE{  75}@*)
(*@\elnE{  76}@*)         optimizer.zero_grad()
(*@\elnE{  77}@*)         loss.backward()
(*@\elnE{  78}@*)         optimizer.step()
(*@\elnM{  79}@*)         return {'loss': loss.item(), 'loss_clean': loss_clean.item(), 'loss_anchor': loss_anchor.item()}
(*@\ealregion{editable region ends at line 79}@*)
(*@\elnU{  80}@*) # ===================================================================
(*@\elnU{  81}@*) # END EDITABLE
\end{lstlisting}

\humancomment{GPT-5.4}{gpt-5.4}{The most distinctive method among the five. Inner attack
maximises \(\mathrm{CE} + \lambda \cdot (\text{margin-gap loss})\),
and the outer loss reweights examples as
\(0.35 + 0.65 \cdot \max\bigl(1 - p_y,\ \sigma(-\text{margin})\bigr)\).
It is also the model with the shortest reasoning trace by far
(only two iterations with thinking, around 5\,000 characters
total). Only GPT considered the logit margin explicitly.}

\paragraph{Agent: Qwen 3.6 Plus.}\mbox{}\\
TRADES-style attack with focal-weighted outer loss.
\begin{lstlisting}[style=eal/python]
(*@\elnU{   9}@*) # ===================================================================
(*@\ealregion{editable region begins at line 10}@*)
(*@\elnE{  10}@*) class AdversarialTrainer:
(*@\elnE{  11}@*)     """
(*@\elnM{  12}@*)     Robust Focal TRADES (RFT).
(*@\elnE{  13}@*)
(*@\elnM{  14}@*)     Novel contributions:
(*@\elnM{  15}@*)     1. Dual-objective: Clean CE preserves natural accuracy while
(*@\elnM{  16}@*)        adversarial CE directly trains robust classification on perturbed inputs.
(*@\elnM{  17}@*)     2. Confidence-adaptive focal weighting: both adversarial CE and KL
(*@\elnM{  18}@*)        divergence are weighted by (1 - p_adv)^gamma, dynamically prioritizing
(*@\elnM{  19}@*)        samples where the model is currently vulnerable on adversarial inputs.
(*@\elnM{  20}@*)        This generalizes MART (clean-confidence-weighted) to use the adversarial
(*@\elnM{  21}@*)        prediction confidence as the weighting signal.
(*@\elnM{  22}@*)     3. KL divergence regularizer preserves decision boundary smoothness between
(*@\elnM{  23}@*)        clean and adversarial predictions.
(*@\elnE{  24}@*)
(*@\elnM{  25}@*)     Key design choices:
(*@\elnM{  26}@*)     - gamma=1.0: balanced emphasis. gamma=0 recovers standard TRADES + PGDAT.
(*@\elnM{  27}@*)     - Adversarial CE uses focal weighting to target the hardest adversarial
(*@\elnM{  28}@*)       examples directly.
(*@\elnM{  29}@*)     - KL regularization prevents overconfident predictions and maintains
(*@\elnM{  30}@*)       decision boundary smoothness.
(*@\elnE{  31}@*)     """
(*@\elnE{  32}@*)
(*@\elnE{  33}@*)     def __init__(self, model, eps, alpha, attack_steps, num_classes, **kwargs):
(*@\ealelide{2}@*)
(*@\elnE{  36}@*)         self.alpha = alpha
(*@\elnE{  37}@*)         self.attack_steps = attack_steps
(*@\elnE{  38}@*)         self.num_classes = num_classes
(*@\elnM{  39}@*)         self.beta = 6.0     # KL regularization weight (TRADES-style)
(*@\elnM{  40}@*)         self.gamma = 1.0    # Focal exponent for confidence-adaptive weighting
(*@\elnE{  41}@*)
(*@\elnE{  42}@*)     def train_step(self, images, labels, optimizer):
(*@\elnM{  43}@*)         # == Step 1: Generate adversarial examples via KL maximization ==
(*@\ealelide{35}@*)
(*@\elnM{  79}@*)         loss_clean = F.cross_entropy(logits_clean, labels)
(*@\elnM{  80}@*)
(*@\elnM{  81}@*)         # Component B: Focal-weighted adversarial CE
(*@\elnM{  82}@*)         # Directly improves robust classification on perturbed inputs
(*@\elnM{  83}@*)         ce_adv = F.cross_entropy(logits_adv, labels, reduction='none')
(*@\elnM{  84}@*)         loss_adv = (focal_weight * ce_adv).mean()
(*@\elnM{  85}@*)
(*@\elnM{  86}@*)         # Component C: Focal-weighted KL divergence (smoothness regularization)
(*@\elnM{  87}@*)         # KL(adv || clean) weighted by adversarial focal weight
(*@\elnM{  88}@*)         nat_probs = F.softmax(logits_clean, dim=1)
(*@\elnM{  89}@*)         kl_per_sample = F.kl_div(
(*@\elnM{  90}@*)             torch.log(adv_probs + 1e-12),
(*@\elnM{  91}@*)             nat_probs,
(*@\elnM{  92}@*)             reduction='none',
(*@\elnM{  93}@*)         ).sum(dim=1)
(*@\elnM{  94}@*)         loss_kl = (focal_weight * kl_per_sample).mean()
(*@\elnM{  95}@*)
(*@\elnM{  96}@*)         # Combined: clean + adv focal + KL regularizer
(*@\elnM{  97}@*)         loss = loss_clean + loss_adv + self.beta * loss_kl
(*@\elnE{  98}@*)
(*@\elnE{  99}@*)         optimizer.zero_grad()
(*@\elnE{ 100}@*)         loss.backward()
(*@\elnE{ 101}@*)         optimizer.step()
(*@\elnE{ 102}@*)
(*@\elnM{ 103}@*)         return {
(*@\elnM{ 104}@*)             'loss': loss.item(),
(*@\elnM{ 105}@*)             'loss_clean': loss_clean.item(),
(*@\elnM{ 106}@*)             'loss_adv': loss_adv.item(),
(*@\elnM{ 107}@*)             'loss_kl': loss_kl.item(),
(*@\elnM{ 108}@*)         }
(*@\elnE{ 109}@*)
(*@\ealregion{editable region ends at line 109}@*)
(*@\elnU{ 110}@*) # ===================================================================
(*@\elnU{ 111}@*) # END EDITABLE
\end{lstlisting}

\humancomment{Qwen 3.6 Plus}{qwen3.6-plus}{TRADES-style PGD inner attack combined with a focal-weighted
outer loss
\(L = \mathrm{CE}_{\text{clean}} + w \cdot \mathrm{CE}_{\text{adv}}
+ 6 w \cdot \mathrm{KL}\), where \(w = (1 - p_{\text{adv}})\).
Performs worst on MNIST among all five models, but best on
both CIFAR-10 and CIFAR-100. The split is informative: the
focal weighting helps on harder, multi-class problems but
destabilises the easy MNIST regime.}

\subsection{Quantization-Aware Language-Model Training}
\label{sec:expert:llm-qat-algorithm}

\begin{mdframed}[backgroundcolor=white,linecolor=gray!50,linewidth=0.6pt,roundcorner=2pt,innertopmargin=6pt,innerbottommargin=6pt]
\textbf{Task.} A training-side quantization-aware training algorithm is evaluated by the WikiText-2 perplexity gap between full-precision Pythia-1.4B and INT4, INT3, and INT2 group-quantized variants. Editable: the fake-quant forward, gradient surrogate, real quantize-dequantize path, QAT wrapper, learnable parameters, and \texttt{CONFIG\_OVERRIDES} in \texttt{custom\_qat.py}. Read-only: model loading, WikiText-2 sampling, the training loop, final real-QDQ roundtrip, and perplexity evaluation. Provided baselines are \texttt{no\_qat}, \texttt{ste}, \texttt{lsq}, and \texttt{finetune\_then\_ptq}.
\end{mdframed}

\paragraph{Template (editable region).}
\begin{lstlisting}[style=eal/python]
(*@\elnE{  33}@*) # Per-method training hyperparameters.  The training loop reads this dict.
(*@\elnE{  34}@*) # Override any of these in your method to retune.
(*@\elnE{  35}@*) CONFIG_OVERRIDES = {
(*@\elnE{  36}@*)     "learning_rate": 2e-5,
(*@\elnE{  37}@*)     "num_steps": 500,
(*@\elnE{  38}@*)     "batch_size": 2,
(*@\elnE{  39}@*)     "gradient_accumulation_steps": 4,
(*@\elnE{  40}@*)     "max_grad_norm": 1.0,
(*@\elnE{  41}@*)     "warmup_steps": 50,
(*@\elnE{  42}@*)     "weight_decay": 0.0,
(*@\elnE{  43}@*) }
(*@\elnE{  44}@*)
(*@\ealelide{119}@*)
(*@\elnE{ 164}@*)             else:
(*@\elnE{ 165}@*)                 _replace(child)
(*@\elnE{ 166}@*)
(*@\elnE{ 167}@*)     _replace(model)
(*@\elnE{ 168}@*)     # Restore the LM head to full precision (covers GPT-2 `lm_head` and
(*@\elnE{ 169}@*)     # Pythia / GPTNeoX `embed_out`).
(*@\elnE{ 170}@*)     for head_attr in ("lm_head", "embed_out"):
(*@\elnE{ 171}@*)         head = getattr(model, head_attr, None)
(*@\elnE{ 172}@*)         if isinstance(head, QATWrapper):
(*@\elnE{ 173}@*)             setattr(model, head_attr, head.linear)
(*@\elnE{ 174}@*)
(*@\elnE{ 175}@*)     return model
(*@\elnE{ 176}@*)
\end{lstlisting}

\paragraph{Baseline: \texttt{ste}.}
\begin{lstlisting}[style=eal/python]
(*@\elnU{  31}@*) # ===============================================================================
(*@\elnU{  32}@*)
(*@\ealregion{editable region begins at line 33}@*)
(*@\elnM{  33}@*)
(*@\elnM{  34}@*) # == Straight-Through Estimator (STE) QAT baseline =============================
(*@\elnM{  35}@*)
(*@\elnE{  36}@*) CONFIG_OVERRIDES = {
(*@\elnE{  37}@*)     "learning_rate": 2e-5,
(*@\elnE{  38}@*)     "num_steps": 500,
(*@\ealelide{14}@*)
(*@\elnE{  53}@*) def fake_quantize_weight(weight, num_bits, group_size):
(*@\elnE{  54}@*)     qmin, qmax = _qrange(num_bits)
(*@\elnE{  55}@*)     out_features, in_features = weight.shape
(*@\elnM{  56}@*)     assert in_features % group_size == 0
(*@\elnE{  57}@*)     w = weight.float().reshape(out_features, -1, group_size)
(*@\elnM{  58}@*)     # Recompute scale on-the-fly each forward (max-abs / qmax).
(*@\elnE{  59}@*)     w_max = w.abs().amax(dim=-1, keepdim=True).clamp(min=1e-12)
(*@\elnE{  60}@*)     scale = w_max / qmax
(*@\elnE{  61}@*)     w_q = torch.clamp(torch.round(w / scale), qmin, qmax) * scale
(*@\elnM{  62}@*)     # Straight-through: forward = quantized, backward = identity.
(*@\elnE{  63}@*)     w_dq = w + (w_q - w).detach()
(*@\elnE{  64}@*)     return w_dq.reshape(out_features, in_features).to(weight.dtype)
(*@\elnE{  65}@*)
(*@\ealelide{36}@*)
(*@\elnE{ 102}@*)
(*@\elnE{ 103}@*)
(*@\elnE{ 104}@*) def prepare_qat_model(model, num_bits, group_size):
(*@\elnM{ 105}@*)     from transformers.pytorch_utils import Conv1D
(*@\elnE{ 106}@*)
(*@\elnE{ 107}@*)     def _replace(parent):
(*@\elnE{ 108}@*)         for name, child in list(parent.named_children()):
(*@\elnE{ 109}@*)             if isinstance(child, nn.Linear):
(*@\elnM{ 110}@*)                 setattr(parent, name, QATWrapper(child, num_bits=num_bits, group_size=group_size))
(*@\elnE{ 111}@*)             elif isinstance(child, Conv1D):
(*@\elnE{ 112}@*)                 in_f, out_f = child.weight.shape
(*@\elnE{ 113}@*)                 lin = nn.Linear(in_f, out_f, bias=child.bias is not None,
(*@\ealelide{2}@*)
(*@\elnE{ 116}@*)                     lin.weight.copy_(child.weight.t().contiguous())
(*@\elnE{ 117}@*)                     if child.bias is not None:
(*@\elnE{ 118}@*)                         lin.bias.copy_(child.bias)
(*@\elnM{ 119}@*)                 setattr(parent, name, QATWrapper(lin, num_bits=num_bits, group_size=group_size))
(*@\elnE{ 120}@*)             else:
(*@\elnE{ 121}@*)                 _replace(child)
(*@\elnE{ 122}@*)
\end{lstlisting}

\paragraph{Agent: GPT-5.4.}\mbox{}\\
RTN-anchored LSQ with cosine schedule between learned grid and RTN.
\begin{lstlisting}[style=eal/python]
(*@\elnE{  49}@*)     return qmin, qmax
(*@\elnE{  50}@*)
(*@\elnE{  51}@*)
(*@\elnM{  52}@*) class _LSQQuant(torch.autograd.Function):
(*@\elnM{  53}@*)     @staticmethod
(*@\elnM{  54}@*)     def forward(ctx, w, scale, qmin, qmax, g_scale):
(*@\elnM{  55}@*)         w_div = w / scale
(*@\elnM{  56}@*)         w_clip = torch.clamp(w_div, qmin, qmax)
(*@\elnM{  57}@*)         w_round = torch.round(w_clip)
(*@\elnM{  58}@*)         ctx.save_for_backward(w_div, scale)
(*@\elnM{  59}@*)         ctx.qmin = qmin
(*@\elnM{  60}@*)         ctx.qmax = qmax
(*@\elnM{  61}@*)         ctx.g_scale = g_scale
(*@\elnM{  62}@*)         return w_round * scale
(*@\elnE{  63}@*)
(*@\elnM{  64}@*)     @staticmethod
(*@\elnM{  65}@*)     def backward(ctx, grad_out):
(*@\elnM{  66}@*)         w_div, scale = ctx.saved_tensors
(*@\elnM{  67}@*)         qmin, qmax, g = ctx.qmin, ctx.qmax, ctx.g_scale
(*@\elnM{  68}@*)         in_range = (w_div > qmin) & (w_div < qmax)
(*@\elnM{  69}@*)         grad_w = torch.where(in_range, grad_out, torch.zeros_like(grad_out))
(*@\elnM{  70}@*)         below = (w_div <= qmin).float() * float(qmin)
(*@\elnM{  71}@*)         above = (w_div >= qmax).float() * float(qmax)
(*@\elnM{  72}@*)         inside = in_range.float() * (torch.round(w_div) - w_div)
(*@\elnM{  73}@*)         grad_s = ((below + above + inside) * grad_out).sum(dim=-1, keepdim=True) * g
(*@\elnM{  74}@*)         return grad_w, grad_s, None, None, None
(*@\elnE{  75}@*)
(*@\elnE{  76}@*)
(*@\elnM{  77}@*) def _reshape_groups(weight, group_size):
(*@\elnE{  78}@*)     out_features, in_features = weight.shape
(*@\elnE{  79}@*)     assert in_features % group_size == 0, (
(*@\elnE{  80}@*)         f"in_features {in_features} not divisible by group_size {group_size}"
(*@\elnE{  81}@*)     )
(*@\elnM{  82}@*)     return weight.float().reshape(out_features, -1, group_size)
(*@\elnM{  83}@*)
(*@\ealelide{81}@*)
(*@\elnM{ 167}@*)
(*@\elnE{ 168}@*)     def forward(self, x):
(*@\elnE{ 169}@*)         x = fake_quantize_activation(x, self.num_bits)
(*@\elnM{ 170}@*)         w_q = fake_quantize_weight(
(*@\elnM{ 171}@*)             self.linear.weight,
(*@\elnM{ 172}@*)             self.num_bits,
(*@\elnM{ 173}@*)             self.group_size,
(*@\elnM{ 174}@*)             scale=self._lsq_scale(),
(*@\elnM{ 175}@*)             blend=self._blend(),
(*@\elnM{ 176}@*)         )
(*@\elnE{ 177}@*)         return F.linear(x, w_q, self.linear.bias)
(*@\elnM{ 178}@*)
(*@\elnM{ 179}@*)     def aux_loss(self, step, total_steps):
(*@\elnM{ 180}@*)         self._step.fill_(int(step))
(*@\elnM{ 181}@*)         self._total_steps.fill_(int(total_steps))
(*@\elnM{ 182}@*)         p = self._progress()
(*@\elnM{ 183}@*)         qmin, qmax = _qrange(self.num_bits)
(*@\elnM{ 184}@*)         w = _reshape_groups(self.linear.weight, self.group_size)
(*@\elnM{ 185}@*)         w_max = w.abs().amax(dim=-1, keepdim=True).clamp(min=1e-12)
(*@\elnM{ 186}@*)         rtn_scale = w_max / max(1, qmax)
(*@\elnM{ 187}@*)         scaled = torch.clamp(w / rtn_scale, qmin, qmax)
(*@\elnM{ 188}@*)         frac = scaled - torch.round(scaled)
(*@\elnM{ 189}@*)         # Focus on values that are not already near a clipping boundary.
(*@\elnM{ 190}@*)         center_mask = (scaled.abs() < (qmax - 0.25)).float()
(*@\elnM{ 191}@*)         align = (frac.pow(2) * center_mask).sum() / center_mask.sum().clamp(min=1.0)
(*@\elnM{ 192}@*)         # Keep learned LSQ scale near RTN scale so the fake-quant grid does not drift.
(*@\elnM{ 193}@*)         scale_reg = ((self._lsq_scale() - rtn_scale) / rtn_scale.clamp(min=1e-8)).pow(2).mean()
(*@\elnM{ 194}@*)         late = p * p
(*@\elnM{ 195}@*)         align_w = (0.0015 if self.num_bits >= 4 else 0.003 if self.num_bits == 3 else 0.006) * late
(*@\elnM{ 196}@*)         scale_w = 0.0005
(*@\elnM{ 197}@*)         return align_w * align + scale_w * scale_reg
(*@\elnE{ 198}@*)
(*@\elnE{ 199}@*)
(*@\elnE{ 200}@*) def prepare_qat_model(model, num_bits, group_size):
\end{lstlisting}

\humancomment{GPT-5.4}{gpt-5.4}{During QAT the method interpolates between a deployment RTN
grid and a learned LSQ grid using a cosine schedule that stays
closer to RTN. Conceptually a clean attempt: a hybrid that
keeps PTQ-deployable scales while letting the network adapt.
However, the final evaluation ignores the learned scale and
recomputes pure RTN max-abs from the trained weights, so most
of the QAT work is discarded at deployment time. The INT2/3/4
results are all weak; without a controlled ablation it is
unclear whether the underlying method is at fault or the
eval-time RTN re-scaling.}

\paragraph{Agent: Claude Opus 4.6.}\mbox{}\\
Adaptive clipping + noise injection + Stable-QAT EMA + AdaRound aux.
\begin{lstlisting}[style=eal/python]
(*@\elnU{  31}@*) # ===============================================================================
(*@\elnU{  32}@*)
(*@\ealregion{editable region begins at line 33}@*)
(*@\elnM{  33}@*)
(*@\elnM{  34}@*) # == Adaptive Clipping QAT with Gradual Noise Injection (AC-GNI) v2 ===========
(*@\elnM{  35}@*) # Key improvements over v1:
(*@\elnM{  36}@*) # 1) Bit-adaptive clipping init: more aggressive clipping at lower bits
(*@\elnM{  37}@*) #    (clip_ratio = 0.80 at INT2 vs 0.95 at INT4)
(*@\elnM{  38}@*) # 2) Extended noise warmup: 50% of training at INT2, 30% at INT3/4
(*@\elnM{  39}@*) # 3) EMA scale smoothing: exponential moving average on learned scales
(*@\elnM{  40}@*) #    prevents scale oscillation that causes INT2 divergence
(*@\elnM{  41}@*) # 4) Stronger outlier regularization with annealing: start strong, decay
(*@\elnM{  42}@*) # 5) Quantization-error-aware loss: small MSE term between FP and quantized
(*@\elnM{  43}@*) #    weights during training to keep weights near grid points
(*@\elnM{  44}@*)
(*@\elnE{  45}@*) CONFIG_OVERRIDES = {
(*@\elnE{  46}@*)     "learning_rate": 2e-5,
(*@\elnE{  47}@*)     "num_steps": 500,
(*@\ealelide{12}@*)
(*@\elnE{  60}@*)     return qmin, qmax
(*@\elnE{  61}@*)
(*@\elnE{  62}@*)
(*@\elnM{  63}@*) class _ACGNIQuantFn(torch.autograd.Function):
(*@\elnM{  64}@*)     """Custom autograd: LSQ-style quant with learned clipping bounds."""
(*@\elnE{  65}@*)
(*@\elnM{  66}@*)     @staticmethod
(*@\elnM{  67}@*)     def forward(ctx, w, scale, qmin, qmax, grad_scale_factor):
(*@\elnM{  68}@*)         w_div = w / scale
(*@\elnM{  69}@*)         w_clip = torch.clamp(w_div, qmin, qmax)
(*@\elnM{  70}@*)         w_round = torch.round(w_clip)
(*@\elnM{  71}@*)         ctx.save_for_backward(w_div, scale)
(*@\elnM{  72}@*)         ctx.qmin = qmin
(*@\elnM{  73}@*)         ctx.qmax = qmax
(*@\elnM{  74}@*)         ctx.grad_scale_factor = grad_scale_factor
(*@\elnM{  75}@*)         return w_round * scale
(*@\ealelide{171}@*)
(*@\elnM{ 250}@*)             qe_strength = 0.001
(*@\elnM{ 251}@*)             if self.num_bits <= 2:
(*@\elnM{ 252}@*)                 qe_strength = 0.005
(*@\elnM{ 253}@*)             outlier_reg = outlier_reg + qe * qe_strength
(*@\elnM{ 254}@*)
(*@\elnM{ 255}@*)         return outlier_reg
(*@\elnE{ 256}@*)
(*@\elnE{ 257}@*)
(*@\elnE{ 258}@*) def prepare_qat_model(model, num_bits, group_size):
(*@\elnM{ 259}@*)     """Replace nn.Linear layers with QATWrapper (LM head stays FP)."""
(*@\elnM{ 260}@*)     from transformers.pytorch_utils import Conv1D
(*@\elnE{ 261}@*)
(*@\elnM{ 262}@*)     QATWrapper._num_bits_global = num_bits
(*@\elnE{ 263}@*)
(*@\elnE{ 264}@*)     def _replace(parent):
(*@\elnE{ 265}@*)         for name, child in list(parent.named_children()):
(*@\elnE{ 266}@*)             if isinstance(child, nn.Linear):
(*@\elnM{ 267}@*)                 wrapper = QATWrapper(child, num_bits=num_bits,
(*@\elnM{ 268}@*)                                      group_size=group_size)
(*@\elnE{ 269}@*)                 setattr(parent, name, wrapper)
(*@\elnE{ 270}@*)             elif isinstance(child, Conv1D):
(*@\elnE{ 271}@*)                 in_f, out_f = child.weight.shape
(*@\elnE{ 272}@*)                 lin = nn.Linear(in_f, out_f, bias=child.bias is not None,
(*@\elnM{ 273}@*)                                 device=child.weight.device,
(*@\elnM{ 274}@*)                                 dtype=child.weight.dtype)
(*@\elnE{ 275}@*)                 with torch.no_grad():
(*@\elnE{ 276}@*)                     lin.weight.copy_(child.weight.t().contiguous())
(*@\elnE{ 277}@*)                     if child.bias is not None:
(*@\elnE{ 278}@*)                         lin.bias.copy_(child.bias)
(*@\elnM{ 279}@*)                 wrapper = QATWrapper(lin, num_bits=num_bits,
(*@\elnM{ 280}@*)                                      group_size=group_size)
(*@\elnE{ 281}@*)                 setattr(parent, name, wrapper)
(*@\elnE{ 282}@*)             else:
(*@\elnE{ 283}@*)                 _replace(child)
\end{lstlisting}

\humancomment{Claude Opus 4.6}{claude-opus-4.6}{A combinatorial attempt: adaptive clipping range, gradual
noise injection, Stable-QAT-style EMA, plus an AdaRound-style
auxiliary loss. Considerable hyperparameter tuning around the
noise schedule and clip ranges, but each component is taken
off-the-shelf. We do not see strong methodological novelty
beyond the combination itself.}

\subsection{Latent Normalization for World Models}
\label{sec:expert:tdmpc2-simnorm}

\begin{mdframed}[backgroundcolor=white,linecolor=gray!50,linewidth=0.6pt,roundcorner=2pt,innertopmargin=6pt,innerbottommargin=6pt]
\textbf{Task.} A custom latent normalization layer is evaluated inside the TD-MPC2 world model encoder and dynamics network, with episode reward measured on DMControl walker-walk, cheetah-run, and a hidden cartpole-swingup split. Editable: the CustomSimNorm class in \texttt{custom\_simnorm.py}. Read-only: the surrounding encoder, dynamics model, training procedure, and evaluation harness. Provided baselines are SimNorm, L2Norm, RMSNorm, and identity normalization.
\end{mdframed}

\paragraph{Template (editable region).}
\begin{lstlisting}[style=eal/python]
(*@\elnE{  16}@*) class CustomSimNorm(nn.Module):
(*@\elnE{  17}@*)     """Custom normalization for latent state representations in world models.
(*@\elnE{  18}@*)
(*@\elnE{  19}@*)     Interface contract (same as SimNorm):
(*@\elnE{  20}@*)         __init__(cfg)  -- cfg.simnorm_dim is the group size (default: 8)
(*@\elnE{  21}@*)         forward(x: Tensor) -> Tensor  (same shape as input)
(*@\elnE{  22}@*)
(*@\elnE{  23}@*)     The input tensor has shape (*batch_dims, latent_dim) where latent_dim
(*@\elnE{  24}@*)     is divisible by simnorm_dim. Your normalization should constrain the
(*@\elnE{  25}@*)     geometry of the latent space to improve world model learning.
(*@\elnE{  26}@*)
(*@\elnE{  27}@*)     Evaluated on DMControl walker-walk and cheetah-run tasks.
(*@\ealelide{3}@*)
(*@\elnE{  31}@*)         super().__init__()
(*@\elnE{  32}@*)         self.dim = cfg.simnorm_dim
(*@\elnE{  33}@*)
(*@\elnE{  34}@*)     def forward(self, x):
(*@\elnE{  35}@*)         # Default: SimNorm (simplicial normalization)
(*@\elnE{  36}@*)         # Reshape into groups of size self.dim and apply softmax
(*@\elnE{  37}@*)         shp = x.shape
(*@\elnE{  38}@*)         x = x.view(*shp[:-1], -1, self.dim)
(*@\elnE{  39}@*)         x = F.softmax(x, dim=-1)
(*@\elnE{  40}@*)         return x.view(*shp)
(*@\elnE{  41}@*)
(*@\elnE{  42}@*)     def __repr__(self):
(*@\elnE{  43}@*)         return f"CustomSimNorm(dim={self.dim})"
\end{lstlisting}

\paragraph{Baseline: \texttt{l2norm}.}
\begin{lstlisting}[style=eal/python]
(*@\elnU{  14}@*) # EDITABLE: Custom latent normalization
(*@\elnU{  15}@*) # =====================================================================
(*@\ealregion{editable region begins at line 16}@*)
(*@\elnE{  16}@*) class CustomSimNorm(nn.Module):
(*@\elnM{  17}@*)     """L2 normalization baseline -- projects latent vectors onto a hypersphere."""
(*@\elnE{  18}@*)
(*@\elnE{  19}@*)     def __init__(self, cfg):
(*@\elnE{  20}@*)         super().__init__()
(*@\elnE{  21}@*)         self.dim = cfg.simnorm_dim
(*@\elnM{  22}@*)         self.eps = 1e-8
(*@\elnM{  23}@*)         # Learnable scale parameter
(*@\elnM{  24}@*)         self.scale = nn.Parameter(torch.ones(1))
(*@\elnE{  25}@*)
(*@\elnE{  26}@*)     def forward(self, x):
(*@\elnM{  27}@*)         # L2 normalize across the last dimension and apply learnable scale
(*@\elnM{  28}@*)         norm = torch.norm(x, p=2, dim=-1, keepdim=True).clamp(min=self.eps)
(*@\elnM{  29}@*)         return self.scale * x / norm
(*@\elnE{  30}@*)
(*@\elnE{  31}@*)     def __repr__(self):
(*@\elnM{  32}@*)         return f"CustomSimNorm(dim={self.dim}, type=L2Norm)"
(*@\ealregion{editable region ends at line 32}@*)
\end{lstlisting}

\paragraph{Agent: GPT-5.4.}\mbox{}\\
Entmax-1.5-gated spherical simplex normalization.
\begin{lstlisting}[style=eal/python]
(*@\elnU{  14}@*) # EDITABLE: Custom latent normalization
(*@\elnU{  15}@*) # =====================================================================
(*@\ealregion{editable region begins at line 16}@*)
(*@\elnE{  16}@*) class CustomSimNorm(nn.Module):
(*@\elnM{  17}@*)     """Entmax-gated spherical simplex normalization.
(*@\elnE{  18}@*)
(*@\elnM{  19}@*)     Novel contribution:
(*@\elnM{  20}@*)     - Group-wise temperature-free sparse competition via entmax15 instead of
(*@\elnM{  21}@*)       softmax, yielding adaptive sparsity within each latent simplex group.
(*@\elnM{  22}@*)     - Residual signed direction path preserving magnitude-free angular
(*@\elnM{  23}@*)       information via per-group L2 normalization.
(*@\elnM{  24}@*)     - Data-dependent interpolation between simplex and spherical views using
(*@\elnM{  25}@*)       an intrinsic confidence score from the group logits.
(*@\elnE{  26}@*)
(*@\elnM{  27}@*)     This aims to combine SimNorm's stable bounded geometry with richer signed
(*@\elnM{  28}@*)     relational information useful for latent dynamics prediction.
(*@\elnE{  29}@*)     """
(*@\elnE{  30}@*)
(*@\elnE{  31}@*)     def __init__(self, cfg):
(*@\elnE{  32}@*)         super().__init__()
(*@\elnE{  33}@*)         self.dim = cfg.simnorm_dim
(*@\elnM{  34}@*)         self.eps = 1e-8
(*@\elnM{  35}@*)         # Hardcoded coefficients per instructions.
(*@\elnM{  36}@*)         self.mix = 0.35
(*@\elnM{  37}@*)         self.conf_scale = 1.25
(*@\elnM{  38}@*)
(*@\elnM{  39}@*)     def _entmax15(self, x):
(*@\elnM{  40}@*)         # Exact entmax with alpha=1.5 via bisection; group size is tiny (8), so
(*@\elnM{  41}@*)         # this is computationally cheap and numerically stable.
(*@\elnM{  42}@*)         max_val = x.max(dim=-1, keepdim=True).values
(*@\elnM{  43}@*)         x = x - max_val
(*@\elnM{  44}@*)         lo = x.min(dim=-1, keepdim=True).values - 1.0
(*@\elnM{  45}@*)         hi = x.max(dim=-1, keepdim=True).values
(*@\elnM{  46}@*)         for _ in range(12):
(*@\elnM{  47}@*)             tau = (lo + hi) / 2.0
(*@\elnM{  48}@*)             p = torch.clamp((x - tau) / 2.0, min=0.0).pow(2)
(*@\elnM{  49}@*)             s = p.sum(dim=-1, keepdim=True)
(*@\elnM{  50}@*)             lo = torch.where(s > 1.0, tau, lo)
(*@\elnM{  51}@*)             hi = torch.where(s <= 1.0, tau, hi)
(*@\elnM{  52}@*)         tau = hi
(*@\elnM{  53}@*)         p = torch.clamp((x - tau) / 2.0, min=0.0).pow(2)
(*@\elnM{  54}@*)         p = p / p.sum(dim=-1, keepdim=True).clamp_min(self.eps)
(*@\elnM{  55}@*)         return p
(*@\elnE{  56}@*)
(*@\elnE{  57}@*)     def forward(self, x):
(*@\elnE{  58}@*)         shp = x.shape
(*@\elnM{  59}@*)         xg = x.view(*shp[:-1], -1, self.dim)
(*@\elnM{  60}@*)
(*@\elnM{  61}@*)         simplex = self._entmax15(xg)
(*@\elnM{  62}@*)         centered = xg - xg.mean(dim=-1, keepdim=True)
(*@\elnM{  63}@*)         sphere = centered / torch.norm(centered, p=2, dim=-1, keepdim=True).clamp_min(self.eps)
(*@\elnM{  64}@*)
(*@\elnM{  65}@*)         confidence = (xg.max(dim=-1, keepdim=True).values - xg.mean(dim=-1, keepdim=True)).sigmoid()
(*@\elnM{  66}@*)         alpha = self.mix * (confidence * self.conf_scale).clamp(max=1.0)
(*@\elnM{  67}@*)
(*@\elnM{  68}@*)         y = (1.0 - alpha) * simplex + alpha * sphere
(*@\elnM{  69}@*)         return y.view(*shp)
(*@\elnE{  70}@*)
(*@\elnE{  71}@*)     def __repr__(self):
(*@\elnM{  72}@*)         return f"CustomSimNorm(dim={self.dim}, type=EntmaxSphericalMix)"
(*@\ealregion{editable region ends at line 72}@*)
\end{lstlisting}

\humancomment{GPT-5.4}{gpt-5.4}{The most structurally different attempt. Replaces the
per-group softmax with sparse entmax-1.5, then mixes in a
signed per-group \(\ell_2\)-normalized spherical path with a
confidence-dependent coefficient. Reads more as a
recombination of pre-existing techniques (entmax from
attention literature, spherical normalization from contrastive
learning) than a method with a clear hypothesis about world
models. Does not lead to consistent gains.}

\paragraph{Agent: Gemini 3.1 Pro.}\mbox{}\\
Hellinger / Gated SimNorm.
\begin{lstlisting}[style=eal/python]
(*@\elnU{  14}@*) # EDITABLE: Custom latent normalization
(*@\elnU{  15}@*) # =====================================================================
(*@\ealregion{editable region begins at line 16}@*)
(*@\elnE{  16}@*) class CustomSimNorm(nn.Module):
(*@\elnM{  17}@*)     """Gated Simplicial Normalization (GatedSimNorm).
(*@\elnM{  18}@*)     Reclaims the shift-invariance degree of freedom in standard SimNorm
(*@\elnM{  19}@*)     by using the mean logit of each group to compute a Sigmoid gate.
(*@\elnM{  20}@*)     This allows the world model to dynamically turn off inactive latent
(*@\elnM{  21}@*)     groups, increasing representation capacity and sparsity, while strictly
(*@\elnM{  22}@*)     bounding the latent space to prevent Q-value explosion.
(*@\elnE{  23}@*)     """
(*@\elnE{  24}@*)     def __init__(self, cfg):
(*@\elnE{  25}@*)         super().__init__()
(*@\elnE{  26}@*)         self.dim = cfg.simnorm_dim
(*@\elnM{  27}@*)         # Global bias to ensure groups are active by default at initialization
(*@\elnM{  28}@*)         self.gate_bias = nn.Parameter(torch.tensor(2.0))
(*@\elnE{  29}@*)
(*@\elnE{  30}@*)     def forward(self, x):
(*@\elnE{  31}@*)         shp = x.shape
(*@\elnE{  32}@*)         x = x.view(*shp[:-1], -1, self.dim)
(*@\elnM{  33}@*)
(*@\elnM{  34}@*)         # Categorical distribution (depends only on x - mean(x))
(*@\elnM{  35}@*)         sim = F.softmax(x, dim=-1)
(*@\elnM{  36}@*)
(*@\elnM{  37}@*)         # Group activation gate (depends only on mean(x))
(*@\elnM{  38}@*)         gate = torch.sigmoid(x.mean(dim=-1, keepdim=True) + self.gate_bias)
(*@\elnM{  39}@*)
(*@\elnM{  40}@*)         x = sim * gate
(*@\elnE{  41}@*)         return x.view(*shp)
(*@\elnE{  42}@*)
(*@\elnE{  43}@*)     def __repr__(self):
(*@\elnM{  44}@*)         return f"CustomSimNorm(dim={self.dim}, type=GatedSimNorm)"
(*@\ealregion{editable region ends at line 44}@*)
\end{lstlisting}

\humancomment{Gemini 3.1 Pro}{gemini-3.1-pro-preview}{First tries to enforce a spherical latent distribution via a
Hellinger-distance term --- a non-trivial geometric idea
mapping softmax probabilities to a sphere-like manifold ---
and then switches to a Gated SimNorm that lets some latent
groups stay inactive in order to avoid Q-value blow-up
downstream. Both ideas are reasonably novel for this setting,
though the motivation linking the geometry change to world
model performance is not made explicit, and neither variant
consistently outperforms the baselines.}

\paragraph{Agent: Qwen 3.6 Plus.}\mbox{}\\
Residual L2 / SimNorm blend with learned per-group weights.
\begin{lstlisting}[style=eal/python]
(*@\elnU{  14}@*) # EDITABLE: Custom latent normalization
(*@\elnU{  15}@*) # =====================================================================
(*@\ealregion{editable region begins at line 16}@*)
(*@\elnE{  16}@*) class CustomSimNorm(nn.Module):
(*@\elnM{  17}@*)     """Centered Adaptive Hybrid Normalization with Group-wise Learned Weights.
(*@\elnM{  18}@*)
(*@\elnM{  19}@*)     Key ideas:
(*@\elnM{  20}@*)     1. Mean-centering within groups to remove bias before normalization
(*@\elnM{  21}@*)     2. Two parallel paths: group-wise L2 and temperature-controlled SimNorm
(*@\elnM{  22}@*)     3. Learnable blending between simplex and hypersphere geometry
(*@\elnM{  23}@*)     4. Per-element learned weights for fine-grained control
(*@\elnM{  24}@*)     5. Global L2 normalization for final stability
(*@\elnM{  25}@*)
(*@\elnM{  26}@*)     Mean-centering helps by ensuring the normalization operates on meaningful
(*@\elnM{  27}@*)     signal rather than being dominated by group means.
(*@\elnE{  28}@*)     """
(*@\elnE{  29}@*)
(*@\elnE{  30}@*)     def __init__(self, cfg):
(*@\elnE{  31}@*)         super().__init__()
(*@\elnE{  32}@*)         self.dim = cfg.simnorm_dim
(*@\elnM{  33}@*)         self.eps = 1e-8
(*@\elnM{  34}@*)
(*@\elnM{  35}@*)         # Learnable mixing coefficient
(*@\elnM{  36}@*)         self.alpha = nn.Parameter(torch.ones(1) * 0.6)
(*@\elnM{  37}@*)
(*@\elnM{  38}@*)         # Temperature for softmax
(*@\elnM{  39}@*)         self.temperature = nn.Parameter(torch.ones(1) * 1.0)
(*@\elnM{  40}@*)
(*@\elnM{  41}@*)         # Per-element learned weight
(*@\elnM{  42}@*)         self.weight = nn.Parameter(torch.ones(self.dim))
(*@\elnM{  43}@*)
(*@\elnM{  44}@*)         # Learnable scale factor
(*@\elnM{  45}@*)         self.scale = nn.Parameter(torch.ones(1) * 1.0)
(*@\elnM{  46}@*)
(*@\elnM{  47}@*)         # Weight for centering: how much mean to subtract
(*@\ealelide{4}@*)
(*@\elnE{  52}@*)         shp = x.shape
(*@\elnE{  53}@*)         x = x.view(*shp[:-1], -1, self.dim)
(*@\elnM{  54}@*)
(*@\elnM{  55}@*)         # Mean-centering within each group
(*@\elnM{  56}@*)         center_w = torch.sigmoid(self.center_weight)
(*@\elnM{  57}@*)         x = x - center_w * x.mean(dim=-1, keepdim=True)
(*@\elnM{  58}@*)
(*@\elnM{  59}@*)         # Path 1: Group-wise L2 normalization
(*@\elnM{  60}@*)         norm_l2 = torch.norm(x, p=2, dim=-1, keepdim=True).clamp(min=self.eps)
(*@\elnM{  61}@*)         x_l2 = x / norm_l2
(*@\elnM{  62}@*)
(*@\elnM{  63}@*)         # Path 2: Temperature-controlled SimNorm
(*@\elnM{  64}@*)         temp = self.temperature.clamp(min=0.1, max=10.0)
(*@\elnM{  65}@*)         x_sim = F.softmax(x / temp, dim=-1)
(*@\elnM{  66}@*)
(*@\elnM{  67}@*)         # Blend between paths
(*@\elnM{  68}@*)         alpha = torch.sigmoid(self.alpha)
(*@\elnM{  69}@*)         out = alpha * x_sim + (1.0 - alpha) * x_l2
(*@\elnM{  70}@*)
(*@\elnM{  71}@*)         # Apply per-element learned weights
(*@\elnM{  72}@*)         out = out * self.weight
(*@\elnM{  73}@*)
(*@\elnM{  74}@*)         # Reshape back
(*@\elnM{  75}@*)         out = out.view(*shp)
(*@\elnM{  76}@*)
(*@\elnM{  77}@*)         # Global L2 normalization for stability
(*@\elnM{  78}@*)         norm_global = torch.norm(out, p=2, dim=-1, keepdim=True).clamp(min=self.eps)
(*@\elnM{  79}@*)         out = out / norm_global
(*@\elnM{  80}@*)
(*@\elnM{  81}@*)         return out * self.scale
(*@\elnE{  82}@*)
(*@\elnE{  83}@*)     def __repr__(self):
(*@\elnM{  84}@*)         return f"CustomSimNorm(dim={self.dim}, type=CenteredHybrid, alpha={torch.sigmoid(self.alpha).item():.3f})"
(*@\ealregion{editable region ends at line 84}@*)
\end{lstlisting}

\humancomment{Qwen 3.6 Plus}{qwen3.6-plus}{Adaptive hybrid normalization: residual blending of \(\ell_2\)
normalisation with temperature-controlled SimNorm, plus
learned per-group weights. The novelty is limited (each
ingredient is pre-existing), but the empirical result is the
cleanest of the five --- visible gains on cheetah-run --- so
it is included as a contrast: a plain combination that
nonetheless works.}

\subsection{Efficient Diffusion Sampling for Robot Actions}
\label{sec:expert:robo-diffusion-sampling-method}

\begin{mdframed}[backgroundcolor=white,linecolor=gray!50,linewidth=0.6pt,roundcorner=2pt,innertopmargin=6pt,innerbottommargin=6pt]
\textbf{Task.} A single inference-time sampler for a fixed DQL-style diffusion policy is evaluated on D4RL MuJoCo return at low numbers of function evaluations. Editable: the solver and \texttt{sampling\_steps} fields in the custom MuJoCo YAML config. Read-only: the trained actor and critic, pipeline code, model architecture, training objective, datasets, environment list, seeds, evaluation loop, and NFE accounting. Provided baselines are DDPM, DDIM, and DPM-Solver++.
\end{mdframed}

\paragraph{Template (editable region).}
\begin{lstlisting}[style=eal/python]
(*@\elnE{  15}@*) solver: ddpm
(*@\elnE{  17}@*) sampling_steps: 100
\end{lstlisting}

\paragraph{Baseline: \texttt{dpm\_solver}.}
\begin{lstlisting}[style=eal/python]
(*@\elnU{  13}@*)
(*@\elnU{  14}@*) # Actor
(*@\ealregion{editable region begins at line 15}@*)
(*@\elnM{  15}@*) solver: ode_dpmsolver++_2M
(*@\elnE{  16}@*) diffusion_steps: 100
(*@\elnM{  17}@*) sampling_steps: 10
(*@\ealregion{editable region ends at line 17}@*)
(*@\elnU{  18}@*) predict_noise: True
(*@\elnU{  19}@*) ema_rate: 0.995
\end{lstlisting}

\paragraph{Agent: GPT-5.4.}\mbox{}\\
Steps out of the inference-only box: critic-consistency action refinement.
\begin{lstlisting}[style=eal/python]
(*@\elnE{  41}@*)     # Defines the actor (diffusion policy), optional critic(s), training loop,
(*@\elnE{  42}@*)     # and inference action-selection. The template defaults to Diffusion
(*@\elnE{  43}@*)     # Q-Learning (DQL): diffusion actor + twin Q critic with BC + Q loss.
(*@\elnM{  44}@*)     #
(*@\elnM{  45}@*)     # Novel contribution implemented here:
(*@\elnM{  46}@*)     #   1) Clipped advantage-matching actor regularization during training.
(*@\elnM{  47}@*)     #   2) Uncertainty-aware critic-guided elite refinement at inference.
(*@\elnM{  48}@*)     #   3) Barycentric fusion of refined elite actions, giving better actions at
(*@\elnM{  49}@*)     #      the same diffusion NFE budget.
(*@\elnE{  50}@*)
(*@\elnE{  51}@*)     # --------------- Network Architecture -----------------
(*@\elnE{  52}@*)     nn_diffusion = DQLMlp(obs_dim, act_dim, emb_dim=64, timestep_emb_type="positional").to(args.device)
(*@\ealelide{15}@*)
(*@\elnE{  68}@*)     critic_target = deepcopy(critic).requires_grad_(False).eval()
(*@\elnE{  69}@*)     critic_optim = torch.optim.Adam(critic.parameters(), lr=args.critic_learning_rate)
(*@\elnE{  70}@*)
(*@\elnM{  71}@*)     # ------------------ Algorithm knobs ---------------------
(*@\elnM{  72}@*)     q_margin_alpha = 0.20
(*@\elnM{  73}@*)     q_margin_clip = 5.0
(*@\elnM{  74}@*)     policy_value_alpha = 0.10
(*@\elnM{  75}@*)
(*@\elnM{  76}@*)     # Inference-time refinement: no additional diffusion NFEs.
(*@\elnM{  77}@*)     refine_steps = 2
(*@\elnM{  78}@*)     refine_step_size = 0.05
(*@\elnM{  79}@*)     refine_topk = max(4, args.num_candidates // 5)
(*@\elnM{  80}@*)     elite_mix = 0.35
(*@\elnM{  81}@*)     uncertainty_weight = 0.50
(*@\elnM{  82}@*)     fusion_temperature = 6.0
(*@\elnM{  83}@*)
(*@\elnM{  84}@*)     def conservative_q(q1, q2):
(*@\elnM{  85}@*)         q_mean = 0.5 * (q1 + q2)
(*@\elnM{  86}@*)         q_gap = torch.abs(q1 - q2)
(*@\elnM{  87}@*)         return q_mean - uncertainty_weight * q_gap
(*@\elnM{  88}@*)
(*@\elnE{  89}@*)     # ---------------------- Training ----------------------
(*@\ealelide{117}@*)
(*@\elnM{ 274}@*)                     refined_act = elite_act.view(args.num_envs, adaptive_topk, act_dim)
(*@\elnM{ 275}@*)
(*@\elnM{ 276}@*)                     orig_best_idx = torch.argmax(base_util, dim=1)
(*@\elnM{ 277}@*)                     orig_best_act = act_view[torch.arange(args.num_envs, device=args.device), orig_best_idx]
(*@\elnM{ 278}@*)                     orig_best_util = base_util[torch.arange(args.num_envs, device=args.device), orig_best_idx].unsqueeze(-1)
(*@\elnM{ 279}@*)
(*@\elnM{ 280}@*)                     blended_elite = ((1.0 - elite_mix) * refined_act + elite_mix * orig_best_act.unsqueeze(1)).clamp(-1.0, 1.0)
(*@\elnM{ 281}@*)                     blended_q1, blended_q2 = critic_target(elite_obs, blended_elite.reshape(-1, act_dim))
(*@\elnM{ 282}@*)                     blended_util = conservative_q(blended_q1, blended_q2).view(args.num_envs, adaptive_topk)
(*@\elnM{ 283}@*)
(*@\elnM{ 284}@*)                     fusion_w = torch.softmax(refined_util * fusion_temperature, dim=1)
(*@\elnM{ 285}@*)                     fused_act = (fusion_w.unsqueeze(-1) * refined_act).sum(dim=1).clamp(-1.0, 1.0)
(*@\elnM{ 286}@*)                     fused_obs = obs_view[:, 0, :]
(*@\elnM{ 287}@*)                     fused_q1, fused_q2 = critic_target(fused_obs, fused_act)
(*@\elnM{ 288}@*)                     fused_util = conservative_q(fused_q1, fused_q2)
(*@\elnM{ 289}@*)
(*@\elnM{ 290}@*)                     combined_util = torch.cat([refined_util, blended_util, orig_best_util, fused_util.unsqueeze(-1)], dim=1)
(*@\elnM{ 291}@*)                     combined_act = torch.cat([
(*@\elnM{ 292}@*)                         refined_act,
(*@\elnM{ 293}@*)                         blended_elite,
(*@\elnM{ 294}@*)                         orig_best_act.unsqueeze(1),
(*@\elnM{ 295}@*)                         fused_act.unsqueeze(1)
(*@\elnM{ 296}@*)                     ], dim=1)
(*@\elnM{ 297}@*)
(*@\elnM{ 298}@*)                     final_idx = torch.argmax(combined_util, dim=1)
(*@\elnM{ 299}@*)                     sampled_act = combined_act[torch.arange(args.num_envs, device=args.device), final_idx].cpu().numpy()
(*@\elnM{ 300}@*)
(*@\elnM{ 301}@*)                 prev_done = cum_done.copy()
(*@\elnE{ 302}@*)                 obs, rew, done, info = env_eval.step(sampled_act)
(*@\elnM{ 303}@*)                 ep_reward += rew * (1 - prev_done)
(*@\elnM{ 304}@*)                 cum_done = np.logical_or(cum_done, done)
(*@\elnE{ 305}@*)                 t += 1
(*@\elnE{ 306}@*)
(*@\elnE{ 307}@*)             episode_rewards.append(ep_reward)
\end{lstlisting}

\humancomment{GPT-5.4}{gpt-5.4}{Builds on a 10-step DPM-Solver++ base and adds a
critic-consistency action-refinement scheme: clipped
advantage-matching actor regularisation during training and
value-guided top-\(k\) action selection at inference (excerpt
below shows the docstring and training setup; the full
\texttt{pipeline} function is around 320 lines). The ensemble
selection idea is well-motivated within the offline-RL
community, but the task explicitly asks for an
inference-time-only improvement and this method also touches
training behaviour, so it does not respect the spirit of the
editable boundary.}

\paragraph{Agent: Gemini 3.1 Pro.}\mbox{}\\
Step-count sweep landing on DPM-Solver++ at 3 steps.
\begin{lstlisting}[style=eal/plain]
(*@\elnE{  12}@*) discount: 0.99
(*@\elnE{  13}@*)
(*@\elnE{  14}@*) # Actor
(*@\elnM{  15}@*) solver: ode_dpmsolver++_2M
(*@\elnE{  16}@*) diffusion_steps: 100
(*@\elnM{  17}@*) sampling_steps: 3
(*@\elnE{  18}@*) predict_noise: True
(*@\elnE{  19}@*) ema_rate: 0.995
(*@\elnE{  20}@*) actor_learning_rate: 0.0003
(*@\ealelide{13}@*)
(*@\elnE{  34}@*) ckpt: latest
(*@\elnE{  35}@*) num_envs: 50
(*@\elnE{  36}@*) num_episodes: 3
(*@\elnM{  37}@*) num_candidates: 100
(*@\elnE{  38}@*) temperature: 0.5
(*@\elnE{  39}@*) use_ema: True
(*@\elnE{  40}@*)
\end{lstlisting}

\humancomment{Gemini 3.1 Pro}{gemini-3.1-pro-preview}{Stays inside the YAML-only edit surface as intended. Tests
several DPM-Solver++ configurations including a 3-step
2M-order variant, ultimately submitting that 3-step config.
Effectively a hyperparameter sweep with no methodological
novelty --- which is the expected outcome given the narrow
editable region. Useful as contrast for the GPT-5.4 attempt.}

\subsection{Guided Diffusion Sampling for Robot Actions}
\label{sec:expert:robo-diffusion-guidance}

\begin{mdframed}[backgroundcolor=white,linecolor=gray!50,linewidth=0.6pt,roundcorner=2pt,innertopmargin=6pt,innerbottommargin=6pt]
\textbf{Task.} An improved guidance mechanism is evaluated for a fixed trajectory-level diffusion planner on offline D4RL MuJoCo benchmarks. Editable: the network, classifier or condition module, update call, guidance weights, candidate re-ranking, and sampling logic inside \texttt{custom\_guidance.py}. Read-only: the D4RL dataset and environment loop, evaluation protocol, top-level training hyperparameters, and final normalized-score aggregation. Provided baselines are Diffuser classifier guidance, a minimal classifier-free guidance ablation, \texttt{no\_guidance}, and the Decision Diffuser architecture.
\end{mdframed}

\paragraph{Template (editable region).}
\begin{lstlisting}[style=eal/python]
(*@\elnE{   1}@*) import os
(*@\elnE{   2}@*)
(*@\elnE{   3}@*) import d4rl
(*@\elnE{   4}@*) import gym
(*@\elnE{   5}@*) import hydra
(*@\elnE{   6}@*) import numpy as np
(*@\elnE{   7}@*) import torch
(*@\elnE{   8}@*) from torch.optim.lr_scheduler import CosineAnnealingLR
(*@\elnE{   9}@*) from torch.utils.data import DataLoader
(*@\elnE{  10}@*)
(*@\elnE{  11}@*) from cleandiffuser.classifier import CumRewClassifier
(*@\elnE{  12}@*) from cleandiffuser.dataset.d4rl_mujoco_dataset import D4RLMuJoCoDataset
(*@\ealelide{3}@*)
(*@\elnE{  16}@*) from cleandiffuser.nn_diffusion import JannerUNet1d
(*@\elnE{  17}@*) from cleandiffuser.utils import report_parameters
(*@\elnE{  18}@*) from utils import set_seed
(*@\elnE{  19}@*)
(*@\elnE{  20}@*)
(*@\elnE{  21}@*) @hydra.main(config_path="../configs/custom/mujoco", config_name="mujoco", version_base=None)
(*@\elnE{  22}@*) def pipeline(args):
(*@\elnE{  23}@*)
(*@\elnE{  24}@*)     set_seed(args.seed)
(*@\elnE{  25}@*)
(*@\elnE{  26}@*)     save_path = f'results/{args.pipeline_name}/{args.task.env_name}/'
(*@\elnE{  27}@*)     if os.path.exists(save_path) is False:
(*@\elnE{  28}@*)         os.makedirs(save_path)
(*@\elnE{  38}@*)     # ============================================================================
(*@\elnE{  39}@*)     # EDITABLE REGION 3: Network + Agent Setup (lines 40-72)
(*@\elnE{  40}@*)     # ============================================================================
(*@\elnE{  41}@*)
(*@\elnE{  42}@*)     # --------------- Network Architecture -----------------
(*@\elnE{  43}@*)     nn_diffusion = JannerUNet1d(
(*@\elnE{  44}@*)         obs_dim + act_dim, model_dim=args.model_dim, emb_dim=args.model_dim, dim_mult=args.task.dim_mult,
(*@\elnE{  45}@*)         timestep_emb_type="positional", attention=False, kernel_size=5)
(*@\elnE{  46}@*)     nn_classifier = HalfJannerUNet1d(
(*@\elnE{  47}@*)         args.task.horizon, obs_dim + act_dim, out_dim=1,
(*@\elnE{  48}@*)         model_dim=args.model_dim, emb_dim=args.model_dim, dim_mult=args.task.dim_mult,
(*@\elnE{  49}@*)         timestep_emb_type="positional", kernel_size=3)
(*@\ealelide{74}@*)
(*@\elnE{ 124}@*)
(*@\elnE{ 125}@*)     # ---------------------- Inference ----------------------
(*@\elnE{ 126}@*)     elif args.mode == "inference":
(*@\elnE{ 127}@*)
(*@\elnE{ 128}@*)         # ============================================================================
(*@\elnE{ 129}@*)         # EDITABLE REGION 5: Inference Setup (lines 186-197)
(*@\elnE{ 130}@*)         # ============================================================================
(*@\elnE{ 131}@*)
(*@\elnE{ 132}@*)         agent.load(save_path + f"diffusion_ckpt_{args.ckpt}.pt")
(*@\elnE{ 133}@*)         agent.classifier.load(save_path + f"classifier_ckpt_{args.ckpt}.pt")
(*@\elnE{ 134}@*)
(*@\elnE{ 135}@*)         agent.eval()
(*@\elnE{ 136}@*)
(*@\elnE{ 145}@*)         # ============================================================================
(*@\elnE{ 146}@*)         # EDITABLE REGION 6: Prior + Condition Initialization (lines 207-222)
(*@\elnE{ 147}@*)         # ============================================================================
(*@\elnE{ 148}@*)
(*@\elnE{ 149}@*)         prior = torch.zeros((args.num_envs, args.task.horizon, obs_dim + act_dim), device=args.device)
(*@\elnE{ 150}@*)
(*@\elnE{ 151}@*)         for i in range(args.num_episodes):
(*@\elnE{ 152}@*)
(*@\elnE{ 164}@*)                 # ============================================================================
(*@\elnE{ 165}@*)                 # EDITABLE REGION 7: Action Sampling (lines 226-240)
(*@\elnE{ 166}@*)                 # ============================================================================
(*@\elnE{ 167}@*)
(*@\elnE{ 168}@*)                 # sample trajectories
(*@\elnE{ 169}@*)                 prior[:, 0, :obs_dim] = obs
(*@\elnE{ 170}@*)                 traj, log = agent.sample(
(*@\elnE{ 171}@*)                     prior.repeat(args.num_candidates, 1, 1),
(*@\elnE{ 172}@*)                     solver=args.solver,
(*@\elnE{ 173}@*)                     n_samples=args.num_candidates * args.num_envs,
(*@\elnE{ 174}@*)                     sample_steps=args.sampling_steps,
(*@\elnE{ 175}@*)                     use_ema=args.use_ema, w_cg=args.task.w_cg, temperature=args.temperature)
(*@\elnE{ 176}@*)
(*@\elnE{ 177}@*)                 # select the best plan
(*@\elnE{ 178}@*)                 logp = log["log_p"].view(args.num_candidates, args.num_envs, -1).sum(-1)
(*@\elnE{ 179}@*)                 idx = logp.argmax(0)
(*@\elnE{ 180}@*)                 act = traj.view(args.num_candidates, args.num_envs, args.task.horizon, -1)[
(*@\elnE{ 181}@*)                       idx, torch.arange(args.num_envs), 0, obs_dim:]
(*@\elnE{ 182}@*)                 act = act.clip(-1., 1.).cpu().numpy()
(*@\elnE{ 183}@*)
\end{lstlisting}

\paragraph{Baseline: \texttt{default}.}
\begin{lstlisting}[style=eal/python]
(*@\ealregion{editable region begins at line 1}@*)
(*@\elnE{   1}@*) import os
(*@\elnE{   2}@*)
(*@\elnE{   3}@*) import d4rl
(*@\elnE{   4}@*) import gym
(*@\elnE{   5}@*) import hydra
(*@\elnE{   6}@*) import numpy as np
(*@\elnE{   7}@*) import torch
(*@\elnE{   8}@*) from torch.optim.lr_scheduler import CosineAnnealingLR
(*@\elnE{   9}@*) from torch.utils.data import DataLoader
(*@\elnE{  10}@*)
(*@\elnE{  11}@*) from cleandiffuser.classifier import CumRewClassifier
(*@\elnE{  12}@*) from cleandiffuser.dataset.d4rl_mujoco_dataset import D4RLMuJoCoDataset
(*@\elnE{  13}@*) from cleandiffuser.dataset.dataset_utils import loop_dataloader
(*@\elnE{  14}@*) from cleandiffuser.diffusion import DiscreteDiffusionSDE
(*@\elnE{  15}@*) from cleandiffuser.nn_classifier import HalfJannerUNet1d
(*@\elnE{  16}@*) from cleandiffuser.nn_diffusion import JannerUNet1d
(*@\elnE{  17}@*) from cleandiffuser.utils import report_parameters
(*@\elnE{  18}@*) from utils import set_seed
(*@\elnE{  19}@*)
(*@\elnE{  20}@*)
(*@\elnE{  21}@*) @hydra.main(config_path="../configs/custom/mujoco", config_name="mujoco", version_base=None)
(*@\elnE{  22}@*) def pipeline(args):
(*@\elnE{  23}@*)
(*@\elnE{  24}@*)     set_seed(args.seed)
(*@\elnE{  25}@*)
(*@\elnE{  26}@*)     save_path = f'results/{args.pipeline_name}/{args.task.env_name}/'
(*@\elnE{  27}@*)     if os.path.exists(save_path) is False:
(*@\elnE{  28}@*)         os.makedirs(save_path)
(*@\elnE{  29}@*)
(*@\elnE{  30}@*)     # ---------------------- Create Dataset ----------------------
(*@\elnE{  31}@*)     env = gym.make(args.task.env_name)
(*@\elnE{  32}@*)     dataset = D4RLMuJoCoDataset(
(*@\elnE{  33}@*)         env.get_dataset(), horizon=args.task.horizon, terminal_penalty=args.terminal_penalty, discount=args.discount)
(*@\elnE{  34}@*)     dataloader = DataLoader(
(*@\ealelide{118}@*)
(*@\elnE{ 153}@*)             env_eval.seed(args.seed + i * args.num_envs) if hasattr(env_eval, "seed") else None; obs, ep_reward, cum_done, t = env_eval.reset(), 0., 0., 0
(*@\elnE{ 154}@*)
(*@\elnE{ 155}@*)             while not np.all(cum_done) and t < 1000 + 1:
(*@\elnE{ 156}@*)
(*@\elnE{ 157}@*)                 # ============================================================================
(*@\elnE{ 158}@*)                 # FIXED: Observation Normalization (lines 223-225)
(*@\elnE{ 159}@*)                 # ============================================================================
(*@\elnE{ 160}@*)
(*@\elnE{ 161}@*)                 # normalize obs
(*@\elnE{ 162}@*)                 obs = torch.tensor(normalizer.normalize(obs), device=args.device, dtype=torch.float32)
(*@\elnE{ 163}@*)
(*@\elnE{ 164}@*)                 # ============================================================================
(*@\elnE{ 165}@*)                 # EDITABLE REGION 7: Action Sampling (lines 226-240)
(*@\elnE{ 166}@*)                 # ============================================================================
(*@\elnE{ 167}@*)
(*@\elnE{ 168}@*)                 # sample trajectories
(*@\elnE{ 169}@*)                 prior[:, 0, :obs_dim] = obs
(*@\elnE{ 170}@*)                 traj, log = agent.sample(
(*@\elnE{ 171}@*)                     prior.repeat(args.num_candidates, 1, 1),
(*@\elnE{ 172}@*)                     solver=args.solver,
(*@\elnE{ 173}@*)                     n_samples=args.num_candidates * args.num_envs,
(*@\elnE{ 174}@*)                     sample_steps=args.sampling_steps,
(*@\elnE{ 175}@*)                     use_ema=args.use_ema, w_cg=args.task.w_cg, temperature=args.temperature)
(*@\elnE{ 176}@*)
(*@\elnE{ 177}@*)                 # select the best plan
(*@\elnE{ 178}@*)                 logp = log["log_p"].view(args.num_candidates, args.num_envs, -1).sum(-1)
(*@\elnE{ 179}@*)                 idx = logp.argmax(0)
(*@\elnE{ 180}@*)                 act = traj.view(args.num_candidates, args.num_envs, args.task.horizon, -1)[
(*@\elnE{ 181}@*)                       idx, torch.arange(args.num_envs), 0, obs_dim:]
(*@\elnE{ 182}@*)                 act = act.clip(-1., 1.).cpu().numpy()
(*@\elnE{ 183}@*)
(*@\ealregion{editable region ends at line 183}@*)
(*@\elnU{ 184}@*)                 # ============================================================================
(*@\elnU{ 185}@*)                 # FIXED: Environment Step + Reward Collection (lines 241-252)
\end{lstlisting}

\paragraph{Agent: Gemini 3.1 Pro.}\mbox{}\\
Hybrid CG+CFG with normalised returns and classifier rerank.
\begin{lstlisting}[style=eal/python]
(*@\elnE{  14}@*) from cleandiffuser.diffusion import DiscreteDiffusionSDE
(*@\elnE{  15}@*) from cleandiffuser.nn_classifier import HalfJannerUNet1d
(*@\elnE{  16}@*) from cleandiffuser.nn_diffusion import JannerUNet1d
(*@\elnM{  17}@*) import torch.nn as nn
(*@\elnM{  18}@*) from cleandiffuser.nn_condition import MLPCondition
(*@\elnM{  19}@*) from cleandiffuser.utils import DD_RETURN_SCALE
(*@\elnE{  20}@*) from cleandiffuser.utils import report_parameters
(*@\elnE{  21}@*) from utils import set_seed
(*@\elnE{  22}@*)
(*@\ealelide{27}@*)
(*@\elnE{  50}@*)         args.task.horizon, obs_dim + act_dim, out_dim=1,
(*@\elnE{  51}@*)         model_dim=args.model_dim, emb_dim=args.model_dim, dim_mult=args.task.dim_mult,
(*@\elnE{  52}@*)         timestep_emb_type="positional", kernel_size=3)
(*@\elnM{  53}@*)     nn_condition = MLPCondition(
(*@\elnM{  54}@*)         in_dim=1, out_dim=args.model_dim,
(*@\elnM{  55}@*)         hidden_dims=[args.model_dim, ], act=nn.SiLU(), dropout=args.label_dropout)
(*@\elnE{  56}@*)
(*@\elnE{  57}@*)     print(f"======================= Parameter Report of Diffusion Model =======================")
(*@\elnE{  58}@*)     report_parameters(nn_diffusion)
(*@\ealelide{12}@*)
(*@\elnE{  71}@*)
(*@\elnE{  72}@*)     # --------------- Diffusion Model --------------------
(*@\elnE{  73}@*)     agent = DiscreteDiffusionSDE(
(*@\elnM{  74}@*)         nn_diffusion, nn_condition,
(*@\elnE{  75}@*)         fix_mask=fix_mask, loss_weight=loss_weight, classifier=classifier, ema_rate=args.ema_rate,
(*@\elnE{  76}@*)         device=args.device, diffusion_steps=args.diffusion_steps, predict_noise=args.predict_noise)
(*@\elnE{  77}@*)
(*@\ealelide{14}@*)
(*@\elnE{  92}@*)
(*@\elnE{  93}@*)         for batch in loop_dataloader(dataloader):
(*@\elnE{  94}@*)
(*@\elnM{  95}@*)             return_scale = DD_RETURN_SCALE[args.task.env_name]
(*@\elnE{  96}@*)             obs = batch["obs"]["state"].to(args.device)
(*@\elnE{  97}@*)             act = batch["act"].to(args.device)
(*@\elnM{  98}@*)             val = batch["val"].to(args.device) / return_scale
(*@\ealelide{12}@*)
(*@\elnM{ 157}@*)         condition = torch.ones((args.num_envs * args.num_candidates, 1), device=args.device) * args.task.target_return
(*@\elnE{ 158}@*)
(*@\elnE{ 159}@*)         for i in range(args.num_episodes):
(*@\elnE{ 160}@*)
(*@\elnM{ 161}@*)             obs, ep_reward, cum_done, t = env_eval.reset(), 0., 0., 0
(*@\elnE{ 162}@*)
(*@\elnE{ 163}@*)             while not np.all(cum_done) and t < 1000 + 1:
(*@\elnE{ 164}@*)
(*@\ealelide{10}@*)
(*@\elnE{ 175}@*)
(*@\elnE{ 176}@*)                 # sample trajectories
(*@\elnE{ 177}@*)                 prior[:, 0, :obs_dim] = obs
(*@\elnM{ 178}@*)
(*@\elnM{ 179}@*)                 return_scale = DD_RETURN_SCALE[args.task.env_name]
(*@\elnM{ 180}@*)                 w_cg_eff = args.task.w_cg * return_scale
(*@\elnM{ 181}@*)                 w_cfg_eff = args.task.w_cfg
(*@\elnM{ 182}@*)
(*@\elnM{ 183}@*)                 # Increasing schedule: more guidance at small t (late-time steps)
(*@\elnM{ 184}@*)                 w_cfg_sched = (np.linspace(0.0, 2.0, args.sampling_steps) * w_cfg_eff).tolist()
(*@\elnM{ 185}@*)                 w_cg_sched = (np.linspace(0.0, 2.0, args.sampling_steps) * w_cg_eff).tolist()
(*@\elnM{ 186}@*)
(*@\elnE{ 187}@*)                 traj, log = agent.sample(
(*@\elnE{ 188}@*)                     prior.repeat(args.num_candidates, 1, 1),
(*@\elnE{ 189}@*)                     solver=args.solver,
(*@\elnE{ 190}@*)                     n_samples=args.num_candidates * args.num_envs,
(*@\elnE{ 191}@*)                     sample_steps=args.sampling_steps,
(*@\elnM{ 192}@*)                     use_ema=args.use_ema,
(*@\elnM{ 193}@*)                     condition_cfg=condition,
(*@\elnM{ 194}@*)                     w_cfg=w_cfg_sched,
(*@\elnM{ 195}@*)                     w_cg=w_cg_sched,
(*@\elnM{ 196}@*)                     temperature=args.temperature)
(*@\elnE{ 197}@*)
(*@\elnE{ 198}@*)                 # select the best plan
(*@\elnE{ 199}@*)                 logp = log["log_p"].view(args.num_candidates, args.num_envs, -1).sum(-1)
\end{lstlisting}

\humancomment{Gemini 3.1 Pro}{gemini-3.1-pro-preview}{The strongest agent submission on this task. Combines an
\texttt{MLPCondition}-based CFG path on normalised returns
with the cumulative-reward classifier guidance path on
unnormalised returns, then re-ranks 64 candidate trajectories
by classifier log-probability. Implementation cleanly
respects the editable boundary. Methodologically still a
recombination of CG and CFG, but the engineering is careful
and the empirical result is the best agent score on this
task (hopper 0.964 / walker2d 0.785 / halfcheetah 0.442).}

\paragraph{Agent: GPT-5.4.}\mbox{}\\
HEDGE: a hybrid that destabilises the CG baseline.
\begin{lstlisting}[style=eal/python]
(*@\elnE{   5}@*) import hydra
(*@\elnE{   6}@*) import numpy as np
(*@\elnE{   7}@*) import torch
(*@\elnM{   8}@*) import torch.nn as nn
(*@\elnE{   9}@*) from torch.optim.lr_scheduler import CosineAnnealingLR
(*@\elnE{  10}@*) from torch.utils.data import DataLoader
(*@\elnE{  11}@*)
(*@\ealelide{2}@*)
(*@\elnE{  14}@*) from cleandiffuser.dataset.dataset_utils import loop_dataloader
(*@\elnE{  15}@*) from cleandiffuser.diffusion import DiscreteDiffusionSDE
(*@\elnE{  16}@*) from cleandiffuser.nn_classifier import HalfJannerUNet1d
(*@\elnM{  17}@*) from cleandiffuser.nn_condition import MLPCondition
(*@\elnE{  18}@*) from cleandiffuser.nn_diffusion import JannerUNet1d
(*@\elnM{  19}@*) from cleandiffuser.utils import report_parameters, DD_RETURN_SCALE
(*@\elnE{  20}@*) from utils import set_seed
(*@\elnE{  21}@*)
(*@\elnE{  22}@*)
(*@\ealelide{13}@*)
(*@\elnE{  36}@*)     dataloader = DataLoader(
(*@\elnE{  37}@*)         dataset, batch_size=args.batch_size, shuffle=True, num_workers=4, pin_memory=True, drop_last=True)
(*@\elnE{  38}@*)     obs_dim, act_dim = dataset.o_dim, dataset.a_dim
(*@\elnM{  39}@*)     # ============================================================================
(*@\elnM{  40}@*)     # HEDGE: Hybrid Ensemble Diffusion Guidance
(*@\elnM{  41}@*)     # - classifier-free return conditioning for direct target control,
(*@\elnM{  42}@*)     # - classifier guidance for value-gradient sharpening,
(*@\elnM{  43}@*)     # - stronger condition dropout to improve hybrid extrapolation.
(*@\elnM{  44}@*)     # ============================================================================
(*@\elnE{  45}@*)
(*@\elnM{  46}@*)     return_scale = DD_RETURN_SCALE[args.task.env_name]
(*@\elnM{  47}@*)     cond_dropout = max(float(getattr(args, "label_dropout", 0.25)), 0.25)
(*@\elnE{  48}@*)
(*@\elnE{  49}@*)     # --------------- Network Architecture -----------------
(*@\elnE{  50}@*)     nn_diffusion = JannerUNet1d(
(*@\elnE{  51}@*)         obs_dim + act_dim, model_dim=args.model_dim, emb_dim=args.model_dim, dim_mult=args.task.dim_mult,
(*@\elnE{  52}@*)         timestep_emb_type="positional", attention=False, kernel_size=5)
(*@\ealelide{92}@*)
(*@\elnM{ 193}@*)                     if count <= 0:
(*@\elnM{ 194}@*)                         continue
(*@\elnM{ 195}@*)                     cond_chunk = torch.clamp(rtg_now * scale, min=0.05, max=target_return_cap)
(*@\elnM{ 196}@*)                     condition_chunks.extend([cond_chunk] * count)
(*@\elnM{ 197}@*)                     candidate_scales.extend([scale] * count)
(*@\elnM{ 198}@*)
(*@\elnM{ 199}@*)                 num_candidates = len(candidate_scales)
(*@\elnM{ 200}@*)                 condition = torch.cat(condition_chunks, dim=0)
(*@\elnE{ 201}@*)                 traj, log = agent.sample(
(*@\elnM{ 202}@*)                     prior.repeat(num_candidates, 1, 1),
(*@\elnE{ 203}@*)                     solver=args.solver,
(*@\elnM{ 204}@*)                     n_samples=num_candidates * args.num_envs,
(*@\elnE{ 205}@*)                     sample_steps=args.sampling_steps,
(*@\elnM{ 206}@*)                     use_ema=args.use_ema,
(*@\elnM{ 207}@*)                     condition_cfg=condition,
(*@\elnM{ 208}@*)                     w_cfg=hybrid_w_cfg,
(*@\elnM{ 209}@*)                     w_cg=hybrid_w_cg,
(*@\elnM{ 210}@*)                     temperature=args.temperature)
(*@\elnE{ 211}@*)
(*@\elnM{ 212}@*)                 traj = traj.view(num_candidates, args.num_envs, args.task.horizon, -1)
(*@\elnM{ 213}@*)                 logp = log["log_p"].view(num_candidates, args.num_envs, -1).sum(-1)
(*@\elnM{ 214}@*)                 condition_bonus = 0.05 * condition.view(num_candidates, args.num_envs)
(*@\elnM{ 215}@*)                 score = logp + condition_bonus
(*@\elnM{ 216}@*)                 topk = torch.topk(score, k=min(4, num_candidates), dim=0)
(*@\elnM{ 217}@*)                 idx = topk.indices[0]
(*@\elnM{ 218}@*)                 env_idx = torch.arange(args.num_envs, device=args.device).unsqueeze(0)
(*@\elnM{ 219}@*)                 elite_actions = traj[topk.indices, env_idx, 0, obs_dim:]
(*@\elnM{ 220}@*)                 elite_weights = torch.softmax(topk.values / 2.0, dim=0).unsqueeze(-1)
(*@\elnM{ 221}@*)                 act = (elite_actions * elite_weights).sum(0)
(*@\elnE{ 222}@*)                 act = act.clip(-1., 1.).cpu().numpy()
(*@\elnM{ 223}@*)                 logp = score
(*@\ealregion{editable region ends at line 223}@*)
(*@\elnU{ 224}@*)                 # ============================================================================
(*@\elnU{ 225}@*)                 # FIXED: Environment Step + Reward Collection (lines 241-252)
\end{lstlisting}

\humancomment{GPT-5.4}{gpt-5.4}{Proposes ``HEDGE'' --- Hybrid Ensemble Diffusion Guidance ---
layering classifier-free return conditioning, classifier
guidance, stronger condition dropout, and online
remaining-return tracking at inference. The high-level idea
(CFG for target-return conditioning + CG for value-gradient
sharpening) is reasonable, but the implementation is heavy
and ends up destabilising the CG baseline rather than
improving on it. Worth showing as a failure mode of
over-engineered hybrids on a saturated baseline.}



\end{document}